\documentclass{article}

 \usepackage[eandd, final]{neurips_2026}

\usepackage[utf8]{inputenc} %
\usepackage[T1]{fontenc}    %
\usepackage{hyperref}       %
\usepackage{url}            %
\usepackage{booktabs}       %
\usepackage{amsfonts}       %
\usepackage{nicefrac}       %
\usepackage{microtype}      %
\usepackage{xcolor}         %

\usepackage{amsmath}
\usepackage{amssymb}
\usepackage{mathtools}
\usepackage{amsthm}
\usepackage{graphicx} %
\usepackage{xspace}
\usepackage{array}
\usepackage{adjustbox} %
\usepackage{makecell}
\usepackage{pifont}
\usepackage{textcomp}
\usepackage{rotating}
\usepackage{enumitem}
\usepackage{fvextra}
\usepackage{tcolorbox}
\usepackage{tabularx}
\usepackage{makecell}

\usepackage[capitalize,noabbrev]{cleveref}

\newcommand{\ours}{{\sc TALES}\xspace}

\newcommand{\textworld}{{\sc TextWorld}\xspace}
\newcommand{\jericho}{{\sc Jericho}\xspace}
\newcommand{\twx}{{\sc TextWorldExpress}\xspace}
\newcommand{\alfworld}{{\sc ALFWorld}\xspace}
\newcommand{\scienceworld}{{\sc ScienceWorld}\xspace}
\newcommand{\abbrtextworld}{{\sc TW}\xspace}
\newcommand{\abbrtwx}{{\sc TWX}\xspace}
\newcommand{\abbralfworld}{{\sc AW}\xspace}
\newcommand{\abbrscienceworld}{{\sc SW}\xspace}

\newcommand{\zork}{{\sc Zork1}\xspace}
\newcommand{\simon}{{\sc Simon Says}\xspace}
\newcommand{\simonwm}{{\sc Simon Says with Memory}\xspace}
\newcommand{\nummodels}{76\xspace}
\newcommand{\numtranscripts}{47,022\xspace}
\newcommand{\numlabeledlogs}{28\xspace}
\newcommand{\numlabeledgames}{7\xspace}
\newcommand{\numgames}{122\xspace}
\newcommand{\astra}{gpt-6-astra\xspace}
\newcommand{\opusfivefive}{claude-opus-5.5\xspace}
\usepackage{xcolor}

\definecolor{darkred}{HTML}{C00000}
\definecolor{darkgreen}{HTML}{005e19}
\definecolor{darkblue}{HTML}{240394}
\newcommand{\code}[1]{\texttt{#1}}
\newcommand{\cmd}[1]{\textcolor{darkgreen}{\textbf{{\code{#1}}}}}

\title{\ours: Text Adventure Learning Environment Suite}

\author{%
  Christopher Zhang Cui\textsuperscript{1}\:\:\:\:Xingdi Yuan\textsuperscript{2}\:\:\:\:Ziang Xiao\textsuperscript{3} \\
  \textbf{Prithviraj Ammanabrolu\textsuperscript{1}\:\:\:\:Marc-Alexandre Côté\textsuperscript{2}} \\
  \textsuperscript{1}University of California, San Diego\:\:\:\:
  \textsuperscript{2}Microsoft Research Montréal\:\:\:\:\\
  \textsuperscript{3}Johns Hopkins University \\
  \texttt{czcui@ucsd.edu}\\ 
  \texttt{textworld@microsoft.com}
}

\begin{document}

\maketitle

\begin{abstract}
While Large Language Models (LLMs) are increasingly deployed as agents to complete complex real-world tasks, assessing their baseline agentic capabilities remains challenging.
Many existing benchmarks for LLM agents require the model to operate through structured interfaces that sit one layer removed from natural language, making it difficult to distinguish failures of interface interpretation from genuine limitations in agentic capability.
Because they rely entirely on grounded, natural-language interaction, text-adventure games serve as a prime sandbox for evaluating these capabilities.
However, current implementations in popular benchmarks obscure performance through the leakage of privileged information or through scaffolding modifications to the tasks themselves.
To solve this, we introduce \ours, a diverse collection of synthetic and human-written text-adventure games, presented in their canonical forms, designed to evaluate baseline agentic competencies in progressively more challenging environments.
We present results over a range of LLMs, open- and closed-weight; evaluate when and where current models fail as agents; and perform a qualitative analysis of top agents to identify key failure modes.
Despite an impressive showing on synthetic games, even the top LLM-driven agents fail to achieve 25\% on games designed for human enjoyment within 100 steps.
Code and visualizations of the experiments can be found at \textcolor{darkblue}{\url{https://github.com/microsoft/tale-suite}}.
\end{abstract}

\section{Introduction}\label{sec:intro}

Despite being trained primarily on ungrounded language, Large Language Models (LLMs) are increasingly deployed as agents in complex, situated environments.
This mismatch becomes apparent when LLM agents struggle even in relatively simple environments with well-defined dynamics. 
In real-world applications, with longer horizons, stricter constraints, and higher costs for failure, the ability of LLMs to reliably operate under these conditions becomes paramount.

Many agentic benchmarks in high-impact domains exist \citep{jimenez2024swebench, OSWorld, küttler2020nethacklearningenvironment, patil2025bfcl, chen2024scienceagentbenchrigorousassessmentlanguage}.
However, when measuring agentic capabilities, these frameworks share a fundamental limitation: agents must operate through structured interfaces that deviate from natural language, such as code, function schemas, or symbolic representations.
These abstractions sit one level removed from the modality LLMs are trained on, making failure attribution ambiguous.
When an agent fails, it is unclear whether the failure stems from limitations in planning and reasoning or from difficulty interpreting the interface itself.

Text-adventure games provide a natural testbed for evaluating baseline agentic capabilities. Observations and feedback are expressed in natural language, and agents interact with the environment through short action phrases. This removes intermediate representations between the model and the task, isolating agentic capability from interface familiarity.
In addition, these text simulators are lightweight compared to alternative benchmarks that require containerized systems or browser-based execution, enabling scalable evaluation. 

However, existing benchmarks that have adopted text-adventure games as part of their evaluation set still obscure true model performance. Prior work often introduces scaffolding that either leaks privileged information into prompts (e.g., BALROG~\citep{paglieri2024balrogbenchmarkingagenticllm}, verl-agent~\citep{feng2025group}) or modifies tasks to improve tractability (e.g., AgentBoard~\citep{Chang2024}).
This practice has two consequences.
First, it inflates reported scores, with similar environments yielding inconsistent results across benchmarks.
Second, it imposes an artificial ceiling on task difficulty, compressing differences between models and masking capability gaps (e.g., privileged information can lower task difficulty to the point that weaker models seemingly approach frontier performance).

To address this gap, we introduce \ours, a unified benchmark spanning \textworld (\abbrtextworld), \twx (\abbrtwx), \alfworld (\abbralfworld), \scienceworld (\abbrscienceworld) and \jericho in their canonical forms, a total of \numgames tasks evaluated through a single minimal harness applied uniformly across all frameworks, following Evidence-Centered Benchmark Design (ECBD) principles~\citep{liu2024ecbd}.
These frameworks form a calibrated progression of difficulty from simple synthetic tasks (i.e., \abbrtextworld's 10 games solvable in an average of 13.7 steps) to long-horizon, human-authored games (i.e., \jericho, a collection of 54 human-authored games requiring 87.2 steps on average).
We also introduce \simon, a simple instruction-following game that serves as a minimal competency check for the suite. 

We evaluate \nummodels model configurations spanning open- and closed-weight LLMs, producing a dataset of \numtranscripts trajectories.\footnote{This is more than \nummodels{} configurations $\times$ 5 seeds $\times$ \numgames{} tasks $=$ 46,360 because we also release transcripts for models that were not evaluated on the entirety of \ours or across all 5 seeds, including the 500-step \jericho runs of \astra and \opusfivefive discussed in Section~\ref{sec:frontier}.} To characterize agent behavior, we propose a framework for identifying baseline agentic competency failures.

\begin{figure*}[t]
\centering
 \includegraphics[width=\textwidth]{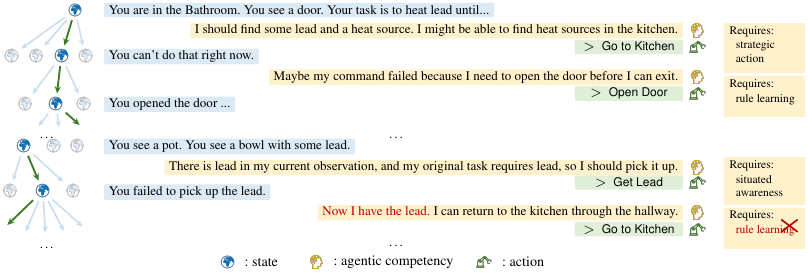}
\caption{Example of a gameplay trajectory presenting the conversation between the game engine and an agent. We additionally fabricate the internal thinking of the agent to demonstrate the agentic competency at each step, detailed in Section~\ref{sec:agent}. 
Here, the agent made a rule learning \textcolor{darkred}{mistake} in failing to account for the environment feedback, which further caused the generation of a sub-optimal action. 
}
    \label{fig:grounded_reasoning_failure}
\end{figure*}

Through the lens of the Problem-Space Hypothesis for complex problem solving \citep{newell1979reasoning}, we identify three atomic agentic competencies for an LLM operating as an agent: \textbf{Rule Learning} (inferring environment dynamics from feedback), \textbf{Situated Awareness} (maintaining an accurate representation of the current state), and \textbf{Strategic Action} (selecting goal-directed actions under constraints). This framework enables more precise analysis of failures beyond aggregate performance. Figure \ref{fig:grounded_reasoning_failure} illustrates a mock trajectory labeled with where these competencies come into play.

Across this benchmark, we find that while modern LLMs perform strongly on synthetic environments, they struggle on human-authored games. Even the strongest agents achieve less than 25\% of the available score in these settings, highlighting a significant gap between current capabilities and the demands of long-horizon, grounded reasoning.

In summary, our contributions are as follows:
\begin{itemize}[itemsep=0pt, topsep=0pt, partopsep=0pt, leftmargin=*]
    \item We introduce \ours, a unified benchmark for evaluating an LLM's baseline agentic competencies in text-adventure environments without privileged information or task modifications.
    \item We provide a standardized evaluation harness across \numgames tasks and define an evaluation path starting with \simon, a simple instruction-following litmus test.
    \item We release a dataset of \numtranscripts trajectories across \nummodels model configurations evaluated on \ours over 5 seeds for large-scale analysis of agent behavior.\footnote{\url{https://huggingface.co/datasets/ccui46/tale_suite_trajectories}}
    \item We provide automated failure detection for scalable analysis of Rule Learning and Situated Awareness failures. 
    \item We empirically show that current LLM agents perform well on synthetic environments but fail to scale to long-horizon, human-authored tasks.
\end{itemize}

\section{\ours}\label{sec:tales}
All frameworks included in \ours are \textit{text-adventure game} environments where players are provided a textual observation, and sometimes an explicit goal, and are able to interact with the environment through short action phrases. 
If these action phrases are invalid, the parser will typically return an error message indicative of whether the action has been understood by the environment but cannot be done, or if the parser just does not understand the action.
Table \ref{tab:benchmark_table} contrasts properties for each framework.
The following section briefly describes each environment and any notable characteristics about the world dynamics or rewards, with further detail in Appendix~\ref{app:framework_details}.
We discuss frameworks in the order of recommended progression such that users do not burn compute testing models against environments too challenging for them. 
\ours is built on the standard Gymnasium interface and allows for integration of any Gymnasium-compatible environment.\footnote{Any change to the task set or scoring will increment the \ours major version.}
We discuss predicting which models benefit from a longer step horizon in Appendix \ref{app:100steps}.
We do not use all environments within each framework. See Appendix \ref{app:env_subselection} for details.

\begin{table*}[t]
\small
\centering
\resizebox{\textwidth}{!}{%
   \begin{tabular}{c|cccc|c} 
    \toprule
    & \textworld & \twx & \alfworld & \scienceworld & \jericho \\
    \midrule
    Properties & (\abbrtextworld) & (\abbrtwx) & (\abbralfworld) & (\abbrscienceworld) & (\jericho) \\
    \midrule
    \# Games & 10 & 16 & 12 & 30 & 54 \\
    Avg. walkthrough steps & 13.70 & 33.06 & 5.83 & 41.67 & 87.15 \\
    Informative feedback & \textcolor{green}{\ding{51}} & \textcolor{green}{\ding{51}} & \textcolor{red}{\ding{55}} & \textcolor{green}{\ding{51}} & \textcolor{green}{\ding{51}} \\
    Intermediate rewards  & \textcolor{green}{\ding{51}} & \textcolor{green}{\ding{51}} & \textcolor{red}{\ding{55}} & \textcolor{green}{\ding{51}} & \textcolor{green}{\ding{51}} \\
    Nearest-neighbor parser  & \textcolor{green}{\ding{51}} & \textcolor{red}{\ding{55}} & \textcolor{red}{\ding{55}} & \textcolor{green}{\ding{51}} & \textcolor{green}{\ding{51}} \\
    Dead states & \textcolor{green}{\ding{51}} & \textcolor{green}{\ding{51}} & \textcolor{red}{\ding{55}} & \textcolor{green}{\ding{51}} & \textcolor{green}{\ding{51}} \\
    \bottomrule

    \end{tabular}%
}
\caption{
Key properties of each framework. 
Informative feedback: when an action fails, specific details about the failure's nature are returned.
Nearest-neighbor parser: environment can understand similar action phrases to mean the same thing.
Dead states: actions can prevent completing the task until a reset.
All pre-\jericho frameworks are synthetic: created for training embodied AI agents.
}
\label{tab:benchmark_table}
\end{table*}

\textbf{\simon: stop here if you cannot solve this task.}
With the release of \ours, we also introduce a new \abbrtwx game in the form of ``\simon'' to serve as a diagnostic entry point.
The basic \simon simply provides the agent an action to repeat at each turn while \simonwm provides a list of actions to follow at the start of the game.
Both versions award a point for every correct action. 
The game restarts if any action is performed out of order or is wrong.  
We find it serves as a good heuristic for an LLM's baseline competency with a Pearson correlation coefficient of 0.79 when taken with respect to the average of all \simon games against the entirety of \ours. 
A prerequisite to success in \ours is the ability to at least properly attend to information over long horizons.
\simon is the simplest form of this, posed in a straightforward, instruction-following task.
We include a graph visualizing the success correlation in Appendix \ref{simonsaysvtales}.

\textworld (\abbrtextworld) \citep{cote18textworld} is a framework originally designed for training agents with Reinforcement Learning (RL) on text-based games. It can generate synthetic text-adventure games of varying complexity. In \ours, we integrate the ``CookingWorld'' games that were used as part of the NeurIPS 2018 Competition\footnote{\url{https://competitions.codalab.org/competitions/21557}}. 

\twx (\abbrtwx) \citep{jansen2022textworldexpress} is a highly optimized re-implementation of many \abbrtextworld game scenarios that runs approximately three orders of magnitude faster than its \abbrtextworld counterparts. 
We opt to use \abbrtwx over \abbrtextworld for the performance improvement where applicable. 
While significantly faster, one potential drawback of using \abbrtwx over \abbrtextworld is also in its stricter parser. 
\abbrtwx simplifies its parser for speed and thus does not allow for nearest-neighbor action phrases.
In \abbrtwx, the feedback from \cmd{help} only reiterates the game objective.

\alfworld (\abbralfworld) \citep{ALFWorld20} is a multi-modal framework, combining complementary visual and textual observations, where agents are asked to navigate and perform tasks in a household setting. All tasks provide only a terminal reward of 1 upon task completion.
For \ours, we only use its textual modality as it has become the standard in the LLM literature when evaluated on \abbralfworld\citep{yao2023react, Shinn2023ReflexionAA}. The \abbralfworld environments are unique in their lack of informative feedback.

\scienceworld (\abbrscienceworld) \citep{scienceworld2022} is a framework focused on the completion of elementary-level science curriculum tasks. 
Notably for many of its tasks, \abbrscienceworld emulates an open-world setting where the player can complete the task in different ways that do not follow one expected trajectory. 
When it comes to heating objects, this part of the task can be completed by either the oven in the kitchen or the blast furnace in the workshop. 
Similarly, \abbrscienceworld also allows the player the freedom to reset the game on command.
This is especially important as a number of \abbrscienceworld games have failure modes where it is no longer possible to complete the assigned task in that playthrough.

\jericho \citep{hausknecht2020interactive} is a suite of 54 human-written, interactive fiction games.\footnote{
We exclude \textsc{hollywood.z3} because of segfault errors and \textsc{theatre.z5} due to game engine errors.
} 
We consider \jericho to be the most difficult framework due to the length and complexity of many of the games. Some can be completed within 17 steps while some others require over 500.
These games also cover an exceptionally broad range of genres and styles and lack the consistency of many other text-game environment suites designed for evaluating agents.
For example, \textsc{9:05} follows the morning of an ordinary office worker while \textsc{Anchorhead} is a Lovecraftian horror story. See Appendix \ref{app:walkthroughs} for maximum scores achievable at different step budgets.

\section{Operating as an Agent}\label{sec:agent}
The Problem-Space Hypothesis frames problem solving as navigation through a structured space defined by states, operators (actions), and the constraints governing which operators are applicable where \citep{newell1979reasoning}.
Classical formulations assume this space is given to an algorithmic solver that can effectively approach the space as a planning problem.
An LLM operating as an agent in a grounded setting can face a harder version of this setting where the set of operators may not be provided in full scope in advance, the constraints on those operators must be inferred from feedback, and the current state must be reconstructed from a message history of textual observations.
This difficulty is most evident in \jericho, where even the top-performing model fails to breach 25\% of the max possible score within 100 steps.

The gap between the classical setting and what an LLM faces when operating as an agent results in three minimal competencies we argue an LLM must display to operate as an agent on the most basic level.
At a minimum, an agent must be able to acquire knowledge of available operators and their constraints, maintain an accurate representation of the current state, and select operators that progress towards the goal state under finite resources. 
We name these competencies \textbf{Rule Learning}, \textbf{Situated Awareness}, and \textbf{Strategic Action}.
While these competencies do not explicitly enumerate all attributes that may be valuable in an agent, they are sufficient as a lens to perform a finer-grained analysis on LLM agent failure modes that aren't exposed through the \ours score alone.

\textbf{Rule Learning:} The ability to draw conclusions through interaction and observation.
Given the diversity of tasks, the environment's affordances may be unknown or contradict general expectations (e.g., a software interface element behaving inconsistently across operating systems).
A successful agent must be able to infer undocumented environmental constraints and discover available operators (actions) through feedback.
This competency encompasses both adjustments to API calls to adhere to strict function signatures and learning from system feedback~\citep{zhong2024learningfromlanguagefeedback} and is analogous to inductive reasoning in humans~\citep{heit2000properties}.

\textbf{Situated Awareness:} The ability to make decisions based on relevant information and current context.
An LLM with situated awareness can accurately distinguish the current state from redundant information in its context.
Agents are required to operate in discrete spaces as they move between initial, intermediate, and goal states within the problem space.
These transitions form a causality chain as the agent makes a decision at the current state to move to the next state.
The ability to accurately distinguish what information is relevant to the current state versus what information is a non-relevant artifact from past states becomes critical as the number of decisions the agent makes increases and the amount of contextually irrelevant information accumulates.

\textbf{Strategic Action:} The ability to choose between available operators based on logical application of general principles in a manner that progresses towards the goal under finite resources.
This competency is analogous to deductive reasoning in humans~\citep{johnson1999deductive}. 
It is infeasible to allow an LLM to run forever to complete a task.
While this may be possible in simulated environments where the cost of failure is incurred primarily through inference, real-world tasks often have a limited allocation of resources.
In such scenarios, the LLM must be able to leverage pre-existing knowledge to decide how best to use the available resources to progress most directly towards the goal.

The three competencies are designed to be atomic properties of the agent-environment loop rather than properties of any interface.
Rule Learning failures manifest as repeating a command that has already errored or reproducing the same syntax or formatting mistake across attempts.
Situated Awareness failures manifest as acting on state the agent itself has invalidated, for example reading a file it just deleted or pattern matching against code it has already modified.
Strategic Action failures manifest as long-horizon breakdowns in multi-step changes, where the agent makes locally valid edits while losing track of the overall goal.
These competencies aren't specific to text games. The same kinds of failures occur when agents interact with websites or embodied environments. For instance, a web agent can lose Situated Awareness when it fails to account for how its own actions have changed the page, while \alfworld shows how the same concepts extend to embodied tasks through a textual interface.
Our contribution is measuring these competencies in a setting where structured-interface formatting cannot mask or be conflated with them.

\section{Scalable Error Analysis}\label{sec:error_analysis}
With \ours, we also release \numtranscripts logs from LLM agents attempting to complete the gauntlet of frameworks from \abbrtextworld to \jericho alongside two automated detectors that target the described failure modes for Rule Learning and Situated Awareness.
By leveraging privileged information exposed by the environments, we construct reliable detectors for Rule Learning and Situated Awareness failures.

\textbf{Strategic Action failures cannot be cleanly distinguished from legitimate affordance exploration.} 
At first glance, a Strategic Action failure might be classified as a step that consumes resources without advancing the agent toward its goal. In an unfamiliar environment, however, an agent must spend some early steps mapping the available actions and environment dynamics.
This can appear structurally identical to burning steps without progress. 
Whether a sequence of actions that fails to make progress toward the goal constitutes a Strategic Action failure depends on what the agent already knows. Because that knowledge is partly determined by pre-training and cannot be directly audited, such failures are difficult to identify objectively. In contrast, Rule Learning and Situated Awareness failures can be more cleanly defined and detected from the interaction trajectory itself.

\textbf{Rule Learning Failure Detection.}
Rule Learning failures can be characterized by the agent attempting to repeat the same action in a state where an identical attempt had just failed.
To minimize false positives, a Rule Learning failure is only assigned when the agent attempts an action, receives explicit negative feedback from the environment, and attempts the exact same action phrase again over the course of multiple cycles.
Notably, this uses exact action phrase matching to exclude the structured attempts where an LLM agent slightly adjusts its phrasing to have the parser accept the command.
We detect Rule Learning errors by taking two passes through the transcript.
Pass one records trivial failures to adapt to parser error messages where an action is recorded as a Rule Learning failure if the environment responds with an error message and the agent continues to repeat the action despite consistent error feedback from the environment.
Pass two detects behavioral cycles where an action is recorded as a Rule Learning failure if the agent's behavior has settled into a verbatim repeating loop despite repeated feedback from the environment.

\textbf{Situated Awareness Failure Detection.}
Situated Awareness failures can be cleanly classified as the agent attempting to interact with an entity that does not exist within its current scope.
To do this, we replay the agent's trajectory through the environments while extracting ground-truth entity relation information such as agent location, inventory, entity location, and the set of rooms visited so far.
This allows us to clearly categorize Situated Awareness failures into three bins by comparing against the world state at the turn.
Let $X$ be the entity referenced by the agent.
We classify the failure as a hallucination if $X$ exists nowhere in the game's object table and has never appeared in any prior environment response; as a stale reference if $X$ appeared in an earlier environment response but no longer exists in the world state; and as a scope violation if $X$ exists at a previously visited location but the agent is no longer there and $X$ is not in its inventory.

\section{Evaluation} \label{ref:zero-shot}

\ours enables evaluation by customizing models with prompts and agentic strategies. For the initial release, we evaluate models using a minimal agent scaffolding with only the following prompt.

\begin{tcolorbox}[fontupper=\footnotesize\sffamily, left=-0.5pt, right=-0.5pt, top=0pt, bottom=0pt]\label{ref:zero-shot_prompt}
You are playing a text-based game and your goal is to finish it with the highest score. %
Upon reading the text observation, provide a *single* short phrase to interact with the game, e.g. \`{}get lamp\`{} (without backticks). %
When stuck, try using the \`{}help\`{} command to see what commands are available.
\end{tcolorbox}\label{ref:zero-shot prompt}

For the main results, we do not provide any other instructions to the LLMs. We aim to measure LLMs' raw capabilities without additional inductive bias from a human expert with domain knowledge.
In Section~\ref{sec:scaffolds}, we explore the effects of providing different forms of privileged or domain knowledge directly to the agent.
When calling the LLMs, the observation and feedback are provided as the \emph{user} inputs while the LLM actions are recorded as the \emph{assistant} outputs.
 
For the initial release of \ours, we cap the agent's max steps for all environments to 100 as this allows ample time for exploration in easier frameworks while providing sufficient signal to indicate which models deserve a higher step budget in \jericho.
In most frameworks, 100 steps serves as ample opportunity to make mistakes, self-correct, and eventually find success through focused exploration without allowing for a randomly acting agent to eventually happen upon success through chance.
In Section \ref{sec:quant} we discuss how automated metrics from 100-step performance in pre-\jericho frameworks can be used as a predictor for which models can effectively leverage a larger step budget in \jericho and provide results from allocating select top performing models at the time of the experiment a step budget of 300. See Appendix \ref{app:1000zork} for a case study of top performing models at the time of the experiment given a budget of 1000 steps in \zork.

\ours captures the model's capability evidence by the score from each game environment, ranging from 0--100. Although each game environment has its own customized scoring rules, those rules mark significant milestones in solving the game. 
Every task carries equal weight in the \ours score, so each framework counts in proportion to its number of tasks: as in an exam, the 68 pre-\jericho tasks make up most of the available points, while the remaining points, concentrated in \jericho, cost far more effort per point.
We do not normalize \jericho scores by what is achievable within 100 steps, so that scores stay comparable across step budgets (e.g., 100 versus 500 steps), and because a walkthrough's score at a given step is only a lower bound on what is achievable by then; Table~\ref{ref:score_table} reports it as a reference instead.
\ours can also replay a released trajectory through the environment and hand control back to the model at its final step, extending a run to a larger budget without re-running the steps already taken.
When supported by the LLM, we set the temperature to 0, provide fixed random seeds, and either cap reasoning at 1024 tokens (rows labeled (1024)) or set a reasoning effort (Appendix~\ref{app:command}).\footnote{We run 5 times for stability at 100 steps, though 3 runs may suffice if budget is limited.}

\begin{table*}[t]
    \caption{Average scores per framework and total \ours score for the top 10 models. \ours score is calculated by averaging success over all tasks. $^\ast$Submits each action through an action tool (Appendix \ref{app:tool_control}). Bold marks the best 5-seed score among the top 10 models in each column. $^\dagger$One seed. RL + SA: Rule Learning and Situated Awareness failures (Section~\ref{sec:error_analysis}).}
    \footnotesize
    \centering
    \renewcommand{\arraystretch}{1.2}
    \resizebox{\textwidth}{!}{
        \begin{tabular}{l c c c c c c c}
            \toprule
            \textbf{Model}
            & \textworld
            & \twx
            & \alfworld
            & \scienceworld
            & \jericho
            & \textbf{\ours Score}
            & \shortstack{\textbf{Avg. RL + SA Failures}\\\textbf{per Game}} \\
            \midrule
claude-fable-5 (thinking)$^{\ast\dagger}$
    & 100.0 & 100.0 & 100.0
    & 93.3 & 22.8 & 64.2 & 4.4 \\

claude-opus-5 (thinking)$^\ast$
    & 100.0 & 96.8 & 100.0
    & 95.0 & \textbf{22.2} & \textbf{63.9} & 5.5 \\

claude-opus-4.7 (thinking)$^\ast$
    & 100.0 & 96.8 & 100.0
    & \textbf{96.3} & 19.0 & 62.8 & \textbf{5.3} \\

gpt-5.4 (thinking)
    & 100.0 & 94.7 & 100.0
    & 95.2 & 16.9 & 61.3 & 5.9 \\

claude-opus-4.6 (thinking)$^\ast$
    & 100.0 & \textbf{98.6} & 100.0
    & 91.0 & 17.8 & 61.2 & 8.0 \\

claude-opus-4.8 (thinking)$^\ast$
    & 100.0 & 97.4 & 100.0
    & 91.6 & 17.3 & 61.0 & 6.7 \\

claude-opus-4.5 (thinking)
    & 100.0 & 89.1 & 100.0
    & 94.0 & 16.8 & 60.3 & 7.7 \\

o3 (medium)
    & 100.0 & 91.9 & 88.3
    & 93.0 & 15.7 & 58.7 & 9.5 \\

gpt-5.6-terra (thinking)
    & 98.7 & 81.9 & 100.0
    & 94.6 & 14.5 & 58.4 & 7.3 \\

gpt-5.1 (thinking)
    & 99.5 & 89.8 & 91.7
    & 90.2 & 16.1 & 58.2 & 6.9 \\

            \midrule

\astra{}
    & \textemdash & \textemdash & \textemdash
    & \textemdash & 24.9 & \textemdash & \textemdash \\

\opusfivefive{}
    & \textemdash & \textemdash & \textemdash
    & \textemdash & 20.0 & \textemdash & \textemdash \\

            \midrule

\textit{Walkthrough score at 100 steps}
    & 100.0 & 100.0 & 100.0
    & 100.0 & 58.2 & 81.5 & \textemdash \\

            \bottomrule
        \end{tabular}
    }
    \label{ref:score_table}
\end{table*}

Table \ref{ref:score_table} shows the per-framework scores of the top 10 overall-scoring models along with scores over \jericho for the current top frontier models. 
See Appendix \ref{app:all_scores} for all model configurations and scores and Appendices \ref{app:tw_all_scores}--\ref{app:jer_all_scores} for a per-model-task breakdown.
For the absolute best models at the time of writing, \astra and \opusfivefive, we skip early evaluation on synthetic environments and run them directly on \jericho with max reasoning effort up to 500 steps.
The score at 100 steps is shown in Table \ref{ref:score_table}.
We discuss the performance of \astra and \opusfivefive further in Section \ref{sec:frontier}.

The automatic failure detectors from Section \ref{sec:error_analysis} track the leaderboard closely. 
The top models typically average an extremely low number of Rule Learning or Situated Awareness failures per game, 4.4 to 9.5 compared to a median of 32 over all evaluated configurations.
At the bottom of the leaderboard, nearly every trajectory contains at least one such failure with the rate rising almost monotonically with \ours score ranking across all \nummodels configurations (Spearman $\rho = 0.87$).

\section{Analysis}\label{sec:quant}
In this section, we provide insights from a qualitative analysis of strong LLMs' game transcripts, focusing on \jericho as the framework where current frontier models still struggle.
While earlier generations of models struggled broadly across \abbralfworld, \abbrscienceworld, and \jericho, the current leaderboard shows substantial saturation in \abbralfworld and a tight cluster of top scores in \abbrscienceworld.
\jericho remains far from solved: even the strongest models achieve low normalized scores, frequently exhaust the 100-step budget, and must integrate exploration, state tracking, parser adaptation, and long-horizon decision making.
We therefore use \jericho to examine whether the core agentic failure modes persist for both modern non-thinking models and models run with explicit thinking budgets.

We compare the strongest non-thinking models in our leaderboard (Claude-4-Sonnet, Claude-3.7-Sonnet without thinking, Claude-3.5-Sonnet, GPT-4.1, and Claude-Sonnet-4.5) against the strongest thinking models at the time of the experiment (Claude-Opus-4.6 high, Claude-Opus-4.5 high, o3 medium, GPT-5.1 high, and o3 high).
For each group, we analyze the canonical 100-step \jericho trajectories over 54 games and 5 seeds per game.
Since most current closed-weight thinking models do not expose their internal thinking traces, we study thinking through observable behavior: late score progress, final-quarter novelty, and whether novel actions continue to produce new environment states.
Claude-3.7-Sonnet is an exception, and we analyze its thinking traces in Appendix~\ref{app:claude_thinking}.

\textbf{Strategic Action remains the central bottleneck.}
Across \jericho, both thinking and non-thinking models usually run out of steps before completing the game.
Non-thinking models reach the 100-step cap while incomplete in 99.2\% of trajectories, and thinking models do so in 96.4\%.
Despite 7 of 54 games being relatively easier and completable in 50 steps, even the best models fail to consistently complete these games.
By the end of many trajectories, the model has exhausted its obvious action repertoire and entered low-productivity search.
For example, Claude-3.5-Sonnet (non-thinking) reaches a score of 40 by step 71 in \zork and then spends the final 30 actions moving through already visited regions and retrying unproductive interactions.
This interpretation is reinforced by the extended step experiments done for the step-budget analysis.
Most exhausted \jericho trajectories are already stalled near the cutoff, so additional steps would usually extend low-productivity search rather than reveal hidden competence. 
The main exception is Claude-Opus-4.6, whose sustained late-state novelty and gains under longer budgets suggest that the strongest thinking models are beginning to make the step cap a real constraint.

\textbf{Most LLM agents learn rules locally but not reliably over the trajectory.}
The best agents often adapt to immediate parser feedback, but this local adaptation does not consistently become a durable strategy.
Non-thinking models receive parser-error feedback on 9.7\% of \jericho actions, compared with 6.9\% for thinking models, indicating that thinking models are slightly better at avoiding invalid command forms and using environmental affordances.
However, both groups still enter repeated action patterns: adjacent action cycles occur in 52.4\% of non-thinking trajectories and 51.2\% of thinking trajectories.
Models can recognize that a command failed while still failing to infer the broader puzzle rule, abandon a bad plan, or transfer a lesson across resets or long horizons.

\textbf{Reasoning alleviates late-horizon exploration failure but does not eliminate it.}
Thinking models achieve a higher average \jericho score than non-thinking models (16.2 vs. 10.0) and complete more games (3.6\% vs. 0.4\%), but both completion rates remain low.
The clearest difference appears late in the trajectory.
Thinking models make score progress over the final 25 actions in 33.6\% of runs, compared with 19.5\% for non-thinking models.
They also maintain higher final-quarter observation novelty (52.8\% vs. 37.9\%) and higher novel-action yield, meaning first-time actions are more likely to produce new environment states without parser-error feedback (76.5\% vs. 65.8\%).
In matched game-seed comparisons, thinking models exceed non-thinking models by 14.7 percentage points in final-quarter observation novelty and 10.7 percentage points in novel-action yield.

\paragraph{Pre-\jericho signals strongly predict which models would benefit from a higher step budget.}
\jericho's average walkthrough length of 87.15 steps approaches the 100-step limit used in our initial evaluation of \nummodels model configurations.
A natural question is whether this acts as an artificial ceiling on performance.
To investigate this, we ran seven models on \ours with extended step budgets over one seed and compared max normalized score achieved with a 100-step budget versus a 300-step budget. 
For a number of top-performing models, the extra 200 steps produced gains of up to 16.4\% in \jericho. 
We find a number of non-\jericho trajectory markers strongly indicate potential for score gain past 100 steps.
Aggregating across trajectory-shape metrics computed on non-\jericho runs can predict improvements in \jericho beyond 100 steps with Spearman rank correlations up to $\rho = 0.93$, where $\rho$ measures whether models are ordered similarly by the predictor and by later gain. 
Rule Learning failures predict in the opposite direction ($\rho = -0.82$): models that repeat failed actions or enter loops tend not to convert extra steps into progress. 
We discuss metrics in greater detail and minimal gains in non-\jericho frameworks in Appendix \ref{app:100steps}.

\section{Information as Agentic Scaffolds}\label{sec:scaffolds}
Liu et al.~\citep{liu2024ecbd} define an adaptation module as ``how test conditions are constructed'', noting a need for fairness in evaluation conditions across all objects of evaluation.
Providing excessive information during training and evaluation can lead to reported improvements of a method being unreliable when compared to baselines due to the scaffolding this provides.
In \ours, we motivate a minimal adaptation module to evaluate the baseline agentic competencies of LLM agents in complex, long-horizon, situated environments with text-adventure games.
A number of other works explore individual frameworks within \ours, but all of these works, to our knowledge, introduce privileged information or domain knowledge through their implementation of the adaptation module~\citep{Chang2024, paglieri2024balrogbenchmarkingagenticllm, lu2025intelligent, feng2025group}.

\begin{table}[!tbh]
\centering
\caption{Zero-shot scores in \abbrtextworld for Qwen3-8B when different forms of commonly leaked privileged or domain knowledge are included in the prompt for \abbrtextworld.}
\label{tab:adapt}
\begin{tabular}{c|ccc|c}
\toprule
In Prompt
& \makecell{Admissible \\ Actions}
& \makecell{Action \\ Templates}
& \makecell{Env. \\ Dynamics}
& \makecell{\abbrtextworld \\ Score} \\
\midrule
BALROG       & \textcolor{red}{\ding{55}}   & \textcolor{green}{\ding{51}} & \textcolor{green}{\ding{51}} & 28\% \\
AgentBoard   & \textcolor{red}{\ding{55}}   & \textcolor{green}{\ding{51}} & \textcolor{red}{\ding{55}}   & 32\% \\
IGE~\citep{lu2025intelligent} & \textcolor{red}{\ding{55}}   & \textcolor{green}{\ding{51}} & \textcolor{green}{\ding{51}} & 28\% \\
verl-agent   & \textcolor{green}{\ding{51}} & \textcolor{red}{\ding{55}}   & \textcolor{red}{\ding{55}}   & 39\% \\
\ours        & \textcolor{red}{\ding{55}}   & \textcolor{red}{\ding{55}}   & \textcolor{red}{\ding{55}}   & 18\% \\
\bottomrule
\end{tabular}
\end{table}

\textbf{Privileged and Domain Information.} We define ``privileged information'' as knowledge the agent has no access to within the scope of the environment, and ``domain information'' as knowledge the agent would otherwise need to discover through exploration.
For example, the per-step \textit{Admissible Actions} are privileged information as they are explicitly generated and verified by the framework.
Within the scope of the environment itself, the model has no direct access to this set of commands.

While the action templates are accessible via the \cmd{help} command and environmental dynamics may be eventually discovered by the agent on its own, we argue this information should not be directly provided to the model at the start.
The need to discover domain information through focused exploration represents one of the most challenging aspects of text-adventure games. 
Synthesizing this information requires strong rule learning in parsing over game-play history and identifying the relationships between the agent, its actions, and feedback from the environment.

Table \ref{tab:adapt} illustrates common modifications to the adaptation module, all with a non-trivial impact on downstream performance.
Admissible Actions are the actions the agent can take at any step, directly queried from the environment; Action Templates are those otherwise provided by the \cmd{help} command; and Environment Dynamics refer to environment constraints such as ``You can only cook meals in the kitchen'' that the agent may otherwise need to discover on its own.
Each is problematic when presented as implicit knowledge the model is provided at the start rather than effective modifications to the environment via the adaptation module. 
Even the contents of the help action can almost double a model's score in the zero-shot setting.
We investigate this further in Appendix~\ref{app:RL}, finding these uses of privileged information in RL can inflate success rate by up to 60\%.
Scores under our minimal harness measure the raw LLM; our intention is not to advocate for limitations on scaffolds so much as improved documentation when they are used.
\ours was designed with modularity such that agent designs and modifications, such as memory, planning, or retry, can easily be added and benchmarked, and our hope is that by establishing a standardized, minimal floor, the community will be more thorough in documenting any score-improving deviations from this minimalist baseline.

\section{At the Frontier}\label{sec:frontier}
To evaluate the current state of the frontier, we run \astra and \opusfivefive on all 54 \jericho games for 500 steps over 5 seeds, with maximum reasoning effort.
Within the first 100 steps, \astra reaches 24.9, the highest \jericho score in Table~\ref{ref:score_table}, and it climbs to 66.9 at 500 steps while \opusfivefive reaches only 57.0 at 500 steps.
Yet, even at 500 steps, \astra gets only two-thirds of the \jericho score while a walkthrough reaches 95.8 within the same budget. Also, \astra trails the walkthrough by 29 to 33 points at every budget from 100 to 500 steps.
The missing third is not a step-budget artifact: in 29 of the 54 games no \astra seed reaches the maximum score, although a walkthrough reaches it within 500 steps in 21 of them, and these games account for 76\% of \astra's gap to the walkthrough (74\% for \opusfivefive; per-game scores in Appendix~\ref{app:frontier_pergame}).

How the frontier gets this far matters for what \jericho now measures.
The \jericho games are heavily documented, and the frontier models certainly have substantial prior knowledge of them: on the first step of \zork, \astra's reasoning summary states ``I know the classic walkthrough'' (see also Appendix~\ref{app:data_contamination}).
What separates \astra from earlier models appears to be less this knowledge than something akin to planning recovery: where earlier models start a remembered sequence and fall into the low-productivity loops of Section~\ref{sec:quant} after the first death, reset, or unexpected event, \astra gets back on track and follows the sequence to its conclusion.
In its first 100 steps, \astra averages 7.1 Rule Learning and Situated Awareness failures per \jericho game and \opusfivefive 8.2, against 7.7 to 11.6 on \jericho for the models in Table~\ref{ref:score_table} (4.4 to 9.5 across all games).

Claude models released after claude-opus-4.6 (the other $^\ast$ rows in Table~\ref{ref:score_table}), when run without an action tool, frequently fail to terminate generation after the first command and instead continue with an imagined multi-turn transcript, including fabricated observations and invented endings that the model subsequently treats as fact.
In an audit of \jericho trajectories generated by claude-opus-4.8 without the tool, 478 of 1,100 responses were such continuations, and mean peak normalized score on matched cells fell from 38.9\% to 5.3\%.
\ours forwards the model's response to the game verbatim by design: single-command extraction or output repair is exactly the kind of bespoke, per-model scaffolding \ours exists to avoid.
We therefore evaluate these models with an action tool, which provides a standardized interface affordance rather than repairing their generated output. Re-running claude-opus-4.6 with the tool changes its scores minimally (Appendix~\ref{app:tool_control}).
\opusfivefive breaks this trend and plays with bare-text actions.

Results achieved with substantial agentic scaffolding, such as memory management and self-summarization, leave the contribution of the harness itself unquantified.
If such a system were to solve the \jericho games, the informative result would be its delta from the minimal-harness floor, which is only measurable because the floor exists; in that sense, harness-driven saturation is not a failure mode for \ours but its central use case.
Should model-level performance on the floor itself eventually saturate, we expect \jericho to retain the same role for the broader model distribution, as benchmarks like GSM8K have at the frontier.

\section{Related Work}
A large body of work exists in teaching agents to navigate and successfully complete text-based games. 
We specifically divide this section into RL-based agents, where the text-based game is defined as a Partially Observable Markov Decision Process (POMDP) \citep{kaelbling1998planning} and LLM-based agents, where information is fed to the LLM as an input with the output taken as an action.

\textbf{RL-based:}
Prior work explores text-based games as benchmarks for non-LLM-based agents \citep{narasimhan2015language, hausknecht2020interactive}.
Due to the intractable action space of language, prior RL approaches used action templates to reduce the possible command space to a subset learnable by an RL agent \citep{narasimhan2015language, Ammanabrolu2018PlayingTG, Yuan2018CountingTE,Hausknecht2019NAILAG, Ammanabrolu2020GraphCR, Ammanabrolu2020HowTA, Murugesan2021EfficientTR,Ryu2023AMA}.
These agents are often augmented with a knowledge graph for better state tracking representation or for direction \citep{Ammanabrolu2018PlayingTG, Hausknecht2019NAILAG, Ammanabrolu2020GraphCR, Murugesan2021EfficientTR, peng2023story, cui2023thespian, cui2024mixture}.
Other approaches still retain a base RL agent while using an LLM to guide it or generate diverse environments for generalization \citep{Yao2020KeepCA, Basavatia2024STARLINGST, golchha2024languageguidedexplorationrl}. 

\textbf{LLM-based:}
Early results showed that even state-of-the-art LLMs face difficulty when playing text-adventure games meant for human players \citep{Tsai2023CanLL}.
Previous work explored leveraging an external buffer or knowledge base to guide the agent \citep{Shinn2023ReflexionAA, zhu2024knowagent}.
Other approaches decompose tasks into simpler sub-goals \citep{Lin2023SwiftSageAG, prasad2024adaptasneededdecompositionplanning}.
Wang et al.~\citep{wang-etal-2024-soft} and Zhao et al.~\citep{zhao2024empowering} leverage LLMs to modify the provided action space.
Using contrasting trajectories with LLMs to improve performance has also been explored \citep{song2024trial, Yang2024ReActMA, qiao2025agent}.
\section{Conclusion}\label{sec:conclusion}
In this work, we introduce \ours, a unified benchmark for evaluating LLM agents in text-adventure game environments, releasing all code and \numtranscripts agent logs as an artifact for the community.
\ours spans synthetic and human-written games in their canonical forms, without scaffolds that obscure baseline agentic capability.
We introduce \simon as a simple instruction-following diagnostic for whether a model has the baseline competency to attempt \ours at all.
Across \nummodels model configurations, we find that current frontier LLMs perform strongly on synthetic games but still struggle as environments become longer, less regular, and more human-authored.
Even the top models fail to reach 25\% on \jericho within the main evaluation setting.

We identify three core competencies that are vital for LLM agents: Rule Learning, Situated Awareness, and Strategic Action.
By replaying trajectories against privileged simulator state, we detect Rule Learning and Situated Awareness failures without giving that information to the agent during evaluation.
These failures track model rank closely, while qualitative analysis shows that strong models still repeat trajectory-level mistakes, lose track of what is currently actionable, and spend scarce steps on unfocused exploration.

Finally, we justify the 100-step cap as an informative baseline while showing that pre-\jericho trajectory markers can identify which models benefit from a larger budget.
We also quantify the effect of privileged information in prompts, demonstrating how common scaffolds can inflate scores.
Overall, while progress has been made on synthetic text-adventure games, LLMs are still far from being agents competent enough to complete games designed primarily for human enjoyment.

\textbf{Limitations.}
The main \ours evaluation uses a 100-step cap.
This cap provides a focused measure of exploration and prevents easier environments from being solved by exhaustive retrying, but it is not the full horizon required by every game in \jericho.
Our automated detectors also only cover Rule Learning and Situated Awareness; Strategic Action remains difficult to distinguish from legitimate affordance discovery without assuming what the model already knows.

Our qualitative analysis is necessarily partial and may reflect author familiarity with the included frameworks.
Broader annotation, as well as validated judge-model annotation pipelines, would further strengthen these observations.
We also inherit data contamination concerns from \jericho.
Frontier models now recall walkthroughs of well-documented games (Section~\ref{sec:frontier}), so part of their \jericho score reflects executing a remembered plan rather than discovering one, and future work should include newly authored held-out games to better separate memorization from general agentic competence.
The aggregate \ours score depends on the mix of tasks in the suite, which is why we also report per-framework scores.
For games with commercial or restrictive licensing, we release only the agent actions, seeds, and scores, with regeneration instructions for anyone with access to the relevant story file.
Exact reproduction is also limited for closed-source or deprecated API models.

\section*{Acknowledgements and Disclosures}
Compute credits for this work were provided by Microsoft. The authors declare no competing interests.

\bibliographystyle{plain} %
\bibliography{neurips2026/neurips}

\medskip

\appendix
\newpage
\textbf{\large{Table of Contents:}}
\begin{itemize}
    \item In Appendix~\ref{app:all_scores}, we list all max normalized scores per model averaged over 5 seeds.
    \item In Appendix~\ref{app:walkthroughs}, we calculate the percent score reached by following the walkthrough across 50, 100, 200, 300, 400, 500, and 1000 steps in \jericho. 
    \item In Appendix~\ref{app:token_use}, we list tokens used per model per framework.
    \item In Appendix~\ref{app:all_stds}, we list the standard deviation of scores across seeds per model per framework.
    \item In Appendix~\ref{app:all_games}, we list all tasks across all frameworks.
    \item In Appendix~\ref{app:tw_all_scores}, we list average max normalized score per model for each task in \textworld.
    \item In Appendix~\ref{app:twx_all_scores}, we list average max normalized score per model for each task in \twx.
    \item In Appendix~\ref{app:alf_all_scores}, we list average max normalized score per model for each task in \alfworld.
    \item In Appendix~\ref{app:sw_all_scores}, we list average max normalized score per model for each task in \scienceworld.
    \item In Appendix~\ref{app:jer_all_scores}, we list average max normalized score per model for each task in \jericho.
    \item In Appendix~\ref{app:frontier_pergame}, we list the per-game \jericho scores of \astra and \opusfivefive at 500 steps.
    \item In Appendix~\ref{app:next_jericho}, we discuss how \ours can be extended beyond \jericho.
    \item In Appendix~\ref{simonsaysvtales}, we list average Simon Says scores versus overall \ours score.
    \item In Appendix~\ref{app:resources}, we describe the compute costs and machines used to run our experiments.
    \item In Appendix~\ref{app:claude_thinking}, we compare the thinking and non-thinking modes of Claude-3.7-Sonnet through its thinking traces.
    \item In Appendix~\ref{sec:labeled_subset}, we report counts of labeled reasoning failures for Claude models.
    \item In Appendix~\ref{app:1000zork}, we analyze 1000-step \zork runs for o3 and Claude-3.7-Sonnet.
    \item In Appendix~\ref{app:hyperparameters}, we describe hyperparameters for the RL fine-tuning used in Appendix~\ref{app:RL}.
    \item In Appendix~\ref{app:framework_details}, we give additional detail on each framework.
    \item In Appendix~\ref{app:env_subselection}, we describe how we subselected tasks for \ours from the original frameworks.
    \item In Appendix~\ref{app:annotated}, we describe our annotation criteria for the first stage of our qualitative analysis and provide examples.
    \item In Appendix~\ref{app:data_contamination}, we provide evidence from a Claude-3.7-Sonnet thinking trace of data leakage from \zork.
    \item In Appendix~\ref{app:command}, we give an example of a run command for benchmarking a model on \ours.
    \item In Appendix~\ref{app:100steps}, we describe additional metrics used to predict if models will benefit from a step budget larger than 100 in \jericho.
    \item In Appendix~\ref{app:RL}, we investigate the impact of leaking privileged information to the LLM during RL training.
    \item In Appendix~\ref{app:tool_control}, we compare claude-opus-4.6 (thinking) with and without the action tool.
\end{itemize}

\newpage

\section{All Agent Average Scores per Framework}\label{app:all_scores}
In Table \ref{tab:all_model_framework_scores}, we include the average scores per framework, the average per-game score, and the average RL + SA failures per game of all full-coverage LLM agents. It is the untruncated version of Table \ref{ref:score_table}, with the same selection and labels.

\begin{table}[ht]
    \centering
    \resizebox{\textwidth}{!}{
        \begin{tabular}{l c c c c c c c}
            \toprule
            \textbf{Model} & \textworld & \twx & \alfworld & \scienceworld & \jericho & \textbf{\ours Score} & \shortstack{\textbf{Avg. RL + SA Failures}\\\textbf{per Game}} \\
            \midrule
claude-fable-5 (thinking)$^{\ast\dagger}$                  & 100.0 & 100.0 & 100.0 & 93.3 & 22.8 & 64.2 & 4.4$^\S$ \\
claude-opus-5 (thinking)$^\ast$                            & 100.0 & 96.8 & 100.0 & 95.0 & 22.2 & 63.9 & 5.5 \\
claude-opus-4.7 (thinking)$^\ast$                          & 100.0 & 96.8 & 100.0 & 96.3 & 19.0 & 62.8 & 5.3 \\
gpt-5.4 (thinking)                                         & 100.0 & 94.7 & 100.0 & 95.2 & 16.9 & 61.3 & 5.9 \\
claude-opus-4.6 (thinking)$^\ast$                          & 100.0 & 98.6 & 100.0 & 91.0 & 17.8 & 61.2 & 8.0 \\
claude-opus-4.8 (thinking)$^\ast$                          & 100.0 & 97.4 & 100.0 & 91.6 & 17.3 & 61.0 & 6.7 \\
claude-opus-4.5 (thinking)                                 & 100.0 & 89.1 & 100.0 & 94.0 & 16.8 & 60.3 & 7.7 \\
o3 (medium)                                                & 100.0 & 91.9 & 88.3 & 93.0 & 15.7 & 58.7 & 9.5 \\
gpt-5.6-terra (thinking)                                   & 98.7 & 81.9 & 100.0 & 94.6 & 14.5 & 58.4 & 7.3 \\
gpt-5.1 (thinking)                                         & 99.5 & 89.8 & 91.7 & 90.2 & 16.1 & 58.2 & 6.9 \\
o3 (high)                                                  & 100.0 & 89.6 & 81.7 & 93.1 & 16.1 & 58.0 & 8.7 \\
claude-sonnet-4.6 (thinking)                               & 100.0 & 90.5 & 100.0 & 88.7 & 13.0 & 57.5 & 10.6 \\
gpt-5 (thinking)                                           & 100.0 & 75.5 & 93.3 & 91.8 & 17.2 & 57.5 & 7.5 \\
kimi-k2.6 (thinking)                                       & 100.0 & 88.5 & 96.7 & 86.9 & 14.4 & 57.1 & 7.5 \\
gpt-5.6-luna (thinking)                                    & 100.0 & 87.1 & 88.3 & 91.4 & 9.9 & 55.2 & 9.7 \\
o3 (low)                                                   & 99.1 & 89.8 & 70.0 & 88.3 & 14.2 & 54.8 & 12.0 \\
claude-4-sonnet                                            & 99.6 & 78.4 & 91.7 & 87.0 & 12.3 & 54.3 & 13.1 \\
kimi-k2.5 (thinking)                                       & 99.1 & 83.0 & 88.3 & 86.4 & 11.2 & 53.9 & 10.2 \\
claude-3.7-sonnet (1024)                                   & 97.3 & 91.4 & 83.3 & 76.5 & 12.5 & 52.5 & 11.2 \\
claude-3.7-sonnet                                          & 97.3 & 95.8 & 81.7 & 72.4 & 13.0 & 52.1 & 15.7 \\
claude-3.5-sonnet-latest                                   & 95.5 & 81.6 & 75.0 & 82.3 & 9.6 & 50.4 & 16.4 \\
gpt-4.1                                                    & 95.3 & 92.6 & 83.3 & 76.1 & 6.8 & 49.9 & 15.7 \\
grok-4.3 (thinking)                                        & 96.4 & 61.1 & 96.7 & 83.2 & 9.0 & 49.9 & 19.1 \\
Qwen3.5-122B-A10B (1024)                                   & 91.5 & 83.4 & 63.3 & 76.5 & 7.6 & 46.8 & 26.1 \\
Qwen3.5-27B (1024)                                         & 92.0 & 86.1 & 75.0 & 68.6 & 7.7 & 46.5 & 20.5 \\
gpt-5-mini (thinking)                                      & 94.7 & 61.9 & 61.7 & 82.7 & 9.5 & 46.5 & 15.6 \\
Qwen3.6-27B (1024)                                         & 94.0 & 89.5 & 70.0 & 68.5 & 6.2 & 45.9 & 26.5 \\
o1 (high)                                                  & 97.8 & 70.2 & 28.3 & 80.1 & 10.3 & 44.2 & 18.3 \\
claude-sonnet-4.5                                          & 100.0 & 71.4 & 83.3 & 56.7 & 8.6 & 43.5 & 13.2 \\
gpt-4o                                                     & 83.6 & 80.6 & 56.7 & 61.4 & 5.6 & 40.6 & 32.7 \\
claude-3.5-haiku                                           & 94.9 & 79.8 & 26.7 & 67.3 & 5.0 & 39.6 & 22.3 \\
Qwen3.6-35B-A3B (1024)                                     & 81.7 & 73.4 & 41.7 & 59.8 & 2.8 & 36.4 & 59.4 \\
Llama-3.1-405B-Instruct                                    & 90.9 & 79.2 & 31.7 & 51.8 & 6.1 & 36.4 & 19.6 \\
gemma-4-31b-it (1024)                                      & 86.4 & 57.6 & 60.0 & 46.7 & 6.9 & 35.1 & 132.6 \\
gemini-2.0-flash                                           & 80.8 & 76.1 & 20.0 & 57.1 & 5.4 & 35.0 & 32.1 \\
Qwen3-32B (1024)                                           & 79.5 & 68.9 & 48.3 & 49.8 & 4.0 & 34.3 & 50.4 \\
Llama-3.3-70B-Instruct                                     & 69.6 & 77.2 & 15.0 & 55.1 & 4.5 & 32.8 & 26.8 \\
Llama-3.1-70B-Instruct                                     & 65.6 & 81.9 & 8.3 & 51.9 & 5.3 & 32.0 & 32.0 \\
Qwen2.5-72B-Instruct                                       & 76.5 & 83.8 & 36.7 & 35.0 & 2.9 & 30.7 & 44.3 \\
Qwen3.5-9B (1024)                                          & 73.3 & 72.6 & 18.3 & 47.4 & 3.5 & 30.5 & 44.9 \\
Mistral-Large-Instruct-2407                                & 82.4 & 68.3 & 6.7 & 46.1 & 5.8 & 30.3 & 38.5 \\
o3-mini (high)                                             & 83.9 & 61.3 & 11.7 & 48.1 & 5.0 & 30.1 & 38.5 \\
MiniMax-M2.7 (1024)                                        & 64.0 & 30.4 & 21.7 & 55.3 & 6.4 & 27.8 & 67.5 \\
gpt-4.1-mini                                               & 62.1 & 74.5 & 5.0 & 41.9 & 3.4 & 27.1 & 44.5 \\
Llama-3.3-Nemotron-Super-49B-v1 (1024)                     & 65.8 & 78.9 & 16.7 & 26.5 & 2.8 & 25.1 & 20.7 \\
gpt-5.4-nano (thinking)                                    & 69.7 & 67.9 & 5.0 & 35.8 & 2.2 & 24.9 & 71.7 \\
Qwen3-30B-A3B (1024)                                       & 55.5 & 58.2 & 16.7 & 31.5 & 3.0 & 22.9 & 67.5 \\
gpt-4o-mini                                                & 56.5 & 73.6 & 0.0 & 27.2 & 1.8 & 21.8 & 57.4 \\
claude-haiku-4.5                                           & 35.5 & 54.3 & 0.0 & 30.5 & 5.4 & 19.9 & 17.1 \\
Llama-4-Scout-17B-16E-Instruct                             & 41.1 & 68.4 & 0.0 & 27.0 & 1.8 & 19.8 & 45.4 \\
Qwen3.5-4B (1024)                                          & 37.7 & 63.8 & 0.0 & 27.9 & 1.8 & 19.1 & 78.6 \\
gpt-5-nano (thinking)                                      & 50.1 & 41.3 & 1.7 & 32.0 & 1.7 & 18.3 & 46.8 \\
gpt-oss-20b (thinking)                                     & 60.6 & 45.1 & 0.0 & 25.7 & 1.6 & 17.9 & 54.0 \\
Llama-4-Maverick-17B-128E-Instruct-FP8                     & 43.5 & 56.1 & 8.3 & 11.5 & 2.0 & 15.5 & 11.6 \\
Mistral-Small-Instruct-2409                                & 56.1 & 27.3 & 0.0 & 24.4 & 1.4 & 14.8 & 87.3 \\
DeepSeek-V4-Flash (1024)                                   & 31.3 & 42.7 & 1.7 & 17.7 & 4.6 & 14.7 & 121.4 \\
Llama-3.1-8B-Instruct                                      & 29.7 & 50.3 & 0.0 & 15.7 & 2.3 & 13.9 & 81.4 \\
DeepSeek-R1 (1024)                                         & 37.1 & 38.6 & 0.0 & 15.8 & 1.0 & 12.4 & 21.4 \\
Qwen2.5-7B-Instruct                                        & 27.7 & 45.6 & 0.0 & 12.6 & 0.7 & 11.7 & 127.6 \\
Llama-3.2-3B-Instruct                                      & 21.4 & 42.0 & 0.0 & 10.0 & 1.5 & 10.4 & 82.3 \\
phi-4                                                      & 20.8 & 43.8 & 0.0 & 8.9 & 1.6 & 10.3 & 101.3 \\
gpt-4.1-nano                                               & 12.8 & 38.7 & 0.0 & 9.4 & 3.6 & 10.0 & 82.4 \\
Mistral-Small-24B-Instruct-2501                            & 15.8 & 23.0 & 0.0 & 15.8 & 1.4 & 8.8 & 28.3 \\
DeepSeek-R1-Distill-Llama-70B (1024)                       & 8.7 & 39.8 & 0.0 & 7.7 & 1.3 & 8.4 & 46.1 \\
Qwen3.5-2B (1024)                                          & 0.0 & 31.3 & 0.0 & 3.3 & 0.5 & 5.1 & 122.4 \\
Ministral-8B-Instruct-2410                                 & 10.9 & 22.8 & 0.0 & 2.3 & 0.4 & 4.6 & 131.6 \\
Mistral-Small-3.1-24B-Instruct-2503                        & 2.5 & 10.3 & 0.0 & 10.5 & 0.8 & 4.5 & 36.3 \\
Mixtral-8x22B-Instruct-v0.1                                & 17.1 & 8.4 & 0.0 & 4.0 & 0.4 & 3.7 & 144.1 \\
Llama-3.2-1B-Instruct                                      & 0.0 & 19.0 & 0.0 & 2.4 & 0.6 & 3.3 & 119.0 \\
Phi-3-mini-128k-instruct                                   & 2.7 & 9.4 & 0.0 & 2.4 & 0.3 & 2.2 & 98.0 \\
Phi-3.5-MoE-instruct                                       & 0.0 & 7.0 & 0.0 & 2.3 & 0.4 & 1.7 & 80.4 \\
Phi-4-mini-instruct                                        & 0.0 & 5.5 & 0.0 & 2.3 & 0.5 & 1.5 & 120.9 \\
Mixtral-8x7B-Instruct-v0.1                                 & 0.0 & 1.6 & 0.0 & 4.0 & 0.3 & 1.3 & 57.4 \\
Phi-3.5-mini-instruct                                      & 0.0 & 2.0 & 0.0 & 2.4 & 0.5 & 1.0 & 59.2 \\
Qwen3.5-0.8B (1024)                                        & 0.0 & 0.0 & 0.0 & 2.3 & 0.3 & 0.7 & 119.9 \\
Phi-3-medium-128k-instruct                                 & 0.0 & 0.0 & 0.0 & 2.3 & 0.3 & 0.7 & 152.9 \\
            \midrule
\textit{Walkthrough score at 100 steps}                     & 100.0 & 100.0 & 100.0 & 100.0 & 58.2 & 81.5 & \textemdash \\
            \bottomrule
        \end{tabular}
    }
    \caption{Average scores per framework, total \ours score and average RL + SA failures per game for every configuration with all 610 canonical trajectories, plus claude-fable-5 ($^\dagger$one seed, 122 trajectories). $^\ast$Submits each action through an action tool; where a model was run both with and without it, only the tool run is listed, and Appendix \ref{app:tool_control} compares the two for claude-opus-4.6. $^\S$Excludes two runs (\scienceworld Mendelian Genetics) in which the model returned no action at any step.}
    \label{tab:all_model_framework_scores}
\end{table}

\clearpage
\newpage
\section{\jericho Walkthrough Scores}
\label{app:walkthroughs}
Table \ref{tab:jericho_envs} shows the percent of achievable score when using the walkthrough for \jericho at different steps. 

\begin{table}[h!]
    \tiny
    \centering
    \renewcommand{\arraystretch}{0.9}
    \resizebox{\textwidth}{!}{
        \begin{tabular}{lccccccc}
            \toprule
            Game & 50 Steps & 100 Steps & 200 Steps & 300 Steps & 400 Steps & 500 Steps & 1000 Steps \\
            \midrule
            JerichoEnv905 & 100.000 & 100.000 & 100.000 & 100.000 & 100.000 & 100.000 & 100.000 \\
JerichoEnvAcorncourt & 100.000 & 100.000 & 100.000 & 100.000 & 100.000 & 100.000 & 100.000 \\
JerichoEnvAdvent & 26.300 & 42.600 & 63.100 & 100.000 & 100.000 & 100.000 & 100.000 \\
JerichoEnvAdventureland & 21.000 & 42.000 & 100.000 & 100.000 & 100.000 & 100.000 & 100.000 \\
JerichoEnvAfflicted & 46.700 & 100.000 & 100.000 & 100.000 & 100.000 & 100.000 & 100.000 \\
JerichoEnvAnchor & 5.000 & 11.000 & 29.000 & 41.000 & 52.000 & 64.000 & 99.000 \\
JerichoEnvAwaken & 60.000 & 100.000 & 100.000 & 100.000 & 100.000 & 100.000 & 100.000 \\
JerichoEnvBalances & 58.800 & 58.800 & 98.000 & 98.000 & 98.000 & 98.000 & 98.000 \\
JerichoEnvBallyhoo & 15.000 & 30.000 & 50.000 & 75.000 & 95.000 & 100.000 & 100.000 \\
JerichoEnvCurses & 3.800 & 5.600 & 12.700 & 28.200 & 38.200 & 47.500 & 81.800 \\
JerichoEnvCutthroat & 12.000 & 28.000 & 36.000 & 44.000 & 100.000 & 100.000 & 100.000 \\
JerichoEnvDeephome & 20.700 & 28.000 & 60.000 & 76.000 & 100.000 & 100.000 & 100.000 \\
JerichoEnvDetective & 100.000 & 100.000 & 100.000 & 100.000 & 100.000 & 100.000 & 100.000 \\
JerichoEnvDragon & 24.000 & 100.000 & 100.000 & 100.000 & 100.000 & 100.000 & 100.000 \\
JerichoEnvEnchanter & 11.300 & 31.200 & 70.000 & 100.000 & 100.000 & 100.000 & 100.000 \\
JerichoEnvEnter & 35.000 & 100.000 & 100.000 & 100.000 & 100.000 & 100.000 & 100.000 \\
JerichoEnvGold & 12.000 & 30.000 & 51.000 & 75.000 & 100.000 & 100.000 & 100.000 \\
JerichoEnvHhgg & 8.300 & 21.200 & 40.000 & 50.000 & 100.000 & 100.000 & 100.000 \\
JerichoEnvHuntdark & 0.000 & 100.000 & 100.000 & 100.000 & 100.000 & 100.000 & 100.000 \\
JerichoEnvInfidel & 12.500 & 20.000 & 70.000 & 100.000 & 100.000 & 100.000 & 100.000 \\
JerichoEnvInhumane & 33.300 & 77.800 & 100.000 & 100.000 & 100.000 & 100.000 & 100.000 \\
JerichoEnvJewel & 15.600 & 26.700 & 77.800 & 100.000 & 100.000 & 100.000 & 100.000 \\
JerichoEnvKarn & 5.900 & 23.500 & 38.200 & 67.600 & 100.000 & 100.000 & 100.000 \\
JerichoEnvLibrary & 100.000 & 100.000 & 100.000 & 100.000 & 100.000 & 100.000 & 100.000 \\
JerichoEnvLoose & 100.000 & 100.000 & 100.000 & 100.000 & 100.000 & 100.000 & 100.000 \\
JerichoEnvLostpig & 28.600 & 42.900 & 85.700 & 85.700 & 85.700 & 85.700 & 85.700 \\
JerichoEnvLudicorp & 13.300 & 25.300 & 58.700 & 92.700 & 100.000 & 100.000 & 100.000 \\
JerichoEnvLurking & 10.000 & 25.000 & 55.000 & 100.000 & 100.000 & 100.000 & 100.000 \\
JerichoEnvMoonlit & 0.000 & 100.000 & 100.000 & 100.000 & 100.000 & 100.000 & 100.000 \\
JerichoEnvMurdac & 6.800 & 18.000 & 18.000 & 48.000 & 99.600 & 99.600 & 99.600 \\
JerichoEnvNight & 60.000 & 100.000 & 100.000 & 100.000 & 100.000 & 100.000 & 100.000 \\
JerichoEnvOmniquest & 40.000 & 100.000 & 100.000 & 100.000 & 100.000 & 100.000 & 100.000 \\
JerichoEnvPartyfoul & 0.000 & 100.000 & 100.000 & 100.000 & 100.000 & 100.000 & 100.000 \\
JerichoEnvPentari & 100.000 & 100.000 & 100.000 & 100.000 & 100.000 & 100.000 & 100.000 \\
JerichoEnvPlanetfall & 7.500 & 26.300 & 35.000 & 60.000 & 100.000 & 100.000 & 100.000 \\
JerichoEnvPlundered & 16.000 & 44.000 & 100.000 & 100.000 & 100.000 & 100.000 & 100.000 \\
JerichoEnvReverb & 60.000 & 100.000 & 100.000 & 100.000 & 100.000 & 100.000 & 100.000 \\
JerichoEnvSeastalker & 28.000 & 44.000 & 90.000 & 100.000 & 100.000 & 100.000 & 100.000 \\
JerichoEnvSherlock & 23.000 & 37.000 & 55.000 & 84.000 & 100.000 & 100.000 & 100.000 \\
JerichoEnvSnacktime & 100.000 & 100.000 & 100.000 & 100.000 & 100.000 & 100.000 & 100.000 \\
JerichoEnvSorcerer & 23.700 & 37.500 & 53.700 & 100.000 & 100.000 & 100.000 & 100.000 \\
JerichoEnvSpellbrkr & 13.300 & 26.700 & 42.500 & 65.000 & 91.700 & 100.000 & 100.000 \\
JerichoEnvSpirit & 2.400 & 3.200 & 9.600 & 14.400 & 18.800 & 27.200 & 71.200 \\
JerichoEnvTemple & 28.600 & 57.100 & 100.000 & 100.000 & 100.000 & 100.000 & 100.000 \\
JerichoEnvTrinity & 15.000 & 22.000 & 32.000 & 47.000 & 58.000 & 78.000 & 100.000 \\
JerichoEnvTryst205 & 2.900 & 14.300 & 24.300 & 41.400 & 58.600 & 74.300 & 100.000 \\
JerichoEnvWeapon & 0.000 & 100.000 & 100.000 & 100.000 & 100.000 & 100.000 & 100.000 \\
JerichoEnvWishbringer & 24.000 & 50.000 & 100.000 & 100.000 & 100.000 & 100.000 & 100.000 \\
JerichoEnvYomomma & 25.700 & 97.100 & 97.100 & 97.100 & 97.100 & 97.100 & 97.100 \\
JerichoEnvZenon & 40.000 & 100.000 & 100.000 & 100.000 & 100.000 & 100.000 & 100.000 \\
JerichoEnvZork1 & 18.000 & 29.100 & 41.700 & 77.400 & 100.000 & 100.000 & 100.000 \\
JerichoEnvZork2 & 6.200 & 22.500 & 47.500 & 100.000 & 100.000 & 100.000 & 100.000 \\
JerichoEnvZork3 & 28.600 & 42.900 & 100.000 & 100.000 & 100.000 & 100.000 & 100.000 \\
JerichoEnvZtuu & 47.000 & 100.000 & 100.000 & 100.000 & 100.000 & 100.000 & 100.000 \\

            \bottomrule
        \end{tabular}
    }
    \caption{Max score percentage reached by following the provided walkthrough for each \jericho game.}
    \label{tab:jericho_envs}
\end{table}

\clearpage
\section{All Agent Average Final Tokens Used per Framework}\label{app:token_use}
In Tables \ref{tab:all_model_token_comparison} and \ref{tab:all_model_token_comparison_2}, we include the average final tokens used per game for each framework of every configuration in Table \ref{tab:all_model_framework_scores}.
\begin{table}[h!]
    \footnotesize
    \centering
    \renewcommand{\arraystretch}{1.2}
    \resizebox{\textwidth}{!}{
\begin{tabular}{l c c c c c }
            \toprule
            \textbf{Model} & \textworld & \twx & \alfworld & \scienceworld & \jericho  \\
            \midrule
claude-fable-5 (thinking)$^{\ast\dagger}$                  & 2152.7 & 1645.6 & 1074.8 & 2035.0 & 9298.0 \\
claude-opus-5 (thinking)$^\ast$                            & 2056.1 & 2526.9 & 1219.2 & 2698.2 & 9037.1 \\
claude-opus-4.7 (thinking)$^\ast$                          & 2416.1 & 2450.9 & 1382.1 & 2549.1 & 8679.0 \\
gpt-5.4 (thinking)                                         & 1553.1 & 2528.2 & 1731.1 & 2509.0 & 9557.3 \\
claude-opus-4.6 (thinking)$^\ast$                          & 2111.3 & 1550.5 & 1469.2 & 2358.2 & 6422.4 \\
claude-opus-4.8 (thinking)$^\ast$                          & 2210.7 & 2412.5 & 1343.0 & 2847.1 & 8739.5 \\
claude-opus-4.5 (thinking)                                 & 1843.5 & 1599.1 & 1300.2 & 1945.4 & 5875.0 \\
\bottomrule
        \end{tabular}
    }
    \caption{Avg final tokens used per LLM per game for each framework. Ordering and labels follow Table \ref{tab:all_model_framework_scores}.}
    \label{tab:all_model_token_comparison}
\end{table}
\begin{table}[h!]
    \footnotesize
    \centering
    \renewcommand{\arraystretch}{1.2}
    \resizebox{\textwidth}{!}{
        \begin{tabular}{l c c c c c }
            \toprule
            \textbf{Model} & \textworld & \twx & \alfworld & \scienceworld & \jericho  \\
            \midrule
Qwen2.5-72B-Instruct                                       & 2832.6 & 1094.5 & 1967.2 & 3484.5 & 3426.2 \\
Qwen3.5-9B (1024)                                          & 3835.4 & 3204.2 & 2275.9 & 3174.1 & 4966.2 \\
Mistral-Large-Instruct-2407                                & 2933.5 & 2211.6 & 2048.7 & 3273.4 & 4214.2 \\
o3-mini (high)                                             & 2785.8 & 4300.6 & 1853.2 & 2165.3 & 3395.8 \\
MiniMax-M2.7 (1024)                                        & 14566.4 & 38039.5 & 17035.5 & 8096.5 & 19121.8 \\
gpt-4.1-mini                                               & 3905.4 & 1891.9 & 2422.6 & 2426.5 & 3313.4 \\
Llama-3.3-Nemotron-Super-49B-v1 (1024)                     & 10614.7 & 21599.4 & 11203.8 & 38957.8 & 25245.2 \\
\bottomrule
        \end{tabular}
    }
    \caption{Avg final tokens used per LLM per game for each framework (continued).}
    \label{tab:all_model_token_comparison_2}
\end{table}

\clearpage
\section{Agent Score Standard Deviations}
In Tables \ref{tab:llm_std_stats} and \ref{tab:llm_std_stats_2}, we include the average standard deviation across seeds per framework of every configuration in Table \ref{tab:all_model_framework_scores}.

\label{app:all_stds}
\begin{table}[h!]
    \footnotesize
    \centering
    \renewcommand{\arraystretch}{1.2}
    \resizebox{\textwidth}{!}{
        \begin{tabular}{l c c c c c }
            \toprule
            \textbf{Model} & \textworld & \twx & \alfworld & \scienceworld & \jericho  \\
            \midrule
claude-fable-5 (thinking)$^{\ast\dagger}$                  & \textemdash & \textemdash & \textemdash & \textemdash & \textemdash \\
claude-opus-5 (thinking)$^\ast$                            & 0.0 & 0.0 & 0.0 & 3.9 & 3.5 \\
claude-opus-4.7 (thinking)$^\ast$                          & 0.0 & 0.0 & 0.0 & 3.4 & 3.5 \\
gpt-5.4 (thinking)                                         & 0.0 & 5.6 & 0.0 & 3.6 & 4.1 \\
claude-opus-4.6 (thinking)$^\ast$                          & 0.0 & 1.8 & 0.0 & 5.5 & 4.8 \\
claude-opus-4.8 (thinking)$^\ast$                          & 0.0 & 1.3 & 0.0 & 2.3 & 4.2 \\
claude-opus-4.5 (thinking)                                 & 0.0 & 4.6 & 0.0 & 5.0 & 4.9 \\
\bottomrule
        \end{tabular}
    }
    \caption{Standard deviation statistics for different LLMs. Ordering and labels follow Table \ref{tab:all_model_framework_scores}; claude-fable-5 has one seed, so it has no standard deviation across seeds.}
    \label{tab:llm_std_stats}
\end{table}
\begin{table}[h!]
    \footnotesize
    \centering
    \renewcommand{\arraystretch}{1.2}
    \resizebox{\textwidth}{!}{
        \begin{tabular}{l c c c c c }
            \toprule
            \textbf{Model} & \textworld & \twx & \alfworld & \scienceworld & \jericho  \\
            \midrule
Qwen2.5-72B-Instruct                                       & 1.8 & 3.1 & 7.4 & 8.2 & 1.2 \\
Qwen3.5-9B (1024)                                          & 21.7 & 5.5 & 28.2 & 15.4 & 1.5 \\
Mistral-Large-Instruct-2407                                & 8.2 & 2.9 & 10.7 & 16.4 & 1.5 \\
o3-mini (high)                                             & 13.9 & 3.1 & 15.6 & 17.8 & 2.0 \\
MiniMax-M2.7 (1024)                                        & 39.6 & 36.5 & 33.0 & 21.9 & 2.7 \\
gpt-4.1-mini                                               & 16.8 & 6.2 & 7.4 & 22.2 & 0.8 \\
Llama-3.3-Nemotron-Super-49B-v1 (1024)                     & 0.0 & 0.0 & 0.0 & 3.6 & 0.0 \\
\bottomrule
        \end{tabular}
    }
    \caption{Standard deviation statistics for different LLMs (continued).}
    \label{tab:llm_std_stats_2}
\end{table}

\clearpage
\section{All Games}

In Table \ref{tab:games_by_framework1} and Table \ref{tab:games_by_framework2} we list all tasks and games in their respective frameworks.
\label{app:all_games}

\begin{table*}[htbp]
\centering
\caption{Games Organized by Framework. Part 1.}
\label{tab:games_by_framework1}

%

\end{table*}

\begin{table*}[htbp]
\centering
\caption{Games Organized by Framework. Part 2.}
\label{tab:games_by_framework2}

%
\end{minipage} \\
\hline
\end{tabular}

\end{table*}

\newpage
\section{All Scores per Game: \textworld}\label{app:tw_all_scores}
Tables \ref{tab:tw_game_scores_1_1}, \ref{tab:tw_game_scores_1_2}, \ref{tab:tw_game_scores_1_3} show the per-game scores of every configuration in Table \ref{tab:all_model_framework_scores}, in the same order and with the same labels, in \textworld across all seeds.

\begin{sidewaystable}[htbp]
\centering
\caption{Model Performance on \textworld tasks, models 1--26.}
\label{tab:tw_game_scores_1_1}
\resizebox{\textwidth}{!}{%
\begin{tabular}{lcccccccccc}
\toprule
Models & \rotatebox{285}{TWCookingLevel1} & \rotatebox{285}{TWCookingLevel10} & \rotatebox{285}{TWCookingLevel2} & \rotatebox{285}{TWCookingLevel3} & \rotatebox{285}{TWCookingLevel4} & \rotatebox{285}{TWCookingLevel5} & \rotatebox{285}{TWCookingLevel6} & \rotatebox{285}{TWCookingLevel7} & \rotatebox{285}{TWCookingLevel8} & \rotatebox{285}{TWCookingLevel9} \\
\midrule
claude-fable-5 (thinking)$^{\ast\dagger}$ & 100.0 & 100.0 & 100.0 & 100.0 & 100.0 & 100.0 & 100.0 & 100.0 & 100.0 & 100.0 \\
claude-opus-5 (thinking)$^\ast$ & 100.0 & 100.0 & 100.0 & 100.0 & 100.0 & 100.0 & 100.0 & 100.0 & 100.0 & 100.0 \\
claude-opus-4.7 (thinking)$^\ast$ & 100.0 & 100.0 & 100.0 & 100.0 & 100.0 & 100.0 & 100.0 & 100.0 & 100.0 & 100.0 \\
gpt-5.4 (thinking) & 100.0 & 100.0 & 100.0 & 100.0 & 100.0 & 100.0 & 100.0 & 100.0 & 100.0 & 100.0 \\
claude-opus-4.6 (thinking)$^\ast$ & 100.0 & 100.0 & 100.0 & 100.0 & 100.0 & 100.0 & 100.0 & 100.0 & 100.0 & 100.0 \\
claude-opus-4.8 (thinking)$^\ast$ & 100.0 & 100.0 & 100.0 & 100.0 & 100.0 & 100.0 & 100.0 & 100.0 & 100.0 & 100.0 \\
claude-opus-4.5 (thinking) & 100.0 & 100.0 & 100.0 & 100.0 & 100.0 & 100.0 & 100.0 & 100.0 & 100.0 & 100.0 \\
\bottomrule
\end{tabular}%
}
\end{sidewaystable}

\begin{sidewaystable}[htbp]
\centering
\caption{Model Performance on \textworld tasks, models 27--51.}
\label{tab:tw_game_scores_1_2}
\resizebox{\textwidth}{!}{%
\begin{tabular}{lcccccccccc}
\toprule
Models & \rotatebox{285}{TWCookingLevel1} & \rotatebox{285}{TWCookingLevel10} & \rotatebox{285}{TWCookingLevel2} & \rotatebox{285}{TWCookingLevel3} & \rotatebox{285}{TWCookingLevel4} & \rotatebox{285}{TWCookingLevel5} & \rotatebox{285}{TWCookingLevel6} & \rotatebox{285}{TWCookingLevel7} & \rotatebox{285}{TWCookingLevel8} & \rotatebox{285}{TWCookingLevel9} \\
\midrule
Qwen3.6-27B (1024) & 100.0 & 47.3 & 100.0 & 100.0 & 100.0 & 100.0 & 100.0 & 100.0 & 100.0 & 92.7 \\
o1 (high) & 100.0 & 78.2 & 100.0 & 100.0 & 100.0 & 100.0 & 100.0 & 100.0 & 100.0 & 100.0 \\
claude-sonnet-4.5 & 100.0 & 100.0 & 100.0 & 100.0 & 100.0 & 100.0 & 100.0 & 100.0 & 100.0 & 100.0 \\
gpt-4o & 86.7 & 14.5 & 70.0 & 100.0 & 86.7 & 100.0 & 100.0 & 100.0 & 100.0 & 78.2 \\
claude-3.5-haiku & 100.0 & 58.2 & 100.0 & 100.0 & 100.0 & 100.0 & 100.0 & 100.0 & 100.0 & 90.9 \\
Qwen3.6-35B-A3B (1024) & 86.7 & 0.0 & 100.0 & 100.0 & 86.7 & 100.0 & 100.0 & 100.0 & 100.0 & 43.6 \\
Llama-3.1-405B-Instruct & 100.0 & 54.5 & 85.0 & 100.0 & 100.0 & 100.0 & 100.0 & 100.0 & 100.0 & 69.1 \\
\bottomrule
\end{tabular}%
}
\end{sidewaystable}

\begin{sidewaystable}[htbp]
\centering
\caption{Model Performance on \textworld tasks, models 52--76.}
\label{tab:tw_game_scores_1_3}
\resizebox{\textwidth}{!}{%
\begin{tabular}{lcccccccccc}
\toprule
Models & \rotatebox{285}{TWCookingLevel1} & \rotatebox{285}{TWCookingLevel10} & \rotatebox{285}{TWCookingLevel2} & \rotatebox{285}{TWCookingLevel3} & \rotatebox{285}{TWCookingLevel4} & \rotatebox{285}{TWCookingLevel5} & \rotatebox{285}{TWCookingLevel6} & \rotatebox{285}{TWCookingLevel7} & \rotatebox{285}{TWCookingLevel8} & \rotatebox{285}{TWCookingLevel9} \\
\midrule
gpt-5-nano (thinking) & 60.0 & 12.7 & 85.0 & 100.0 & 33.3 & 0.0 & 60.0 & 84.0 & 20.0 & 45.5 \\
gpt-oss-20b (thinking) & 66.7 & 0.0 & 100.0 & 100.0 & 60.0 & 0.0 & 86.7 & 92.0 & 64.0 & 36.4 \\
Llama-4-Maverick-17B-128E-Instruct-FP8 & 100.0 & 0.0 & 25.0 & 100.0 & 20.0 & 0.0 & 100.0 & 32.0 & 16.0 & 41.8 \\
Mistral-Small-Instruct-2409 & 100.0 & 0.0 & 25.0 & 100.0 & 100.0 & 20.0 & 0.0 & 100.0 & 76.0 & 40.0 \\
DeepSeek-V4-Flash (1024) & 73.3 & 14.5 & 25.0 & 100.0 & 0.0 & 0.0 & 20.0 & 60.0 & 0.0 & 20.0 \\
Llama-3.1-8B-Instruct & 73.3 & 1.8 & 25.0 & 0.0 & 100.0 & 0.0 & 0.0 & 28.0 & 60.0 & 9.1 \\
DeepSeek-R1 (1024) & 66.7 & 0.0 & 25.0 & 75.0 & 100.0 & 0.0 & 0.0 & 28.0 & 40.0 & 36.4 \\
\bottomrule
\end{tabular}%
}
\end{sidewaystable}

\section{All Scores per Game: \twx}\label{app:twx_all_scores}
Tables \ref{tab:twx_game_scores_1_1}, \ref{tab:twx_game_scores_1_2}, \ref{tab:twx_game_scores_1_3} show the average per-game scores of every configuration in Table \ref{tab:all_model_framework_scores}, in the same order and with the same labels, in \twx across all seeds.

\begin{sidewaystable}[htbp]
\centering
\caption{Model Performance on \twx tasks, models 1--26.}
\label{tab:twx_game_scores_1_1}
\resizebox{\textwidth}{!}{%
\begin{tabular}{lcccccccccccccccc}
\toprule
Models & \rotatebox{285}{TWXArithmetic} & \rotatebox{285}{TWXCoinCollector} & \rotatebox{285}{TWXCookingWorld} & \rotatebox{285}{TWXMapReader} & \rotatebox{285}{TWXPeckingOrder} & \rotatebox{285}{TWXSimonSays10} & \rotatebox{285}{TWXSimonSays100} & \rotatebox{285}{TWXSimonSays50} & \rotatebox{285}{TWXSimonSaysWithMemory10} & \rotatebox{285}{TWXSimonSaysWithMemory100} & \rotatebox{285}{TWXSimonSaysWithMemory100Verbose} & \rotatebox{285}{TWXSimonSaysWithMemory10Verbose} & \rotatebox{285}{TWXSimonSaysWithMemory50} & \rotatebox{285}{TWXSimonSaysWithMemory50Verbose} & \rotatebox{285}{TWXSorting} & \rotatebox{285}{TWXTextWorldCommonsense} \\
\midrule
claude-fable-5 (thinking)$^{\ast\dagger}$ & 100.0 & 100.0 & 100.0 & 100.0 & 100.0 & 100.0 & 100.0 & 100.0 & 100.0 & 100.0 & 100.0 & 100.0 & 100.0 & 100.0 & 100.0 & 100.0 \\
claude-opus-5 (thinking)$^\ast$ & 100.0 & 100.0 & 100.0 & 100.0 & 100.0 & 100.0 & 99.2 & 100.0 & 100.0 & 100.0 & 100.0 & 100.0 & 100.0 & 100.0 & 100.0 & 50.0 \\
claude-opus-4.7 (thinking)$^\ast$ & 100.0 & 100.0 & 100.0 & 100.0 & 100.0 & 100.0 & 99.0 & 100.0 & 100.0 & 100.0 & 100.0 & 100.0 & 100.0 & 100.0 & 100.0 & 50.0 \\
gpt-5.4 (thinking) & 60.0 & 100.0 & 100.0 & 100.0 & 100.0 & 100.0 & 87.4 & 98.8 & 100.0 & 94.2 & 100.0 & 100.0 & 94.0 & 100.0 & 100.0 & 80.0 \\
claude-opus-4.6 (thinking)$^\ast$ & 100.0 & 100.0 & 76.8 & 100.0 & 100.0 & 100.0 & 100.0 & 100.0 & 100.0 & 100.0 & 100.0 & 100.0 & 100.0 & 100.0 & 100.0 & 100.0 \\
claude-opus-4.8 (thinking)$^\ast$ & 100.0 & 100.0 & 100.0 & 100.0 & 100.0 & 100.0 & 98.4 & 100.0 & 100.0 & 100.0 & 100.0 & 100.0 & 100.0 & 100.0 & 100.0 & 60.0 \\
claude-opus-4.5 (thinking) & 70.0 & 100.0 & 65.2 & 100.0 & 100.0 & 100.0 & 100.0 & 100.0 & 100.0 & 100.0 & 100.0 & 100.0 & 100.0 & 100.0 & 0.0 & 90.0 \\
\bottomrule
\end{tabular}%
}
\end{sidewaystable}

\begin{sidewaystable}[htbp]
\centering
\caption{Model Performance on \twx tasks, models 27--51.}
\label{tab:twx_game_scores_1_2}
\resizebox{\textwidth}{!}{%
\begin{tabular}{lcccccccccccccccc}
\toprule
Models & \rotatebox{285}{TWXArithmetic} & \rotatebox{285}{TWXCoinCollector} & \rotatebox{285}{TWXCookingWorld} & \rotatebox{285}{TWXMapReader} & \rotatebox{285}{TWXPeckingOrder} & \rotatebox{285}{TWXSimonSays10} & \rotatebox{285}{TWXSimonSays100} & \rotatebox{285}{TWXSimonSays50} & \rotatebox{285}{TWXSimonSaysWithMemory10} & \rotatebox{285}{TWXSimonSaysWithMemory100} & \rotatebox{285}{TWXSimonSaysWithMemory100Verbose} & \rotatebox{285}{TWXSimonSaysWithMemory10Verbose} & \rotatebox{285}{TWXSimonSaysWithMemory50} & \rotatebox{285}{TWXSimonSaysWithMemory50Verbose} & \rotatebox{285}{TWXSorting} & \rotatebox{285}{TWXTextWorldCommonsense} \\
\midrule
Qwen3.6-27B (1024) & 60.0 & 100.0 & 42.0 & 100.0 & 100.0 & 100.0 & 100.0 & 100.0 & 100.0 & 87.2 & 93.0 & 100.0 & 100.0 & 100.0 & 100.0 & 50.0 \\
o1 (high) & 60.0 & 100.0 & 42.0 & 90.0 & 100.0 & 100.0 & 100.0 & 100.0 & 100.0 & 20.4 & 39.6 & 100.0 & 46.0 & 74.8 & 0.0 & 50.0 \\
claude-sonnet-4.5 & 0.0 & 100.0 & 42.0 & 100.0 & 0.0 & 100.0 & 53.0 & 100.0 & 100.0 & 99.0 & 99.0 & 100.0 & 100.0 & 100.0 & 0.0 & 50.0 \\
gpt-4o & 0.0 & 100.0 & 39.2 & 100.0 & 100.0 & 100.0 & 100.0 & 100.0 & 100.0 & 100.0 & 100.0 & 100.0 & 100.0 & 100.0 & 0.0 & 50.0 \\
claude-3.5-haiku & 0.0 & 100.0 & 28.0 & 100.0 & 100.0 & 100.0 & 100.0 & 100.0 & 100.0 & 99.0 & 99.0 & 100.0 & 100.0 & 100.0 & 0.0 & 50.0 \\
Qwen3.6-35B-A3B (1024) & 50.0 & 100.0 & 42.0 & 0.0 & 100.0 & 100.0 & 100.0 & 100.0 & 100.0 & 71.4 & 41.6 & 100.0 & 100.0 & 100.0 & 20.0 & 50.0 \\
Llama-3.1-405B-Instruct & 80.0 & 100.0 & 33.6 & 100.0 & 100.0 & 100.0 & 21.0 & 42.0 & 100.0 & 100.0 & 100.0 & 100.0 & 100.0 & 100.0 & 40.0 & 50.0 \\
\bottomrule
\end{tabular}%
}
\end{sidewaystable}

\begin{sidewaystable}[htbp]
\centering
\caption{Model Performance on \twx tasks, models 52--76.}
\label{tab:twx_game_scores_1_3}
\resizebox{\textwidth}{!}{%
\begin{tabular}{lcccccccccccccccc}
\toprule
Models & \rotatebox{285}{TWXArithmetic} & \rotatebox{285}{TWXCoinCollector} & \rotatebox{285}{TWXCookingWorld} & \rotatebox{285}{TWXMapReader} & \rotatebox{285}{TWXPeckingOrder} & \rotatebox{285}{TWXSimonSays10} & \rotatebox{285}{TWXSimonSays100} & \rotatebox{285}{TWXSimonSays50} & \rotatebox{285}{TWXSimonSaysWithMemory10} & \rotatebox{285}{TWXSimonSaysWithMemory100} & \rotatebox{285}{TWXSimonSaysWithMemory100Verbose} & \rotatebox{285}{TWXSimonSaysWithMemory10Verbose} & \rotatebox{285}{TWXSimonSaysWithMemory50} & \rotatebox{285}{TWXSimonSaysWithMemory50Verbose} & \rotatebox{285}{TWXSorting} & \rotatebox{285}{TWXTextWorldCommonsense} \\
\midrule
gpt-5-nano (thinking) & 0.0 & 100.0 & 0.0 & 0.0 & 60.0 & 100.0 & 79.0 & 86.8 & 74.0 & 7.4 & 9.4 & 74.0 & 18.4 & 21.2 & 0.0 & 30.0 \\
gpt-oss-20b (thinking) & 10.0 & 100.0 & 36.4 & 50.0 & 60.0 & 100.0 & 26.8 & 65.2 & 40.0 & 1.0 & 21.6 & 100.0 & 2.0 & 51.2 & 8.0 & 50.0 \\
Llama-4-Maverick-17B-128E-Instruct-FP8 & 0.0 & 100.0 & 42.0 & 100.0 & 100.0 & 100.0 & 55.0 & 100.0 & 100.0 & 13.0 & 5.0 & 100.0 & 26.0 & 6.0 & 0.0 & 50.0 \\
Mistral-Small-Instruct-2409 & 0.0 & 100.0 & 28.0 & 0.0 & 0.0 & 100.0 & 21.0 & 42.0 & 10.0 & 1.0 & 7.0 & 50.0 & 14.0 & 14.0 & 0.0 & 50.0 \\
DeepSeek-V4-Flash (1024) & 0.0 & 100.0 & 30.8 & 0.0 & 40.0 & 100.0 & 72.4 & 100.0 & 48.0 & 7.2 & 9.0 & 70.0 & 11.2 & 24.0 & 40.0 & 30.0 \\
Llama-3.1-8B-Instruct & 10.0 & 100.0 & 2.8 & 0.0 & 100.0 & 100.0 & 100.0 & 100.0 & 100.0 & 1.0 & 1.0 & 100.0 & 2.0 & 2.0 & 36.0 & 50.0 \\
DeepSeek-R1 (1024) & 0.0 & 100.0 & 42.0 & 0.0 & 100.0 & 100.0 & 21.0 & 42.0 & 10.0 & 1.0 & 12.0 & 100.0 & 28.0 & 12.0 & 0.0 & 50.0 \\
\bottomrule
\end{tabular}%
}
\end{sidewaystable}

\section{All Scores per Game: \alfworld}\label{app:alf_all_scores}
Tables \ref{tab:alfworld_game_scores_1_1}, \ref{tab:alfworld_game_scores_1_2}, \ref{tab:alfworld_game_scores_1_3} show the average per-game scores of every configuration in Table \ref{tab:all_model_framework_scores}, in the same order and with the same labels, in \alfworld across all seeds.

\begin{sidewaystable}[htbp]
\centering
\caption{Model Performance on \alfworld tasks, models 1--26.}
\label{tab:alfworld_game_scores_1_1}
\resizebox{\textwidth}{!}{%
\begin{tabular}{lcccccccccccc}
\toprule
Models & \rotatebox{285}{ALFWorldLookAtObjInLightSeen} & \rotatebox{285}{ALFWorldLookAtObjInLightUnseen} & \rotatebox{285}{ALFWorldPickAndPlaceSimpleSeen} & \rotatebox{285}{ALFWorldPickAndPlaceSimpleUnseen} & \rotatebox{285}{ALFWorldPickCleanThenPlaceInRecepSeen} & \rotatebox{285}{ALFWorldPickCleanThenPlaceInRecepUnseen} & \rotatebox{285}{ALFWorldPickCoolThenPlaceInRecepSeen} & \rotatebox{285}{ALFWorldPickCoolThenPlaceInRecepUnseen} & \rotatebox{285}{ALFWorldPickHeatThenPlaceInRecepSeen} & \rotatebox{285}{ALFWorldPickHeatThenPlaceInRecepUnseen} & \rotatebox{285}{ALFWorldPickTwoObjAndPlaceSeen} & \rotatebox{285}{ALFWorldPickTwoObjAndPlaceUnseen} \\
\midrule
claude-fable-5 (thinking)$^{\ast\dagger}$ & 100.0 & 100.0 & 100.0 & 100.0 & 100.0 & 100.0 & 100.0 & 100.0 & 100.0 & 100.0 & 100.0 & 100.0 \\
claude-opus-5 (thinking)$^\ast$ & 100.0 & 100.0 & 100.0 & 100.0 & 100.0 & 100.0 & 100.0 & 100.0 & 100.0 & 100.0 & 100.0 & 100.0 \\
claude-opus-4.7 (thinking)$^\ast$ & 100.0 & 100.0 & 100.0 & 100.0 & 100.0 & 100.0 & 100.0 & 100.0 & 100.0 & 100.0 & 100.0 & 100.0 \\
gpt-5.4 (thinking) & 100.0 & 100.0 & 100.0 & 100.0 & 100.0 & 100.0 & 100.0 & 100.0 & 100.0 & 100.0 & 100.0 & 100.0 \\
claude-opus-4.6 (thinking)$^\ast$ & 100.0 & 100.0 & 100.0 & 100.0 & 100.0 & 100.0 & 100.0 & 100.0 & 100.0 & 100.0 & 100.0 & 100.0 \\
claude-opus-4.8 (thinking)$^\ast$ & 100.0 & 100.0 & 100.0 & 100.0 & 100.0 & 100.0 & 100.0 & 100.0 & 100.0 & 100.0 & 100.0 & 100.0 \\
claude-opus-4.5 (thinking) & 100.0 & 100.0 & 100.0 & 100.0 & 100.0 & 100.0 & 100.0 & 100.0 & 100.0 & 100.0 & 100.0 & 100.0 \\
o3 (medium) & 100.0 & 100.0 & 100.0 & 100.0 & 60.0 & 100.0 & 100.0 & 100.0 & 60.0 & 60.0 & 100.0 & 80.0 \\
gpt-5.6-terra (thinking) & 100.0 & 100.0 & 100.0 & 100.0 & 100.0 & 100.0 & 100.0 & 100.0 & 100.0 & 100.0 & 100.0 & 100.0 \\
gpt-5.1 (thinking) & 80.0 & 60.0 & 100.0 & 100.0 & 100.0 & 100.0 & 100.0 & 100.0 & 80.0 & 80.0 & 100.0 & 100.0 \\
o3 (high) & 60.0 & 100.0 & 100.0 & 100.0 & 20.0 & 60.0 & 80.0 & 80.0 & 100.0 & 80.0 & 100.0 & 100.0 \\
claude-sonnet-4.6 (thinking) & 100.0 & 100.0 & 100.0 & 100.0 & 100.0 & 100.0 & 100.0 & 100.0 & 100.0 & 100.0 & 100.0 & 100.0 \\
gpt-5 (thinking) & 100.0 & 100.0 & 100.0 & 100.0 & 100.0 & 100.0 & 80.0 & 100.0 & 80.0 & 60.0 & 100.0 & 100.0 \\
kimi-k2.6 (thinking) & 100.0 & 100.0 & 100.0 & 100.0 & 80.0 & 100.0 & 100.0 & 100.0 & 100.0 & 100.0 & 80.0 & 100.0 \\
gpt-5.6-luna (thinking) & 80.0 & 100.0 & 100.0 & 100.0 & 80.0 & 100.0 & 100.0 & 100.0 & 80.0 & 20.0 & 100.0 & 100.0 \\
o3 (low) & 80.0 & 100.0 & 80.0 & 100.0 & 60.0 & 80.0 & 20.0 & 80.0 & 40.0 & 40.0 & 100.0 & 60.0 \\
claude-4-sonnet & 100.0 & 80.0 & 100.0 & 100.0 & 100.0 & 100.0 & 100.0 & 100.0 & 100.0 & 100.0 & 100.0 & 20.0 \\
kimi-k2.5 (thinking) & 80.0 & 100.0 & 100.0 & 100.0 & 60.0 & 100.0 & 100.0 & 100.0 & 60.0 & 60.0 & 100.0 & 100.0 \\
claude-3.7-sonnet (1024) & 20.0 & 100.0 & 100.0 & 100.0 & 80.0 & 100.0 & 100.0 & 100.0 & 60.0 & 80.0 & 80.0 & 80.0 \\
claude-3.7-sonnet & 0.0 & 100.0 & 100.0 & 100.0 & 100.0 & 100.0 & 100.0 & 100.0 & 80.0 & 0.0 & 100.0 & 100.0 \\
claude-3.5-sonnet-latest & 20.0 & 20.0 & 100.0 & 100.0 & 100.0 & 80.0 & 100.0 & 100.0 & 20.0 & 60.0 & 100.0 & 100.0 \\
gpt-4.1 & 80.0 & 100.0 & 100.0 & 100.0 & 100.0 & 80.0 & 60.0 & 60.0 & 80.0 & 40.0 & 100.0 & 100.0 \\
grok-4.3 (thinking) & 100.0 & 100.0 & 100.0 & 100.0 & 100.0 & 100.0 & 100.0 & 100.0 & 100.0 & 100.0 & 100.0 & 60.0 \\
Qwen3.5-122B-A10B (1024) & 100.0 & 0.0 & 100.0 & 100.0 & 100.0 & 100.0 & 0.0 & 20.0 & 20.0 & 100.0 & 100.0 & 20.0 \\
Qwen3.5-27B (1024) & 100.0 & 0.0 & 100.0 & 100.0 & 80.0 & 80.0 & 100.0 & 60.0 & 60.0 & 20.0 & 100.0 & 100.0 \\
gpt-5-mini (thinking) & 100.0 & 0.0 & 100.0 & 100.0 & 20.0 & 60.0 & 80.0 & 40.0 & 40.0 & 0.0 & 100.0 & 100.0 \\
\bottomrule
\end{tabular}%
}
\end{sidewaystable}

\begin{sidewaystable}[htbp]
\centering
\caption{Model Performance on \alfworld tasks, models 27--51.}
\label{tab:alfworld_game_scores_1_2}
\resizebox{\textwidth}{!}{%
\begin{tabular}{lcccccccccccc}
\toprule
Models & \rotatebox{285}{ALFWorldLookAtObjInLightSeen} & \rotatebox{285}{ALFWorldLookAtObjInLightUnseen} & \rotatebox{285}{ALFWorldPickAndPlaceSimpleSeen} & \rotatebox{285}{ALFWorldPickAndPlaceSimpleUnseen} & \rotatebox{285}{ALFWorldPickCleanThenPlaceInRecepSeen} & \rotatebox{285}{ALFWorldPickCleanThenPlaceInRecepUnseen} & \rotatebox{285}{ALFWorldPickCoolThenPlaceInRecepSeen} & \rotatebox{285}{ALFWorldPickCoolThenPlaceInRecepUnseen} & \rotatebox{285}{ALFWorldPickHeatThenPlaceInRecepSeen} & \rotatebox{285}{ALFWorldPickHeatThenPlaceInRecepUnseen} & \rotatebox{285}{ALFWorldPickTwoObjAndPlaceSeen} & \rotatebox{285}{ALFWorldPickTwoObjAndPlaceUnseen} \\
\midrule
Qwen3.6-27B (1024) & 40.0 & 40.0 & 100.0 & 80.0 & 80.0 & 100.0 & 80.0 & 60.0 & 60.0 & 60.0 & 100.0 & 40.0 \\
o1 (high) & 100.0 & 40.0 & 40.0 & 20.0 & 0.0 & 40.0 & 40.0 & 0.0 & 60.0 & 0.0 & 0.0 & 0.0 \\
claude-sonnet-4.5 & 100.0 & 100.0 & 100.0 & 100.0 & 100.0 & 100.0 & 0.0 & 100.0 & 100.0 & 100.0 & 100.0 & 0.0 \\
gpt-4o & 40.0 & 100.0 & 60.0 & 100.0 & 60.0 & 60.0 & 0.0 & 20.0 & 60.0 & 40.0 & 60.0 & 80.0 \\
claude-3.5-haiku & 100.0 & 0.0 & 100.0 & 80.0 & 0.0 & 0.0 & 0.0 & 0.0 & 0.0 & 20.0 & 0.0 & 20.0 \\
Qwen3.6-35B-A3B (1024) & 0.0 & 0.0 & 40.0 & 40.0 & 20.0 & 60.0 & 60.0 & 100.0 & 60.0 & 20.0 & 60.0 & 40.0 \\
Llama-3.1-405B-Instruct & 80.0 & 0.0 & 100.0 & 20.0 & 0.0 & 60.0 & 0.0 & 0.0 & 0.0 & 0.0 & 100.0 & 20.0 \\
gemma-4-31b-it (1024) & 0.0 & 80.0 & 100.0 & 20.0 & 0.0 & 100.0 & 100.0 & 20.0 & 100.0 & 0.0 & 100.0 & 100.0 \\
gemini-2.0-flash & 0.0 & 0.0 & 100.0 & 40.0 & 100.0 & 0.0 & 0.0 & 0.0 & 0.0 & 0.0 & 0.0 & 0.0 \\
Qwen3-32B (1024) & 80.0 & 0.0 & 100.0 & 60.0 & 80.0 & 60.0 & 20.0 & 40.0 & 40.0 & 20.0 & 80.0 & 0.0 \\
Llama-3.3-70B-Instruct & 100.0 & 20.0 & 40.0 & 0.0 & 0.0 & 20.0 & 0.0 & 0.0 & 0.0 & 0.0 & 0.0 & 0.0 \\
Llama-3.1-70B-Instruct & 40.0 & 0.0 & 60.0 & 0.0 & 0.0 & 0.0 & 0.0 & 0.0 & 0.0 & 0.0 & 0.0 & 0.0 \\
Qwen2.5-72B-Instruct & 100.0 & 0.0 & 100.0 & 100.0 & 0.0 & 80.0 & 0.0 & 0.0 & 0.0 & 0.0 & 60.0 & 0.0 \\
Qwen3.5-9B (1024) & 20.0 & 0.0 & 60.0 & 20.0 & 40.0 & 0.0 & 20.0 & 20.0 & 20.0 & 0.0 & 20.0 & 0.0 \\
Mistral-Large-Instruct-2407 & 20.0 & 0.0 & 0.0 & 40.0 & 0.0 & 0.0 & 0.0 & 0.0 & 0.0 & 0.0 & 20.0 & 0.0 \\
o3-mini (high) & 40.0 & 0.0 & 40.0 & 0.0 & 0.0 & 40.0 & 0.0 & 20.0 & 0.0 & 0.0 & 0.0 & 0.0 \\
MiniMax-M2.7 (1024) & 40.0 & 20.0 & 40.0 & 0.0 & 20.0 & 40.0 & 0.0 & 0.0 & 20.0 & 40.0 & 20.0 & 20.0 \\
gpt-4.1-mini & 0.0 & 0.0 & 20.0 & 40.0 & 0.0 & 0.0 & 0.0 & 0.0 & 0.0 & 0.0 & 0.0 & 0.0 \\
Llama-3.3-Nemotron-Super-49B-v1 (1024) & 0.0 & 0.0 & 100.0 & 0.0 & 0.0 & 0.0 & 0.0 & 0.0 & 0.0 & 0.0 & 100.0 & 0.0 \\
gpt-5.4-nano (thinking) & 60.0 & 0.0 & 0.0 & 0.0 & 0.0 & 0.0 & 0.0 & 0.0 & 0.0 & 0.0 & 0.0 & 0.0 \\
Qwen3-30B-A3B (1024) & 0.0 & 0.0 & 80.0 & 20.0 & 40.0 & 0.0 & 0.0 & 0.0 & 0.0 & 0.0 & 40.0 & 20.0 \\
gpt-4o-mini & 0.0 & 0.0 & 0.0 & 0.0 & 0.0 & 0.0 & 0.0 & 0.0 & 0.0 & 0.0 & 0.0 & 0.0 \\
claude-haiku-4.5 & 0.0 & 0.0 & 0.0 & 0.0 & 0.0 & 0.0 & 0.0 & 0.0 & 0.0 & 0.0 & 0.0 & 0.0 \\
Llama-4-Scout-17B-16E-Instruct & 0.0 & 0.0 & 0.0 & 0.0 & 0.0 & 0.0 & 0.0 & 0.0 & 0.0 & 0.0 & 0.0 & 0.0 \\
Qwen3.5-4B (1024) & 0.0 & 0.0 & 0.0 & 0.0 & 0.0 & 0.0 & 0.0 & 0.0 & 0.0 & 0.0 & 0.0 & 0.0 \\
\bottomrule
\end{tabular}%
}
\end{sidewaystable}

\begin{sidewaystable}[htbp]
\centering
\caption{Model Performance on \alfworld tasks, models 52--76.}
\label{tab:alfworld_game_scores_1_3}
\resizebox{\textwidth}{!}{%
\begin{tabular}{lcccccccccccc}
\toprule
Models & \rotatebox{285}{ALFWorldLookAtObjInLightSeen} & \rotatebox{285}{ALFWorldLookAtObjInLightUnseen} & \rotatebox{285}{ALFWorldPickAndPlaceSimpleSeen} & \rotatebox{285}{ALFWorldPickAndPlaceSimpleUnseen} & \rotatebox{285}{ALFWorldPickCleanThenPlaceInRecepSeen} & \rotatebox{285}{ALFWorldPickCleanThenPlaceInRecepUnseen} & \rotatebox{285}{ALFWorldPickCoolThenPlaceInRecepSeen} & \rotatebox{285}{ALFWorldPickCoolThenPlaceInRecepUnseen} & \rotatebox{285}{ALFWorldPickHeatThenPlaceInRecepSeen} & \rotatebox{285}{ALFWorldPickHeatThenPlaceInRecepUnseen} & \rotatebox{285}{ALFWorldPickTwoObjAndPlaceSeen} & \rotatebox{285}{ALFWorldPickTwoObjAndPlaceUnseen} \\
\midrule
gpt-5-nano (thinking) & 0.0 & 0.0 & 0.0 & 20.0 & 0.0 & 0.0 & 0.0 & 0.0 & 0.0 & 0.0 & 0.0 & 0.0 \\
gpt-oss-20b (thinking) & 0.0 & 0.0 & 0.0 & 0.0 & 0.0 & 0.0 & 0.0 & 0.0 & 0.0 & 0.0 & 0.0 & 0.0 \\
Llama-4-Maverick-17B-128E-Instruct-FP8 & 0.0 & 0.0 & 100.0 & 0.0 & 0.0 & 0.0 & 0.0 & 0.0 & 0.0 & 0.0 & 0.0 & 0.0 \\
Mistral-Small-Instruct-2409 & 0.0 & 0.0 & 0.0 & 0.0 & 0.0 & 0.0 & 0.0 & 0.0 & 0.0 & 0.0 & 0.0 & 0.0 \\
DeepSeek-V4-Flash (1024) & 0.0 & 0.0 & 0.0 & 0.0 & 0.0 & 0.0 & 0.0 & 0.0 & 0.0 & 0.0 & 0.0 & 20.0 \\
Llama-3.1-8B-Instruct & 0.0 & 0.0 & 0.0 & 0.0 & 0.0 & 0.0 & 0.0 & 0.0 & 0.0 & 0.0 & 0.0 & 0.0 \\
DeepSeek-R1 (1024) & 0.0 & 0.0 & 0.0 & 0.0 & 0.0 & 0.0 & 0.0 & 0.0 & 0.0 & 0.0 & 0.0 & 0.0 \\
Qwen2.5-7B-Instruct & 0.0 & 0.0 & 0.0 & 0.0 & 0.0 & 0.0 & 0.0 & 0.0 & 0.0 & 0.0 & 0.0 & 0.0 \\
Llama-3.2-3B-Instruct & 0.0 & 0.0 & 0.0 & 0.0 & 0.0 & 0.0 & 0.0 & 0.0 & 0.0 & 0.0 & 0.0 & 0.0 \\
phi-4 & 0.0 & 0.0 & 0.0 & 0.0 & 0.0 & 0.0 & 0.0 & 0.0 & 0.0 & 0.0 & 0.0 & 0.0 \\
gpt-4.1-nano & 0.0 & 0.0 & 0.0 & 0.0 & 0.0 & 0.0 & 0.0 & 0.0 & 0.0 & 0.0 & 0.0 & 0.0 \\
Mistral-Small-24B-Instruct-2501 & 0.0 & 0.0 & 0.0 & 0.0 & 0.0 & 0.0 & 0.0 & 0.0 & 0.0 & 0.0 & 0.0 & 0.0 \\
DeepSeek-R1-Distill-Llama-70B (1024) & 0.0 & 0.0 & 0.0 & 0.0 & 0.0 & 0.0 & 0.0 & 0.0 & 0.0 & 0.0 & 0.0 & 0.0 \\
Qwen3.5-2B (1024) & 0.0 & 0.0 & 0.0 & 0.0 & 0.0 & 0.0 & 0.0 & 0.0 & 0.0 & 0.0 & 0.0 & 0.0 \\
Ministral-8B-Instruct-2410 & 0.0 & 0.0 & 0.0 & 0.0 & 0.0 & 0.0 & 0.0 & 0.0 & 0.0 & 0.0 & 0.0 & 0.0 \\
Mistral-Small-3.1-24B-Instruct-2503 & 0.0 & 0.0 & 0.0 & 0.0 & 0.0 & 0.0 & 0.0 & 0.0 & 0.0 & 0.0 & 0.0 & 0.0 \\
Mixtral-8x22B-Instruct-v0.1 & 0.0 & 0.0 & 0.0 & 0.0 & 0.0 & 0.0 & 0.0 & 0.0 & 0.0 & 0.0 & 0.0 & 0.0 \\
Llama-3.2-1B-Instruct & 0.0 & 0.0 & 0.0 & 0.0 & 0.0 & 0.0 & 0.0 & 0.0 & 0.0 & 0.0 & 0.0 & 0.0 \\
Phi-3-mini-128k-instruct & 0.0 & 0.0 & 0.0 & 0.0 & 0.0 & 0.0 & 0.0 & 0.0 & 0.0 & 0.0 & 0.0 & 0.0 \\
Phi-3.5-MoE-instruct & 0.0 & 0.0 & 0.0 & 0.0 & 0.0 & 0.0 & 0.0 & 0.0 & 0.0 & 0.0 & 0.0 & 0.0 \\
Phi-4-mini-instruct & 0.0 & 0.0 & 0.0 & 0.0 & 0.0 & 0.0 & 0.0 & 0.0 & 0.0 & 0.0 & 0.0 & 0.0 \\
Mixtral-8x7B-Instruct-v0.1 & 0.0 & 0.0 & 0.0 & 0.0 & 0.0 & 0.0 & 0.0 & 0.0 & 0.0 & 0.0 & 0.0 & 0.0 \\
Phi-3.5-mini-instruct & 0.0 & 0.0 & 0.0 & 0.0 & 0.0 & 0.0 & 0.0 & 0.0 & 0.0 & 0.0 & 0.0 & 0.0 \\
Qwen3.5-0.8B (1024) & 0.0 & 0.0 & 0.0 & 0.0 & 0.0 & 0.0 & 0.0 & 0.0 & 0.0 & 0.0 & 0.0 & 0.0 \\
Phi-3-medium-128k-instruct & 0.0 & 0.0 & 0.0 & 0.0 & 0.0 & 0.0 & 0.0 & 0.0 & 0.0 & 0.0 & 0.0 & 0.0 \\
\bottomrule
\end{tabular}%
}
\end{sidewaystable}

\section{All Scores per Game: \scienceworld}\label{app:sw_all_scores}
Tables \ref{tab:sw_game_scores_1_1}, \ref{tab:sw_game_scores_1_2}, \ref{tab:sw_game_scores_1_3}, \ref{tab:sw_game_scores_2_1}, \ref{tab:sw_game_scores_2_2}, and \ref{tab:sw_game_scores_2_3} show the average per-game scores of every configuration in Table \ref{tab:all_model_framework_scores}, in the same order and with the same labels, in \scienceworld across all seeds. \scienceworld tasks are split into two 15-task table sets for readability.

\begin{sidewaystable}[htbp]
\centering
\caption{Model Performance on \scienceworld tasks, task set 1, models 1--26.}
\label{tab:sw_game_scores_1_1}
\resizebox{\textwidth}{!}{%
\begin{tabular}{lccccccccccccccc}
\toprule
Models & \rotatebox{285}{ScienceWorldBoil} & \rotatebox{285}{ScienceWorldChangeTheStateOfMatterOf} & \rotatebox{285}{ScienceWorldChemistryMix} & \rotatebox{285}{ScienceWorldChemistryMixPaintSecondaryColor} & \rotatebox{285}{ScienceWorldChemistryMixPaintTertiaryColor} & \rotatebox{285}{ScienceWorldFindAnimal} & \rotatebox{285}{ScienceWorldFindLivingThing} & \rotatebox{285}{ScienceWorldFindNonLivingThing} & \rotatebox{285}{ScienceWorldFindPlant} & \rotatebox{285}{ScienceWorldFreeze} & \rotatebox{285}{ScienceWorldGrowFruit} & \rotatebox{285}{ScienceWorldGrowPlant} & \rotatebox{285}{ScienceWorldIdentifyLifeStages1} & \rotatebox{285}{ScienceWorldIdentifyLifeStages2} & \rotatebox{285}{ScienceWorldInclinedPlaneDetermineAngle} \\
\midrule
claude-fable-5 (thinking)$^{\ast\dagger}$ & 100.0 & 100.0 & 100.0 & 100.0 & 100.0 & 100.0 & 100.0 & 100.0 & 100.0 & 100.0 & 100.0 & 100.0 & 100.0 & 100.0 & 100.0 \\
claude-opus-5 (thinking)$^\ast$ & 100.0 & 100.0 & 100.0 & 100.0 & 100.0 & 89.8 & 96.6 & 100.0 & 100.0 & 100.0 & 78.0 & 100.0 & 100.0 & 100.0 & 100.0 \\
claude-opus-4.7 (thinking)$^\ast$ & 61.6 & 100.0 & 100.0 & 100.0 & 94.6 & 96.6 & 100.0 & 100.0 & 100.0 & 69.8 & 100.0 & 100.0 & 100.0 & 100.0 & 100.0 \\
gpt-5.4 (thinking) & 96.0 & 100.0 & 100.0 & 100.0 & 100.0 & 84.8 & 100.0 & 100.0 & 100.0 & 59.4 & 100.0 & 100.0 & 64.4 & 100.0 & 100.0 \\
claude-opus-4.6 (thinking)$^\ast$ & 81.8 & 100.0 & 100.0 & 100.0 & 100.0 & 100.0 & 100.0 & 100.0 & 100.0 & 61.6 & 50.6 & 63.8 & 100.0 & 100.0 & 100.0 \\
claude-opus-4.8 (thinking)$^\ast$ & 5.4 & 100.0 & 100.0 & 100.0 & 100.0 & 95.0 & 100.0 & 100.0 & 100.0 & 59.4 & 89.0 & 100.0 & 100.0 & 100.0 & 100.0 \\
claude-opus-4.5 (thinking) & 68.0 & 100.0 & 100.0 & 100.0 & 100.0 & 83.2 & 100.0 & 100.0 & 100.0 & 49.0 & 100.0 & 100.0 & 100.0 & 100.0 & 100.0 \\
\bottomrule
\end{tabular}%
}
\end{sidewaystable}

\begin{sidewaystable}[htbp]
\centering
\caption{Model Performance on \scienceworld tasks, task set 1, models 27--51.}
\label{tab:sw_game_scores_1_2}
\resizebox{\textwidth}{!}{%
\begin{tabular}{lccccccccccccccc}
\toprule
Models & \rotatebox{285}{ScienceWorldBoil} & \rotatebox{285}{ScienceWorldChangeTheStateOfMatterOf} & \rotatebox{285}{ScienceWorldChemistryMix} & \rotatebox{285}{ScienceWorldChemistryMixPaintSecondaryColor} & \rotatebox{285}{ScienceWorldChemistryMixPaintTertiaryColor} & \rotatebox{285}{ScienceWorldFindAnimal} & \rotatebox{285}{ScienceWorldFindLivingThing} & \rotatebox{285}{ScienceWorldFindNonLivingThing} & \rotatebox{285}{ScienceWorldFindPlant} & \rotatebox{285}{ScienceWorldFreeze} & \rotatebox{285}{ScienceWorldGrowFruit} & \rotatebox{285}{ScienceWorldGrowPlant} & \rotatebox{285}{ScienceWorldIdentifyLifeStages1} & \rotatebox{285}{ScienceWorldIdentifyLifeStages2} & \rotatebox{285}{ScienceWorldInclinedPlaneDetermineAngle} \\
\midrule
Qwen3.6-27B (1024) & 25.8 & 43.6 & 88.4 & 86.0 & 67.4 & 56.6 & 40.0 & 100.0 & 70.0 & 17.6 & 32.4 & 50.8 & 37.6 & 36.0 & 100.0 \\
o1 (high) & 21.4 & 80.8 & 100.0 & 100.0 & 82.2 & 91.6 & 96.6 & 100.0 & 100.0 & 37.4 & 29.8 & 62.4 & 35.0 & 28.4 & 100.0 \\
claude-sonnet-4.5 & 0.0 & 100.0 & 100.0 & 20.0 & 62.8 & 86.4 & 0.0 & 0.0 & 100.0 & 48.0 & 11.0 & 100.0 & 0.0 & 52.0 & 100.0 \\
gpt-4o & 2.8 & 2.6 & 75.0 & 28.0 & 22.0 & 100.0 & 90.0 & 100.0 & 70.0 & 0.0 & 27.0 & 11.4 & 20.0 & 64.8 & 82.0 \\
claude-3.5-haiku & 40.0 & 72.8 & 33.0 & 86.0 & 76.6 & 58.2 & 100.0 & 100.0 & 100.0 & 0.0 & 18.4 & 34.0 & 40.4 & 46.8 & 100.0 \\
Qwen3.6-35B-A3B (1024) & 7.0 & 18.4 & 63.4 & 48.0 & 32.0 & 70.0 & 100.0 & 100.0 & 100.0 & 36.4 & 38.6 & 11.2 & 32.0 & 13.6 & 100.0 \\
Llama-3.1-405B-Instruct & 3.2 & 2.2 & 42.0 & 86.0 & 20.0 & 45.0 & 50.0 & 100.0 & 70.0 & 17.6 & 21.0 & 11.6 & 12.0 & 20.0 & 100.0 \\
\bottomrule
\end{tabular}%
}
\end{sidewaystable}

\begin{sidewaystable}[htbp]
\centering
\caption{Model Performance on \scienceworld tasks, task set 1, models 52--76.}
\label{tab:sw_game_scores_1_3}
\resizebox{\textwidth}{!}{%
\begin{tabular}{lccccccccccccccc}
\toprule
Models & \rotatebox{285}{ScienceWorldBoil} & \rotatebox{285}{ScienceWorldChangeTheStateOfMatterOf} & \rotatebox{285}{ScienceWorldChemistryMix} & \rotatebox{285}{ScienceWorldChemistryMixPaintSecondaryColor} & \rotatebox{285}{ScienceWorldChemistryMixPaintTertiaryColor} & \rotatebox{285}{ScienceWorldFindAnimal} & \rotatebox{285}{ScienceWorldFindLivingThing} & \rotatebox{285}{ScienceWorldFindNonLivingThing} & \rotatebox{285}{ScienceWorldFindPlant} & \rotatebox{285}{ScienceWorldFreeze} & \rotatebox{285}{ScienceWorldGrowFruit} & \rotatebox{285}{ScienceWorldGrowPlant} & \rotatebox{285}{ScienceWorldIdentifyLifeStages1} & \rotatebox{285}{ScienceWorldIdentifyLifeStages2} & \rotatebox{285}{ScienceWorldInclinedPlaneDetermineAngle} \\
\midrule
gpt-5-nano (thinking) & 1.8 & 0.8 & 33.0 & 88.0 & 60.0 & 23.4 & 21.8 & 78.2 & 21.8 & 0.0 & 32.0 & 7.0 & 15.4 & 30.2 & 13.0 \\
gpt-oss-20b (thinking) & 7.8 & 15.6 & 31.6 & 32.0 & 13.4 & 20.2 & 31.8 & 38.4 & 20.0 & 0.0 & 16.6 & 7.4 & 4.0 & 15.2 & 12.0 \\
Llama-4-Maverick-17B-128E-Instruct-FP8 & 0.0 & 0.0 & 33.0 & 20.0 & 7.0 & 23.4 & 0.0 & 8.0 & 0.0 & 0.0 & 21.0 & 7.0 & 0.0 & 8.0 & 10.0 \\
Mistral-Small-Instruct-2409 & 0.0 & 0.4 & 43.2 & 72.0 & 8.8 & 25.0 & 25.0 & 95.0 & 36.6 & 0.0 & 13.4 & 7.6 & 0.0 & 9.6 & 0.0 \\
DeepSeek-V4-Flash (1024) & 0.0 & 0.0 & 26.4 & 38.0 & 38.8 & 5.0 & 5.0 & 15.0 & 1.6 & 0.0 & 14.4 & 6.0 & 1.6 & 10.4 & 0.0 \\
Llama-3.1-8B-Instruct & 0.4 & 0.0 & 8.0 & 24.0 & 19.6 & 20.0 & 18.4 & 48.2 & 10.0 & 0.0 & 29.2 & 7.4 & 4.0 & 14.4 & 3.0 \\
DeepSeek-R1 (1024) & 0.0 & 0.0 & 33.0 & 20.0 & 8.8 & 10.2 & 21.8 & 53.4 & 23.4 & 0.0 & 12.0 & 7.0 & 0.0 & 8.0 & 5.0 \\
\bottomrule
\end{tabular}%
}
\end{sidewaystable}

\begin{sidewaystable}[htbp]
\centering
\caption{Model Performance on \scienceworld tasks, task set 2, models 1--26.}
\label{tab:sw_game_scores_2_1}
\resizebox{\textwidth}{!}{%
\begin{tabular}{lccccccccccccccc}
\toprule
Models & \rotatebox{285}{ScienceWorldInclinedPlaneFrictionNamedSurfaces} & \rotatebox{285}{ScienceWorldInclinedPlaneFrictionUnnamedSurfaces} & \rotatebox{285}{ScienceWorldLifespanLongestLived} & \rotatebox{285}{ScienceWorldLifespanLongestLivedThenShortestLived} & \rotatebox{285}{ScienceWorldLifespanShortestLived} & \rotatebox{285}{ScienceWorldMeasureMeltingPointKnownSubstance} & \rotatebox{285}{ScienceWorldMeasureMeltingPointUnknownSubstance} & \rotatebox{285}{ScienceWorldMelt} & \rotatebox{285}{ScienceWorldMendelianGeneticsKnownPlant} & \rotatebox{285}{ScienceWorldMendelianGeneticsUnknownPlant} & \rotatebox{285}{ScienceWorldPowerComponent} & \rotatebox{285}{ScienceWorldPowerComponentRenewableVsNonrenewableEnergy} & \rotatebox{285}{ScienceWorldTestConductivity} & \rotatebox{285}{ScienceWorldTestConductivityOfUnknownSubstances} & \rotatebox{285}{ScienceWorldUseThermometer} \\
\midrule
claude-fable-5 (thinking)$^{\ast\dagger}$ & 100.0 & 100.0 & 100.0 & 100.0 & 100.0 & 100.0 & 100.0 & 100.0 & 0.0 & 0.0 & 100.0 & 100.0 & 100.0 & 100.0 & 100.0 \\
claude-opus-5 (thinking)$^\ast$ & 100.0 & 100.0 & 100.0 & 100.0 & 100.0 & 100.0 & 100.0 & 100.0 & 100.0 & 35.2 & 100.0 & 100.0 & 91.6 & 59.8 & 100.0 \\
claude-opus-4.7 (thinking)$^\ast$ & 100.0 & 100.0 & 100.0 & 100.0 & 100.0 & 100.0 & 100.0 & 100.0 & 100.0 & 66.8 & 100.0 & 100.0 & 100.0 & 100.0 & 100.0 \\
gpt-5.4 (thinking) & 100.0 & 100.0 & 100.0 & 100.0 & 100.0 & 100.0 & 100.0 & 100.0 & 100.0 & 51.4 & 100.0 & 100.0 & 100.0 & 100.0 & 100.0 \\
claude-opus-4.6 (thinking)$^\ast$ & 100.0 & 100.0 & 100.0 & 100.0 & 100.0 & 77.0 & 100.0 & 81.0 & 100.0 & 50.2 & 90.8 & 71.8 & 100.0 & 100.0 & 100.0 \\
claude-opus-4.8 (thinking)$^\ast$ & 100.0 & 100.0 & 100.0 & 100.0 & 100.0 & 100.0 & 84.2 & 100.0 & 100.0 & 17.0 & 100.0 & 100.0 & 100.0 & 96.6 & 100.0 \\
claude-opus-4.5 (thinking) & 100.0 & 100.0 & 100.0 & 100.0 & 100.0 & 83.2 & 100.0 & 100.0 & 84.2 & 83.6 & 86.2 & 81.2 & 100.0 & 100.0 & 100.0 \\
\bottomrule
\end{tabular}%
}
\end{sidewaystable}

\begin{sidewaystable}[htbp]
\centering
\caption{Model Performance on \scienceworld tasks, task set 2, models 27--51.}
\label{tab:sw_game_scores_2_2}
\resizebox{\textwidth}{!}{%
\begin{tabular}{lccccccccccccccc}
\toprule
Models & \rotatebox{285}{ScienceWorldInclinedPlaneFrictionNamedSurfaces} & \rotatebox{285}{ScienceWorldInclinedPlaneFrictionUnnamedSurfaces} & \rotatebox{285}{ScienceWorldLifespanLongestLived} & \rotatebox{285}{ScienceWorldLifespanLongestLivedThenShortestLived} & \rotatebox{285}{ScienceWorldLifespanShortestLived} & \rotatebox{285}{ScienceWorldMeasureMeltingPointKnownSubstance} & \rotatebox{285}{ScienceWorldMeasureMeltingPointUnknownSubstance} & \rotatebox{285}{ScienceWorldMelt} & \rotatebox{285}{ScienceWorldMendelianGeneticsKnownPlant} & \rotatebox{285}{ScienceWorldMendelianGeneticsUnknownPlant} & \rotatebox{285}{ScienceWorldPowerComponent} & \rotatebox{285}{ScienceWorldPowerComponentRenewableVsNonrenewableEnergy} & \rotatebox{285}{ScienceWorldTestConductivity} & \rotatebox{285}{ScienceWorldTestConductivityOfUnknownSubstances} & \rotatebox{285}{ScienceWorldUseThermometer} \\
\midrule
Qwen3.6-27B (1024) & 100.0 & 76.0 & 100.0 & 100.0 & 100.0 & 81.6 & 73.0 & 56.4 & 100.0 & 33.2 & 73.6 & 67.8 & 54.8 & 86.6 & 100.0 \\
o1 (high) & 100.0 & 100.0 & 100.0 & 100.0 & 100.0 & 100.0 & 100.0 & 100.0 & 60.4 & 100.0 & 69.0 & 23.6 & 100.0 & 100.0 & 84.4 \\
claude-sonnet-4.5 & 100.0 & 100.0 & 100.0 & 0.0 & 100.0 & 77.0 & 78.0 & 100.0 & 0.0 & 0.0 & 7.0 & 53.0 & 5.0 & 100.0 & 100.0 \\
gpt-4o & 100.0 & 66.0 & 100.0 & 100.0 & 100.0 & 100.0 & 72.2 & 3.0 & 100.0 & 82.2 & 58.2 & 38.0 & 91.0 & 50.6 & 84.4 \\
claude-3.5-haiku & 100.0 & 100.0 & 80.0 & 96.6 & 90.0 & 78.4 & 89.2 & 59.6 & 47.8 & 13.6 & 57.0 & 38.0 & 100.0 & 61.8 & 100.0 \\
Qwen3.6-35B-A3B (1024) & 100.0 & 70.0 & 100.0 & 100.0 & 100.0 & 100.0 & 100.0 & 24.4 & 100.0 & 65.0 & 11.0 & 20.6 & 16.2 & 17.2 & 100.0 \\
Llama-3.1-405B-Instruct & 81.0 & 100.0 & 100.0 & 100.0 & 100.0 & 100.0 & 9.0 & 4.0 & 83.2 & 100.0 & 33.0 & 28.8 & 17.0 & 58.8 & 37.6 \\
\bottomrule
\end{tabular}%
}
\end{sidewaystable}

\begin{sidewaystable}[htbp]
\centering
\caption{Model Performance on \scienceworld tasks, task set 2, models 52--76.}
\label{tab:sw_game_scores_2_3}
\resizebox{\textwidth}{!}{%
\begin{tabular}{lccccccccccccccc}
\toprule
Models & \rotatebox{285}{ScienceWorldInclinedPlaneFrictionNamedSurfaces} & \rotatebox{285}{ScienceWorldInclinedPlaneFrictionUnnamedSurfaces} & \rotatebox{285}{ScienceWorldLifespanLongestLived} & \rotatebox{285}{ScienceWorldLifespanLongestLivedThenShortestLived} & \rotatebox{285}{ScienceWorldLifespanShortestLived} & \rotatebox{285}{ScienceWorldMeasureMeltingPointKnownSubstance} & \rotatebox{285}{ScienceWorldMeasureMeltingPointUnknownSubstance} & \rotatebox{285}{ScienceWorldMelt} & \rotatebox{285}{ScienceWorldMendelianGeneticsKnownPlant} & \rotatebox{285}{ScienceWorldMendelianGeneticsUnknownPlant} & \rotatebox{285}{ScienceWorldPowerComponent} & \rotatebox{285}{ScienceWorldPowerComponentRenewableVsNonrenewableEnergy} & \rotatebox{285}{ScienceWorldTestConductivity} & \rotatebox{285}{ScienceWorldTestConductivityOfUnknownSubstances} & \rotatebox{285}{ScienceWorldUseThermometer} \\
\midrule
gpt-5-nano (thinking) & 3.0 & 5.0 & 90.0 & 53.0 & 70.0 & 80.2 & 31.0 & 2.0 & 1.0 & 1.0 & 58.2 & 31.4 & 81.0 & 8.0 & 18.2 \\
gpt-oss-20b (thinking) & 46.0 & 62.0 & 50.0 & 33.0 & 50.0 & 19.2 & 48.8 & 0.4 & 40.6 & 42.8 & 27.0 & 8.0 & 25.0 & 9.8 & 41.6 \\
Llama-4-Maverick-17B-128E-Instruct-FP8 & 0.0 & 5.0 & 50.0 & 33.0 & 50.0 & 6.0 & 1.0 & 1.2 & 0.0 & 0.0 & 7.0 & 38.0 & 5.0 & 10.0 & 0.0 \\
Mistral-Small-Instruct-2409 & 5.0 & 0.0 & 50.0 & 33.0 & 50.0 & 80.4 & 73.2 & 0.0 & 0.0 & 0.0 & 57.0 & 38.0 & 5.0 & 0.0 & 3.0 \\
DeepSeek-V4-Flash (1024) & 1.0 & 33.0 & 60.0 & 26.6 & 60.0 & 80.0 & 0.0 & 0.8 & 0.0 & 2.6 & 57.0 & 38.0 & 5.0 & 3.0 & 0.6 \\
Llama-3.1-8B-Instruct & 0.0 & 43.0 & 40.0 & 0.0 & 50.0 & 4.0 & 0.0 & 0.0 & 2.6 & 100.0 & 7.0 & 5.0 & 5.0 & 0.0 & 9.2 \\
DeepSeek-R1 (1024) & 0.0 & 0.0 & 50.0 & 100.0 & 50.0 & 24.8 & 1.0 & 0.0 & 0.0 & 11.0 & 7.0 & 5.0 & 5.0 & 10.0 & 8.8 \\
\bottomrule
\end{tabular}%
}
\end{sidewaystable}

\section{All Scores per Game: \jericho}\label{app:jer_all_scores}
Tables \ref{tab:jericho_game_scores_1_1}, \ref{tab:jericho_game_scores_1_2}, \ref{tab:jericho_game_scores_1_3}, \ref{tab:jericho_game_scores_2_1}, \ref{tab:jericho_game_scores_2_2}, \ref{tab:jericho_game_scores_2_3}, \ref{tab:jericho_game_scores_3_1}, \ref{tab:jericho_game_scores_3_2}, and \ref{tab:jericho_game_scores_3_3} show the per-game scores of every configuration in Table \ref{tab:all_model_framework_scores}, in the same order and with the same labels, in \jericho. \jericho games are split into three 18-game table sets for readability.

\begin{sidewaystable}[htbp]
\centering
\caption{Model Performance on \jericho games, game set 1, models 1--26.}
\label{tab:jericho_game_scores_1_1}
\resizebox{\textwidth}{!}{%
\begin{tabular}{lcccccccccccccccccc} %
\toprule
Models & \rotatebox{285}{905} & \rotatebox{285}{Acorncourt} & \rotatebox{285}{Advent} & \rotatebox{285}{Adventureland} & \rotatebox{285}{Afflicted} & \rotatebox{285}{Anchor} & \rotatebox{285}{Awaken} & \rotatebox{285}{Balances} & \rotatebox{285}{Ballyhoo} & \rotatebox{285}{Curses} & \rotatebox{285}{Cutthroat} & \rotatebox{285}{Deephome} & \rotatebox{285}{Detective} & \rotatebox{285}{Dragon} & \rotatebox{285}{Enchanter} & \rotatebox{285}{Enter} & \rotatebox{285}{Gold} & \rotatebox{285}{Hhgg} \\
\midrule
claude-fable-5 (thinking)$^{\ast\dagger}$ & 100.0 & 100.0 & 41.1 & 7.0 & 61.3 & 2.0 & 20.0 & 29.4 & 5.0 & 1.6 & 28.0 & 20.7 & 88.9 & 8.0 & 13.8 & 75.0 & 21.0 & 7.5 \\
claude-opus-5 (thinking)$^\ast$ & 100.0 & 100.0 & 33.8 & 18.2 & 57.3 & 3.6 & 14.0 & 31.4 & 5.0 & 1.5 & 13.6 & 12.7 & 62.2 & 4.0 & 13.2 & 34.0 & 13.2 & 6.5 \\
claude-opus-4.7 (thinking)$^\ast$ & 100.0 & 100.0 & 33.4 & 7.0 & 44.5 & 0.0 & 4.0 & 29.4 & 3.0 & 1.1 & 16.8 & 13.5 & 32.8 & 4.0 & 11.2 & 59.0 & 11.4 & 3.5 \\
gpt-5.4 (thinking) & 100.0 & 100.0 & 25.0 & 9.8 & 36.5 & 2.0 & 0.0 & 35.3 & 5.0 & 0.4 & 15.2 & 16.5 & 40.6 & 5.6 & 7.8 & 62.0 & 4.2 & 3.2 \\
claude-opus-4.6 (thinking)$^\ast$ & 100.0 & 73.3 & 28.5 & 4.2 & 49.9 & 1.2 & 6.0 & 31.4 & 4.0 & 0.7 & 20.0 & 19.9 & 48.9 & 0.0 & 15.5 & 51.0 & 6.0 & 2.5 \\
claude-opus-4.8 (thinking)$^\ast$ & 100.0 & 100.0 & 19.8 & 2.8 & 39.5 & 0.4 & 4.0 & 27.5 & 2.0 & 0.7 & 18.4 & 8.5 & 62.8 & 4.0 & 8.8 & 50.0 & 7.8 & 2.5 \\
claude-opus-4.5 (thinking) & 100.0 & 46.7 & 27.4 & 2.8 & 33.6 & 2.0 & 4.0 & 37.3 & 4.0 & 0.4 & 21.6 & 18.3 & 56.1 & 4.0 & 15.0 & 77.0 & 9.0 & 4.5 \\
\bottomrule
\end{tabular}%
}
\end{sidewaystable}

\begin{sidewaystable}[htbp]
\centering
\caption{Model Performance on \jericho games, game set 1, models 27--51.}
\label{tab:jericho_game_scores_1_2}
\resizebox{\textwidth}{!}{%
\begin{tabular}{lcccccccccccccccccc} %
\toprule
Models & \rotatebox{285}{905} & \rotatebox{285}{Acorncourt} & \rotatebox{285}{Advent} & \rotatebox{285}{Adventureland} & \rotatebox{285}{Afflicted} & \rotatebox{285}{Anchor} & \rotatebox{285}{Awaken} & \rotatebox{285}{Balances} & \rotatebox{285}{Ballyhoo} & \rotatebox{285}{Curses} & \rotatebox{285}{Cutthroat} & \rotatebox{285}{Deephome} & \rotatebox{285}{Detective} & \rotatebox{285}{Dragon} & \rotatebox{285}{Enchanter} & \rotatebox{285}{Enter} & \rotatebox{285}{Gold} & \rotatebox{285}{Hhgg} \\
\midrule
Qwen3.6-27B (1024) & 0.0 & 20.0 & 13.5 & 0.0 & 9.6 & 0.8 & 4.0 & 9.8 & 0.0 & 0.2 & 2.4 & 4.3 & 36.7 & 4.0 & 5.2 & 48.0 & 4.2 & 2.5 \\
o1 (high) & 80.0 & 66.7 & 13.9 & 1.4 & 1.1 & 0.0 & 0.0 & 23.9 & 0.0 & 0.0 & 9.6 & 6.7 & 27.2 & 4.8 & 12.5 & 22.0 & 0.0 & 2.5 \\
claude-sonnet-4.5 & 0.0 & 33.3 & 10.3 & 0.0 & 0.0 & 0.0 & 0.0 & 19.6 & 0.0 & 0.0 & 4.0 & 0.3 & 2.8 & 4.0 & 11.2 & 65.0 & 9.6 & 2.5 \\
gpt-4o & 0.0 & 0.0 & 10.3 & 0.0 & 1.1 & 0.4 & 0.0 & 13.7 & 1.0 & 0.0 & 0.0 & 6.3 & 34.4 & 7.2 & 4.2 & 38.0 & 7.8 & 2.5 \\
claude-3.5-haiku & 0.0 & 20.0 & 10.3 & 0.0 & 0.0 & 0.0 & 0.0 & 19.6 & 0.0 & 0.0 & 2.4 & 5.7 & 25.0 & 0.8 & 3.2 & 23.0 & 0.0 & 2.5 \\
Qwen3.6-35B-A3B (1024) & 0.0 & 0.0 & 10.3 & 0.0 & 0.0 & 0.0 & 0.0 & 0.0 & 0.0 & 0.0 & 2.4 & 4.9 & 19.4 & 1.6 & 0.0 & 17.0 & 0.0 & 0.0 \\
Llama-3.1-405B-Instruct & 0.0 & 0.0 & 11.7 & 0.0 & 4.5 & 0.0 & 0.0 & 19.6 & 0.0 & 0.0 & 5.6 & 5.6 & 50.0 & 4.0 & 4.0 & 15.0 & 0.0 & 1.0 \\
\bottomrule
\end{tabular}%
}
\end{sidewaystable}

\begin{sidewaystable}[htbp]
\centering
\caption{Model Performance on \jericho games, game set 1, models 52--76.}
\label{tab:jericho_game_scores_1_3}
\resizebox{\textwidth}{!}{%
\begin{tabular}{lcccccccccccccccccc} %
\toprule
Models & \rotatebox{285}{905} & \rotatebox{285}{Acorncourt} & \rotatebox{285}{Advent} & \rotatebox{285}{Adventureland} & \rotatebox{285}{Afflicted} & \rotatebox{285}{Anchor} & \rotatebox{285}{Awaken} & \rotatebox{285}{Balances} & \rotatebox{285}{Ballyhoo} & \rotatebox{285}{Curses} & \rotatebox{285}{Cutthroat} & \rotatebox{285}{Deephome} & \rotatebox{285}{Detective} & \rotatebox{285}{Dragon} & \rotatebox{285}{Enchanter} & \rotatebox{285}{Enter} & \rotatebox{285}{Gold} & \rotatebox{285}{Hhgg} \\
\midrule
gpt-5-nano (thinking) & 0.0 & 0.0 & 10.3 & 0.0 & 0.0 & 0.0 & 0.0 & 0.0 & 0.0 & 0.0 & 0.0 & 1.2 & 12.2 & 0.0 & 0.8 & 0.0 & 0.0 & 0.5 \\
gpt-oss-20b (thinking) & 0.0 & 0.0 & 10.3 & 0.0 & 0.0 & 0.0 & 0.0 & 5.9 & 0.0 & 0.0 & 0.0 & 2.6 & 7.8 & 2.4 & 0.0 & 7.0 & 0.0 & 0.0 \\
Llama-4-Maverick-17B-128E-Instruct-FP8 & 0.0 & 0.0 & 10.3 & 0.0 & 0.0 & 0.0 & 0.0 & 5.9 & 0.0 & 0.0 & 7.2 & 4.1 & 12.8 & 3.2 & 0.0 & 10.0 & 0.0 & 0.0 \\
Mistral-Small-Instruct-2409 & 0.0 & 0.0 & 10.3 & 0.0 & 0.0 & 0.0 & 0.0 & 0.0 & 0.0 & 0.0 & 0.0 & 4.0 & 8.3 & 0.0 & 5.0 & 0.0 & 0.0 & 0.0 \\
DeepSeek-V4-Flash (1024) & 40.0 & 0.0 & 10.3 & 0.0 & 0.5 & 0.0 & 0.0 & 5.9 & 0.0 & 0.1 & 0.0 & 2.3 & 33.9 & 3.2 & 2.2 & 6.0 & 3.0 & 1.0 \\
Llama-3.1-8B-Instruct & 0.0 & 0.0 & 10.3 & 0.0 & 0.5 & 0.0 & 0.0 & 15.7 & 0.0 & 0.0 & 0.8 & 4.3 & 15.6 & 0.0 & 0.0 & 2.0 & 0.0 & 0.0 \\
DeepSeek-R1 (1024) & 0.0 & 0.0 & 10.3 & 0.0 & 0.0 & 0.0 & 0.0 & 0.0 & 0.0 & 0.0 & 4.0 & 0.3 & 11.1 & 0.0 & 0.0 & 0.0 & 0.0 & 0.0 \\
\bottomrule
\end{tabular}%
}
\end{sidewaystable}

\begin{sidewaystable}[htbp]
\centering
\caption{Model Performance on \jericho games, game set 2, models 1--26.}
\label{tab:jericho_game_scores_2_1}
\resizebox{\textwidth}{!}{%
\begin{tabular}{lcccccccccccccccccc} %
\toprule
Models & \rotatebox{285}{Huntdark} & \rotatebox{285}{Infidel} & \rotatebox{285}{Inhumane} & \rotatebox{285}{Jewel} & \rotatebox{285}{Karn} & \rotatebox{285}{Library} & \rotatebox{285}{Loose} & \rotatebox{285}{Lostpig} & \rotatebox{285}{Ludicorp} & \rotatebox{285}{Lurking} & \rotatebox{285}{Moonlit} & \rotatebox{285}{Murdac} & \rotatebox{285}{Night} & \rotatebox{285}{Omniquest} & \rotatebox{285}{Partyfoul} & \rotatebox{285}{Pentari} & \rotatebox{285}{Planetfall} & \rotatebox{285}{Plundered} \\
\midrule
claude-fable-5 (thinking)$^{\ast\dagger}$ & 0.0 & 1.2 & 44.4 & 0.0 & 2.9 & 33.3 & 6.0 & 28.6 & 9.3 & 5.0 & 100.0 & 12.8 & 20.0 & 20.0 & 0.0 & 14.3 & 7.5 & 4.0 \\
claude-opus-5 (thinking)$^\ast$ & 0.0 & 1.2 & 53.3 & 1.3 & 4.1 & 53.3 & 6.0 & 42.9 & 12.0 & 6.0 & 100.0 & 9.3 & 20.0 & 26.0 & 0.0 & 20.0 & 7.5 & 10.4 \\
claude-opus-4.7 (thinking)$^\ast$ & 0.0 & 1.2 & 46.7 & 3.6 & 3.5 & 56.7 & 12.0 & 20.0 & 10.9 & 5.0 & 80.0 & 5.6 & 20.0 & 10.0 & 0.0 & 15.7 & 7.5 & 4.0 \\
gpt-5.4 (thinking) & 0.0 & 1.2 & 37.8 & 0.0 & 2.4 & 51.3 & 3.6 & 20.0 & 10.4 & 6.0 & 20.0 & 10.3 & 28.0 & 10.0 & 0.0 & 10.0 & 7.5 & 4.0 \\
claude-opus-4.6 (thinking)$^\ast$ & 0.0 & 1.2 & 48.9 & 2.7 & 3.5 & 39.3 & 3.2 & 28.6 & 8.7 & 5.0 & 60.0 & 6.6 & 16.0 & 10.0 & 0.0 & 10.0 & 7.5 & 8.8 \\
claude-opus-4.8 (thinking)$^\ast$ & 0.0 & 1.0 & 51.1 & 0.4 & 2.9 & 56.7 & 6.0 & 17.1 & 9.3 & 11.0 & 40.0 & 6.7 & 16.0 & 14.0 & 0.0 & 7.1 & 7.5 & 4.0 \\
claude-opus-4.5 (thinking) & 0.0 & 1.2 & 37.8 & 2.7 & 3.5 & 38.0 & 8.8 & 28.6 & 11.3 & 5.0 & 20.0 & 7.7 & 16.0 & 10.0 & 0.0 & 10.0 & 7.5 & 5.6 \\
\bottomrule
\end{tabular}%
}
\end{sidewaystable}

\begin{sidewaystable}[htbp]
\centering
\caption{Model Performance on \jericho games, game set 2, models 27--51.}
\label{tab:jericho_game_scores_2_2}
\resizebox{\textwidth}{!}{%
\begin{tabular}{lcccccccccccccccccc} %
\toprule
Models & \rotatebox{285}{Huntdark} & \rotatebox{285}{Infidel} & \rotatebox{285}{Inhumane} & \rotatebox{285}{Jewel} & \rotatebox{285}{Karn} & \rotatebox{285}{Library} & \rotatebox{285}{Loose} & \rotatebox{285}{Lostpig} & \rotatebox{285}{Ludicorp} & \rotatebox{285}{Lurking} & \rotatebox{285}{Moonlit} & \rotatebox{285}{Murdac} & \rotatebox{285}{Night} & \rotatebox{285}{Omniquest} & \rotatebox{285}{Partyfoul} & \rotatebox{285}{Pentari} & \rotatebox{285}{Planetfall} & \rotatebox{285}{Plundered} \\
\midrule
Qwen3.6-27B (1024) & 0.0 & 1.0 & 8.9 & 0.0 & 0.6 & 10.0 & 1.2 & 14.3 & 3.7 & 5.0 & 0.0 & 6.6 & 4.0 & 6.0 & 0.0 & 7.1 & 4.5 & 3.2 \\
o1 (high) & 0.0 & 1.0 & 4.4 & 0.0 & 1.8 & 34.7 & 0.0 & 14.3 & 7.9 & 7.0 & 0.0 & 8.6 & 0.0 & 12.0 & 0.0 & 8.6 & 3.8 & 4.0 \\
claude-sonnet-4.5 & 0.0 & 0.0 & 11.1 & 0.0 & 2.9 & 33.3 & 36.0 & 14.3 & 19.3 & 5.0 & 0.0 & 5.6 & 20.0 & 10.0 & 0.0 & 7.1 & 7.5 & 0.0 \\
gpt-4o & 0.0 & 1.2 & 4.4 & 0.0 & 1.2 & 13.3 & 0.8 & 5.7 & 0.8 & 5.0 & 0.0 & 5.6 & 0.0 & 10.0 & 0.0 & 8.6 & 7.5 & 3.2 \\
claude-3.5-haiku & 0.0 & 0.0 & 0.0 & 0.4 & 0.0 & 0.0 & 0.0 & 11.4 & 2.5 & 5.0 & 0.0 & 5.3 & 0.0 & 10.0 & 0.0 & 5.7 & 6.8 & 2.4 \\
Qwen3.6-35B-A3B (1024) & 0.0 & 0.5 & 0.0 & 0.0 & 0.0 & 0.0 & 0.0 & 0.0 & 0.7 & 5.0 & 0.0 & 2.2 & 0.0 & 8.0 & 0.0 & 2.9 & 2.2 & 0.8 \\
Llama-3.1-405B-Instruct & 0.0 & 1.2 & 0.0 & 0.0 & 0.0 & 20.0 & 0.0 & 14.3 & 2.5 & 5.0 & 0.0 & 5.6 & 20.0 & 10.0 & 0.0 & 7.1 & 3.8 & 0.0 \\
\bottomrule
\end{tabular}%
}
\end{sidewaystable}

\begin{sidewaystable}[htbp]
\centering
\caption{Model Performance on \jericho games, game set 2, models 52--76.}
\label{tab:jericho_game_scores_2_3}
\resizebox{\textwidth}{!}{%
\begin{tabular}{lcccccccccccccccccc} %
\toprule
Models & \rotatebox{285}{Huntdark} & \rotatebox{285}{Infidel} & \rotatebox{285}{Inhumane} & \rotatebox{285}{Jewel} & \rotatebox{285}{Karn} & \rotatebox{285}{Library} & \rotatebox{285}{Loose} & \rotatebox{285}{Lostpig} & \rotatebox{285}{Ludicorp} & \rotatebox{285}{Lurking} & \rotatebox{285}{Moonlit} & \rotatebox{285}{Murdac} & \rotatebox{285}{Night} & \rotatebox{285}{Omniquest} & \rotatebox{285}{Partyfoul} & \rotatebox{285}{Pentari} & \rotatebox{285}{Planetfall} & \rotatebox{285}{Plundered} \\
\midrule
gpt-5-nano (thinking) & 0.0 & 0.0 & 0.0 & 0.0 & 0.0 & 0.0 & 0.0 & 8.6 & 1.5 & 4.0 & 0.0 & 0.0 & 0.0 & 2.0 & 0.0 & 0.0 & 3.0 & 0.0 \\
gpt-oss-20b (thinking) & 0.0 & 0.0 & 0.0 & 0.0 & 0.0 & 0.0 & 0.0 & 11.4 & 1.3 & 0.0 & 0.0 & 3.1 & 0.0 & 8.0 & 0.0 & 0.0 & 2.2 & 0.0 \\
Llama-4-Maverick-17B-128E-Instruct-FP8 & 0.0 & 0.0 & 0.0 & 0.0 & 0.0 & 0.0 & 0.0 & 0.0 & 6.7 & 0.0 & 0.0 & 1.3 & 12.0 & 10.0 & 0.0 & 4.3 & 3.8 & 0.0 \\
Mistral-Small-Instruct-2409 & 0.0 & 0.0 & 0.0 & 0.0 & 0.0 & 0.0 & 0.0 & 14.3 & 0.7 & 0.0 & 0.0 & 0.0 & 0.0 & 10.0 & 0.0 & 7.1 & 3.8 & 0.0 \\
DeepSeek-V4-Flash (1024) & 0.0 & 0.0 & 4.4 & 0.0 & 0.0 & 30.0 & 0.0 & 14.3 & 0.8 & 5.0 & 0.0 & 1.1 & 12.0 & 10.0 & 0.0 & 0.0 & 0.0 & 0.8 \\
Llama-3.1-8B-Instruct & 0.0 & 0.0 & 0.0 & 0.0 & 0.0 & 0.0 & 0.0 & 8.6 & 0.7 & 3.0 & 0.0 & 3.4 & 0.0 & 10.0 & 0.0 & 0.0 & 3.0 & 0.0 \\
DeepSeek-R1 (1024) & 0.0 & 0.0 & 0.0 & 0.0 & 0.0 & 0.0 & 0.0 & 2.9 & 1.3 & 0.0 & 0.0 & 0.0 & 0.0 & 10.0 & 0.0 & 0.0 & 0.0 & 0.0 \\
\bottomrule
\end{tabular}%
}
\end{sidewaystable}

\begin{sidewaystable}[htbp]
\centering
\caption{Model Performance on \jericho games, game set 3, models 1--26.}
\label{tab:jericho_game_scores_3_1}
\resizebox{\textwidth}{!}{%
\begin{tabular}{lcccccccccccccccccc} %
\toprule
Models & \rotatebox{285}{Reverb} & \rotatebox{285}{Seastalker} & \rotatebox{285}{Sherlock} & \rotatebox{285}{Snacktime} & \rotatebox{285}{Sorcerer} & \rotatebox{285}{Spellbrkr} & \rotatebox{285}{Spirit} & \rotatebox{285}{Temple} & \rotatebox{285}{Trinity} & \rotatebox{285}{Tryst205} & \rotatebox{285}{Weapon} & \rotatebox{285}{Wishbringer} & \rotatebox{285}{Yomomma} & \rotatebox{285}{Zenon} & \rotatebox{285}{Zork1} & \rotatebox{285}{Zork2} & \rotatebox{285}{Zork3} & \rotatebox{285}{Ztuu} \\
\midrule
claude-fable-5 (thinking)$^{\ast\dagger}$ & 50.0 & 30.0 & 10.0 & 40.0 & 5.0 & 11.7 & 0.8 & 14.3 & 15.0 & 2.9 & 0.0 & 10.0 & 2.9 & 0.0 & 24.3 & 5.0 & 57.1 & 10.0 \\
claude-opus-5 (thinking)$^\ast$ & 44.0 & 19.0 & 12.0 & 64.0 & 10.0 & 11.7 & 1.3 & 21.1 & 13.0 & 3.1 & 0.0 & 19.6 & 2.9 & 0.0 & 32.2 & 0.0 & 42.9 & 6.0 \\
claude-opus-4.7 (thinking)$^\ast$ & 15.6 & 20.0 & 10.8 & 64.0 & 9.0 & 8.5 & 1.3 & 21.1 & 8.8 & 3.7 & 0.0 & 14.0 & 0.0 & 0.0 & 14.5 & 3.2 & 48.6 & 6.0 \\
gpt-5.4 (thinking) & 9.6 & 13.4 & 8.4 & 52.0 & 9.5 & 11.3 & 1.3 & 16.0 & 15.0 & 2.0 & 0.0 & 15.2 & 1.7 & 0.0 & 13.1 & 2.0 & 45.7 & 4.0 \\
claude-opus-4.6 (thinking)$^\ast$ & 20.8 & 16.0 & 10.2 & 48.0 & 10.0 & 11.3 & 0.8 & 8.6 & 15.0 & 3.4 & 0.0 & 18.6 & 1.7 & 0.0 & 20.6 & 3.0 & 42.9 & 5.0 \\
claude-opus-4.8 (thinking)$^\ast$ & 22.4 & 15.0 & 11.0 & 36.0 & 8.0 & 10.3 & 1.0 & 16.0 & 11.6 & 1.7 & 0.0 & 19.0 & 0.0 & 0.0 & 17.9 & 1.0 & 45.7 & 8.0 \\
claude-opus-4.5 (thinking) & 20.8 & 15.4 & 10.4 & 40.0 & 8.0 & 11.7 & 1.0 & 19.4 & 14.8 & 2.3 & 0.0 & 14.6 & 2.3 & 0.0 & 14.8 & 2.0 & 42.9 & 11.4 \\
\bottomrule
\end{tabular}%
}
\end{sidewaystable}

\begin{sidewaystable}[htbp]
\centering
\caption{Model Performance on \jericho games, game set 3, models 27--51.}
\label{tab:jericho_game_scores_3_2}
\resizebox{\textwidth}{!}{%
\begin{tabular}{lcccccccccccccccccc} %
\toprule
Models & \rotatebox{285}{Reverb} & \rotatebox{285}{Seastalker} & \rotatebox{285}{Sherlock} & \rotatebox{285}{Snacktime} & \rotatebox{285}{Sorcerer} & \rotatebox{285}{Spellbrkr} & \rotatebox{285}{Spirit} & \rotatebox{285}{Temple} & \rotatebox{285}{Trinity} & \rotatebox{285}{Tryst205} & \rotatebox{285}{Weapon} & \rotatebox{285}{Wishbringer} & \rotatebox{285}{Yomomma} & \rotatebox{285}{Zenon} & \rotatebox{285}{Zork1} & \rotatebox{285}{Zork2} & \rotatebox{285}{Zork3} & \rotatebox{285}{Ztuu} \\
\midrule
Qwen3.6-27B (1024) & 0.0 & 6.8 & 1.6 & 28.0 & 1.2 & 5.0 & 1.4 & 0.0 & 7.6 & 0.0 & 0.0 & 8.8 & 0.0 & 0.0 & 8.9 & 0.0 & 22.9 & 0.0 \\
o1 (high) & 8.8 & 3.8 & 0.4 & 40.0 & 7.8 & 9.0 & 1.1 & 11.4 & 9.4 & 1.7 & 0.0 & 13.8 & 0.6 & 0.0 & 12.5 & 1.0 & 42.9 & 9.6 \\
claude-sonnet-4.5 & 0.0 & 4.0 & 2.0 & 0.0 & 1.2 & 8.3 & 0.8 & 14.3 & 10.0 & 0.0 & 0.0 & 15.0 & 2.9 & 0.0 & 15.4 & 0.0 & 57.1 & 0.0 \\
gpt-4o & 0.0 & 3.2 & 8.8 & 20.0 & 3.2 & 8.0 & 0.8 & 0.0 & 9.6 & 0.9 & 0.0 & 7.4 & 0.0 & 0.0 & 14.5 & 0.0 & 31.4 & 0.0 \\
claude-3.5-haiku & 14.0 & 10.8 & 0.0 & 8.0 & 2.0 & 4.2 & 0.8 & 0.0 & 9.0 & 0.0 & 0.0 & 11.8 & 1.7 & 0.0 & 13.1 & 0.0 & 31.4 & 0.0 \\
Qwen3.6-35B-A3B (1024) & 0.0 & 21.0 & 1.6 & 0.0 & 2.0 & 3.3 & 0.5 & 2.9 & 1.6 & 0.0 & 0.0 & 6.0 & 0.0 & 0.0 & 2.3 & 0.0 & 34.3 & 0.0 \\
Llama-3.1-405B-Instruct & 0.0 & 10.6 & 2.6 & 24.0 & 1.2 & 4.2 & 1.3 & 14.3 & 4.0 & 0.0 & 0.0 & 6.0 & 0.0 & 0.0 & 11.1 & 0.0 & 28.6 & 13.0 \\
\bottomrule
\end{tabular}%
}
\end{sidewaystable}

\begin{sidewaystable}[htbp]
\centering
\caption{Model Performance on \jericho games, game set 3, models 52--76.}
\label{tab:jericho_game_scores_3_3}
\resizebox{\textwidth}{!}{%
\begin{tabular}{lcccccccccccccccccc} %
\toprule
Models & \rotatebox{285}{Reverb} & \rotatebox{285}{Seastalker} & \rotatebox{285}{Sherlock} & \rotatebox{285}{Snacktime} & \rotatebox{285}{Sorcerer} & \rotatebox{285}{Spellbrkr} & \rotatebox{285}{Spirit} & \rotatebox{285}{Temple} & \rotatebox{285}{Trinity} & \rotatebox{285}{Tryst205} & \rotatebox{285}{Weapon} & \rotatebox{285}{Wishbringer} & \rotatebox{285}{Yomomma} & \rotatebox{285}{Zenon} & \rotatebox{285}{Zork1} & \rotatebox{285}{Zork2} & \rotatebox{285}{Zork3} & \rotatebox{285}{Ztuu} \\
\midrule
gpt-5-nano (thinking) & 0.0 & 2.6 & 1.4 & 0.0 & 1.2 & 5.2 & 0.2 & 0.0 & 3.8 & 0.0 & 0.0 & 6.0 & 0.0 & 0.0 & 0.6 & 0.0 & 28.6 & 0.0 \\
gpt-oss-20b (thinking) & 0.0 & 2.6 & 0.0 & 0.0 & 1.2 & 1.0 & 0.0 & 0.0 & 2.4 & 0.0 & 0.0 & 6.0 & 0.0 & 0.0 & 2.0 & 0.0 & 11.4 & 0.0 \\
Llama-4-Maverick-17B-128E-Instruct-FP8 & 0.0 & 3.0 & 1.2 & 0.0 & 1.2 & 4.7 & 0.0 & 0.0 & 0.6 & 0.0 & 0.0 & 6.0 & 0.0 & 0.0 & 2.0 & 0.0 & 0.0 & 0.0 \\
Mistral-Small-Instruct-2409 & 0.0 & 2.0 & 0.0 & 0.0 & 1.2 & 0.0 & 0.0 & 0.0 & 1.0 & 0.0 & 0.0 & 6.0 & 0.0 & 0.0 & 1.4 & 0.0 & 0.0 & 0.0 \\
DeepSeek-V4-Flash (1024) & 0.0 & 6.6 & 4.0 & 8.0 & 3.2 & 0.8 & 0.6 & 8.6 & 0.6 & 0.0 & 0.0 & 6.0 & 0.0 & 0.0 & 6.0 & 0.0 & 17.1 & 2.0 \\
Llama-3.1-8B-Instruct & 0.0 & 3.0 & 0.0 & 0.0 & 1.2 & 3.3 & 0.2 & 8.6 & 1.0 & 0.0 & 0.0 & 6.0 & 0.0 & 0.0 & 2.0 & 0.0 & 17.1 & 2.0 \\
DeepSeek-R1 (1024) & 0.0 & 3.0 & 0.0 & 0.0 & 1.2 & 0.0 & 0.0 & 0.0 & 0.0 & 0.0 & 0.0 & 6.0 & 0.0 & 0.0 & 2.9 & 0.0 & 0.0 & 0.0 \\
\bottomrule
\end{tabular}%
}
\end{sidewaystable}

\clearpage
\section{\jericho Scores per Game at 500 Steps}\label{app:frontier_pergame}
Table~\ref{tab:frontier_pergame} shows the per-game \jericho scores of \astra and \opusfivefive at 500 steps (Section~\ref{sec:frontier}), averaged over 5 seeds, on the same 54 games as Appendix~\ref{app:jer_all_scores}.
Both models run with maximum reasoning effort, the full interaction history in context, and save, load, restore, and undo commands blocked; the prompt does not state the step budget.
\astra submits actions through a structured action tool, while \opusfivefive plays with bare-text actions.

\begin{table}[h!]
\centering
\small
\resizebox{\textwidth}{!}{%
\begin{tabular}{lcc@{\hskip 1.5em}lcc@{\hskip 1.5em}lcc}
\toprule
Game & \astra & \opusfivefive & Game & \astra & \opusfivefive & Game & \astra & \opusfivefive \\
\midrule
905 & 100.0 & 100.0 & Huntdark & 100.0 & 20.0 & Reverb & 100.0 & 72.4 \\
Acorncourt & 100.0 & 100.0 & Infidel & 59.8 & 6.0 & Seastalker & 82.0 & 57.2 \\
Advent & 89.1 & 99.6 & Inhumane & 20.0 & 22.2 & Sherlock & 30.6 & 25.8 \\
Adventureland & 64.4 & 76.4 & Jewel & 31.1 & 8.0 & Snacktime & 100.0 & 100.0 \\
Afflicted & 99.1 & 95.7 & Karn & 23.5 & 21.8 & Sorcerer & 10.0 & 11.0 \\
Anchor & 22.0 & 17.8 & Library & 92.7 & 80.7 & Spellbrkr & 11.7 & 11.7 \\
Awaken & 100.0 & 82.0 & Loose & 90.0 & 35.6 & Spirit & 4.1 & 6.0 \\
Balances & 66.7 & 69.0 & Lostpig & 100.0 & 91.4 & Temple & 100.0 & 88.6 \\
Ballyhoo & 35.0 & 23.0 & Ludicorp & 21.6 & 34.5 & Trinity & 26.6 & 30.4 \\
Curses & 7.3 & 4.3 & Lurking & 37.0 & 34.0 & Tryst205 & 33.1 & 16.9 \\
Cutthroat & 47.2 & 32.8 & Moonlit & 100.0 & 100.0 & Weapon & 100.0 & 100.0 \\
Deephome & 49.4 & 47.3 & Murdac & 16.1 & 14.1 & Wishbringer & 66.0 & 50.6 \\
Detective & 100.0 & 100.0 & Night & 100.0 & 100.0 & Yomomma & 97.1 & 97.1 \\
Dragon & 97.6 & 87.2 & Omniquest & 100.0 & 42.0 & Zenon & 100.0 & 96.0 \\
Enchanter & 57.8 & 46.2 & Partyfoul & 100.0 & 100.0 & Zork1 & 85.7 & 100.0 \\
Enter & 100.0 & 100.0 & Pentari & 100.0 & 41.4 & Zork2 & 56.9 & 41.0 \\
Gold & 43.2 & 37.2 & Planetfall & 44.0 & 48.8 & Zork3 & 100.0 & 91.4 \\
Hhgg & 39.5 & 27.5 & Plundered & 76.0 & 59.2 & Ztuu & 80.2 & 75.0 \\
\midrule
\multicolumn{7}{l}{Average over all 54 games} & 66.9 & 57.0 \\
\bottomrule
\end{tabular}%
}
\caption{Per-game \jericho scores at 500 steps, averaged over 5 seeds. Appendix~\ref{app:walkthroughs} gives the walkthrough score of each game at 500 steps.}
\label{tab:frontier_pergame}
\end{table}

\clearpage
\section{Extending \ours Beyond \jericho}\label{app:next_jericho}
Section~\ref{sec:frontier} argues that if model-level performance on the minimal-harness floor saturates, \jericho keeps its role for the broader model distribution.
Contamination, however, puts every human-authored game on a timer: a game that is obscure today enters the next training set once its transcripts are published.
Break-In is a less documented game outside \ours. In a 500-step run with the settings of Section~\ref{sec:frontier} and the game's built-in hints disabled, \astra's reasoning at step 177 connects a red hat to the cockerel guarding a chicken coop (``a rooster has a red comb that stands out''); it puts on the hat at step 217, and at step 231 the cockerel looks closely at its head and lets it into the coop.
That is the abstraction \jericho was chosen to measure, and recalling a walkthrough cannot show it.
The way off this treadmill is games that did not exist until the evaluation did: newly authored or generated games, with one model writing, another playing, and a human in between.
Such games are admitted by the same gymnasium interface, and under our versioning scheme they would not change scores on the current 122-task set.

\clearpage
\section{Average \simon Score versus Overall \ours Score}\label{simonsaysvtales}
\begin{figure}[h]
\centering
 \includegraphics[width=\textwidth]{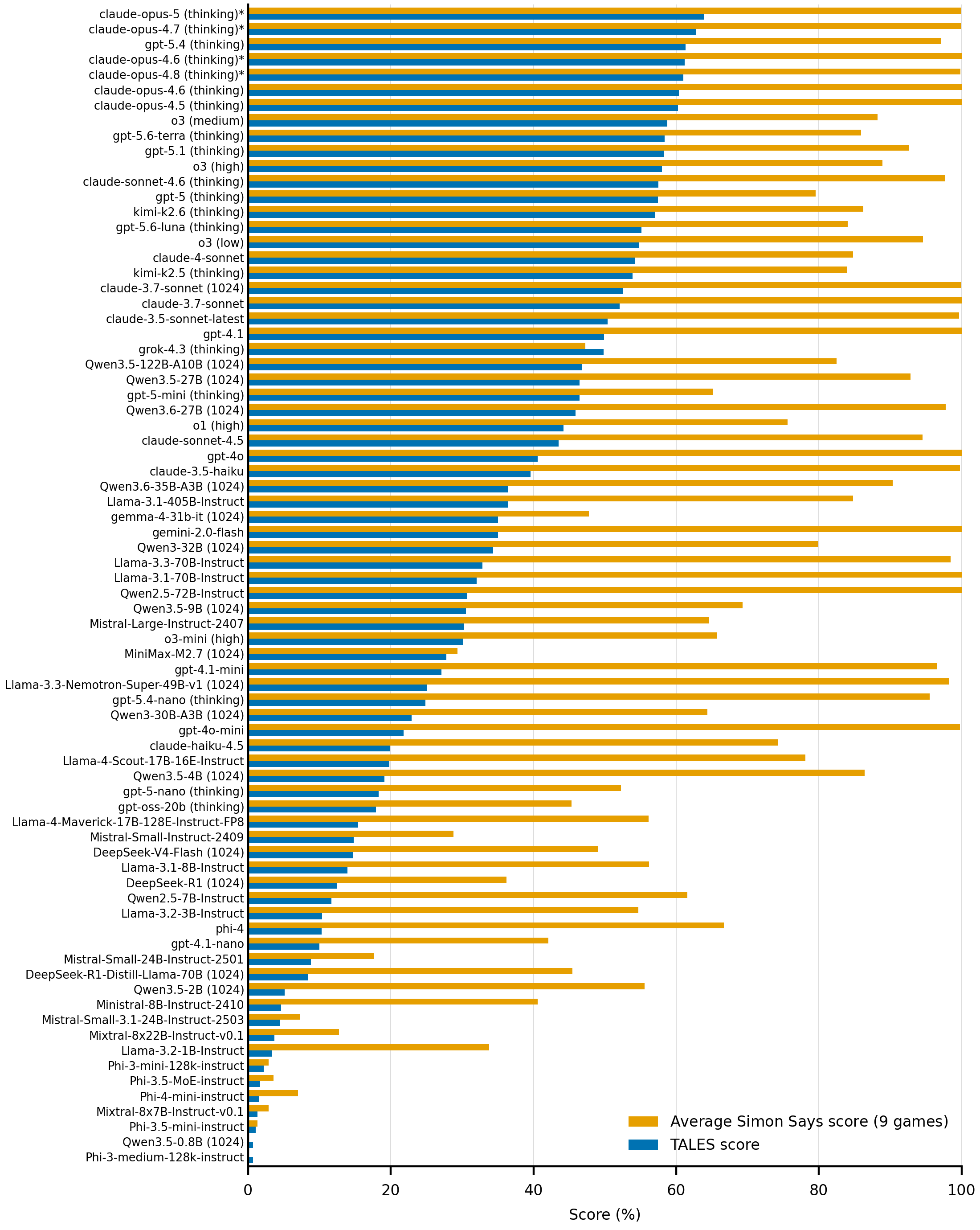}
\caption{All \ours and average \simon scores for each model, sorted by \ours performance. We see that an increase in performance in \simon typically correlates with an increase in performance for \ours overall. $^\ast$Action-tool runs; both claude-opus-4.6 runs are shown, and claude-fable-5 (one seed) is omitted.
}
\end{figure}

\section{Compute}\label{app:resources}
Experiments were run between November 2024 and October 2026 on a mix of A100, H100, and B200 clusters, with compute credits provided by Microsoft.
We estimate roughly 9,000 hours of total run time, 23 billion tokens, and \$20,000 in API costs at list prices across all runs.

\section{To Think or Not to Think? A Claude Case Study}\label{app:claude_thinking}
Due to performance and availability of the thinking traces, we analyze the game transcripts and thinking traces of Claude-3.7-Sonnet, comparing the model's performance with and without the ability to think.
\citep{whitelivebench} reports a significant increase in performance for Claude-3.7-Sonnet with thinking versus without.
Despite this, we see an increase of less than 1\% in \ours score for the thinking versus non-thinking modes of Claude-3.7-Sonnet. 

\textbf{Thinking scaffolds different competencies at different stages.}
We find Claude-3.7-Sonnet often explicitly exhibits one or more agentic competencies in its thinking traces throughout the progression of the game.
Early, the primary competency shown is often Strategic Action when identifying relevant game elements to interact with or areas to explore.
Later steps mainly focus on Rule Learning to synthesize implicit knowledge gained from exploring the environment or Situated Awareness in explicitly verbalizing the current state of the agent in the environment. 

\textbf{Thinking LLMs still fail when required to integrate multiple agentic competencies simultaneously.}
Although Claude-3.7 explicitly verbalized all agentic competencies within its thinking trace at various points throughout the games, failures still occurred when multiple competencies were required at key steps.
Similar to the example shown in Figure \ref{fig:grounded_reasoning_failure}, we see simultaneously one competency on display while another is failing.
In some cases when the missed detail leads to an immediate error, the LLM is capable of self-correcting. 
However, errors occurring later in the horizon often result in the agent being unable to identify and backtrack to the original point of failure. 

\section{Reasoning Failures}\label{sec:labeled_subset}
In this section, we provide explicit reasoning failure numbers from a smaller subset of the top performing LLMs at the time of the experiment, similar to what is shown in Figure \ref{fig:grounded_reasoning_failure}.
We focus again on the Claude family of models due to having access to a range of model performances as well as both thinking and non-thinking modes for their top model.
We outline the specific criteria we used to classify actions as a specific type of reasoning failure.
From a selection of \numlabeledlogs game logs of the most difficult games in the frameworks discussed in Section \ref{sec:quant}, we exhaustively label all actions that exhibit some reasoning failure, discuss our criteria for designating a chosen action as a reasoning failure, and present the cumulative average failures per LLM, finding that \textbf{inductive reasoning failures} where the agent fails to account for explicit or implicit feedback are by far the most common.

\begin{table}[h!]
  \caption{Average reasoning failures for the Claude family of models across \numlabeledgames games (\numlabeledlogs logs total).}
  \centering
  \resizebox{\textwidth}{!}{
  \begin{tabular}{l|ccccc|c}
    \toprule
     & \multicolumn{5}{c|}{Failures($\downarrow$)} & \\
    Model & Spatial & Deductive & Inductive & Grounded & Total & \ours Score ($\uparrow$) \\
    \midrule
     3.7-Sonnet (Thinking) & 1.29 & 1.29 & 2.14 & 1.0 & 5.71 & 52.5 \\
     3.7-Sonnet & 0.86 & 3.57 & 5.43 & 1.0 & 10.86 & 52.1 \\
     3.5-Sonnet & 1.14 & 1.43 & 9.43 & 0.86 & 12.86 & 50.4 \\
     3.5-Haiku  & 3.86 & 7.14 & 25.29 & 2.86 & 39.14 & 39.6 \\
    \bottomrule
  \end{tabular}
  }
  \label{table:labeled_failures}
\end{table}

As shown in Table~\ref{table:labeled_failures}, \textbf{top LLMs still make reasoning errors but resolve them quickly enough to still succeed.}
When explicitly labeling and enumerating reasoning errors made on a per-action basis, we see a clear trend with the lower-performing LLMs also making a larger number of reasoning errors.
These failures ultimately waste a large portion of the weaker LLMs' allotted steps for the environments, leading to a decreased performance.
The stronger LLMs, while still making reasoning errors, make few enough errors to still have sufficient time to complete the tasks.

Claude-3.5-Haiku frequently encounters inductive reasoning failures as weaker models tend to get trapped in cyclical action loops that do not result in any meaningful change in state.
We consider each repeated step in this cycle as a distinct failure because each step represents an instance where the agent fails to incorporate feedback.
See Appendix \ref{app:annotated} for our labeling criteria and examples.

\section{1000 Steps of \zork}\label{app:1000zork}
While even the best performing LLMs make reasoning mistakes, we find what allows them to still find success in \ours is the ability to both avoid making an excessive number of these mistakes and the ability to self-correct.
We argue that 100 steps are sufficient to evaluate the performance of current state-of-the-art LLMs because even the best overall LLM at the time of the experiment, o3, fails to approach the walkthrough score within 100 steps for \zork (13.3\% vs 29.1\%).
In this experiment, we explore whether any of the top models can achieve a score comparable to the walkthrough after 100 steps in \zork while allowed to run for 1000 steps while keeping the entire history in the context. If so, we examine the behaviors that enable this and determine the required number of steps.
We select the overall top 3 performing models at the time of the experiment, o3 and both the thinking and non-thinking modes of Claude-3.7-Sonnet.

\textbf{Scores improve slightly, but the best LLMs are still far from the walkthrough score even with 10 times more steps.}
Even with 1000 steps, the best LLMs fail to reach 29.1\% of the total score. 
o3 manages to achieve a score of 20.9\%, a performance increase of only 7.6 percentage points over its original score of 13.3\% for 10 times the steps.
Claude-3.7 non-thinking\footnote{Despite scoring higher than its thinking variant, non-thinking Claude-3.7-Sonnet suffers a catastrophic inductive reasoning failure by repeatedly issuing the quit command after step 479.} and thinking achieve 16.9\% and 15.3\% respectively. 
The key behavior pattern we see in both thinking models is a slightly directed, random exploration of the area of the game past the bottleneck that stops other, weaker LLMs.
This exploration is far less focused than the iterative search we see agents perform early in the game and in simpler environments such as \abbralfworld.

We tried allowing Claude-3.7-Sonnet to think for up to 4096 tokens.
However, the model never uses more than 700 tokens for its thinking, a similar value seen for the rest of the benchmark where the thinking effort is capped to 1024.
This is a significantly smaller thinking effort than o3 which uses up to 5000 thinking tokens throughout its 1000-step \zork playthrough.
This suggests that o3's performance is due to a willingness to leverage many more thinking tokens at any particular step.
However, the highest thinking efforts do not appear to occur at any significant points during gameplay and we are unable to verify the actual contents since we do not have access to the thinking traces.

\section{RL Fine-tuning Hyperparameters}\label{app:hyperparameters}
We use Qwen3 rather than Qwen2.5 due to its inherent reasoning training.
Following guidance in the appendix of the verl-agent paper~\citep{feng2025group}, we use a learning rate of 1e-6 for the actor and 1e-5 for the critic. 
We train over a total of 150 \textit{epochs}, where verl-agent defines an epoch as one full rollout of the environment.
To more closely adhere to our zero-shot evaluations, we keep the entire history in the trajectory up to 50 steps. This necessitates a far smaller batch size of 8. During evaluation, we use a batch size of 32 to sufficiently sample all evaluation environments.
We use the collection of all existing game seeds and formats not evaluated on in our zero-shot experiments as a training set and the specific environments from the zero-shot experiments as the test set.
We allocate a thinking budget of 512 tokens.
In their \alfworld implementation, \citep{feng2025group} use a terminal reward of 10.
In our RL experiments, we use the native reward from the environment, as our motivation is not to contest \citep{feng2025group}'s results but to evaluate the effects of including privileged or domain knowledge directly in the prompt.

\section{Framework Details}\label{app:framework_details}
This section gives additional detail on the frameworks described in Section~\ref{sec:tales}.

\textbf{\simon.}
\simon serves as a diagnostic entry point to \ours as it requires minimal multi-turn competency in extended horizons for completion.
All frameworks included in \ours require the LLM to have minimal capabilities as an agent to make any non-trivial progress.

\textbf{\textworld.}
In each CookingWorld game, the task involves following a recipe that requires finding ingredients and processing them according to said recipe.
We selected one game per difficulty ranging from level 1 (with one location and a recipe of 1 ingredient) to level 10 (having 12 locations and a recipe with 3 ingredients).
The player receives 1 point after completing sub-goals related to the task in the game.

\textbf{\alfworld.}
Where other environments have a predefined error message relating to the type of error, whether it is due to the parser not recognizing the command or the action not being possible, \abbralfworld has only one error message: \cmd{Nothing happens}.
In the original \abbralfworld framework, the visual component compensates for the lack of detailed text feedback.
However, this makes it significantly harder for agents relying solely on text-based interactions, compounded by the limitation that in \alfworld you can hold one object at a time.

\section{Framework Environment Subselection}\label{app:env_subselection}
\subsection{\textworld}
For \textworld, we use the following environments:

\begin{Verbatim}[breaklines=true, breakanywhere=true]
test/difficulty_level_1/tw-cooking-recipe1+take1+open-0nQyHWbvh6dXFPmhLKX.z8
test/difficulty_level_2/tw-cooking-recipe1+take1+cook+open-0nQyHWbvh6dXFPmhLKX.z8
test/difficulty_level_3/tw-cooking-recipe1+take1+cut+open-0nQyHWbvh6dXFPmhLKX.z8
test/difficulty_level_4/tw-cooking-recipe1+take1+open+go6-0nQyHWbvh6dXFPmhLKX.z8
test/difficulty_level_5/tw-cooking-recipe1+take1+open+go9-0nQyHWbvh6dXFPmhLKX.z8
test/difficulty_level_6/tw-cooking-recipe1+take1+open+go12-0nQyHWbvh6dXFPmhLKX.z8
test/difficulty_level_7/tw-cooking-recipe1+take1+cook+cut+open-0nQyHWbvh6dXFPmhLKX.z8
test/difficulty_level_8/tw-cooking-recipe3+take3+open+go6-0nQyHWbvh6dXFPmhLKX.z8
test/difficulty_level_9/tw-cooking-recipe3+take3+cook+cut+open+go6-0nQyHWbvh6dXFPmhLKX.z8
test/difficulty_level_10/tw-cooking-recipe3+take3+cook+cut+open+go12-0nQyHWbvh6dXFPmhLKX.z8
\end{Verbatim}

\subsection{\twx}
For \twx, we use the game parameters:
\begin{Verbatim}[breaklines=true, breakanywhere=true]
TASKS = [
    (
        "CookingWorld",
        "cookingworld",
        "numLocations=1, numIngredients=2, numDistractorItems=5, includeDoors=0, limitInventorySize=0",
    ),
    (
        "TextWorldCommonsense",
        "twc",
        "numLocations=1,numItemsToPutAway=1,includeDoors=0,limitInventorySize=0",
    ),
    (
        "CoinCollector",
        "coin",
        "numLocations=1, numDistractorItems=5, limitInventorySize=0",
    ),
    ("Arithmetic", "arithmetic", ""),
    (
        "MapReader",
        "mapreader",
        "numLocations=2, maxDistanceApart=1, maxDistractorItemsPerLocation=2, includeDoors=0, limitInventorySize=0",
    ),
    ("Sorting", "sorting", ""),
    ("SimonSays10", "simonsays", "gameLength=10, numDistractors=4, memorization=0"),
    ("SimonSays50", "simonsays", "gameLength=50, numDistractors=4, memorization=0"),
    ("SimonSays100", "simonsays", "gameLength=100, numDistractors=4, memorization=0"),
    (
        "SimonSaysWithMemory10",
        "simonsays",
        "gameLength=10, numDistractors=4, memorization=1, verbose=0",
    ),
    (
        "SimonSaysWithMemory50",
        "simonsays",
        "gameLength=50, numDistractors=4, memorization=1, verbose=0",
    ),
    (
        "SimonSaysWithMemory100",
        "simonsays",
        "gameLength=100, numDistractors=4, memorization=1, verbose=0",
    ),
    (
        "SimonSaysWithMemory10Verbose",
        "simonsays",
        "gameLength=10, numDistractors=4, memorization=1, verbose=1",
    ),
    (
        "SimonSaysWithMemory50Verbose",
        "simonsays",
        "gameLength=50, numDistractors=4, memorization=1, verbose=1",
    ),
    (
        "SimonSaysWithMemory100Verbose",
        "simonsays",
        "gameLength=100, numDistractors=4, memorization=1, verbose=1",
    ),
    ("PeckingOrder", "peckingorder", ""),
]
\end{Verbatim}

\subsection{\alfworld}
We use the following 12 games for \alfworld. Note that these are the games used when \texttt{-{}-game-seed} is not set. Changing this value would cause the games to change.
\begin{Verbatim}[breaklines=true, breakanywhere=true]
valid_seen/pick_and_place_simple-Book-None-SideTable-329/trial_T20190908_050633_745514
valid_seen/look_at_obj_in_light-AlarmClock-None-DeskLamp-323/trial_T20190909_044715_250790
valid_seen/pick_clean_then_place_in_recep-ButterKnife-None-CounterTop-8/trial_T20190909_105559_983897
valid_seen/pick_heat_then_place_in_recep-Apple-None-DiningTable-26/trial_T20190907_060234_011675
valid_seen/pick_cool_then_place_in_recep-Apple-None-CounterTop-14/trial_T20190909_044933_815840
valid_seen/pick_two_obj_and_place-AlarmClock-None-Dresser-305/trial_T20190907_165826_194855
valid_unseen/pick_and_place_simple-Mug-None-Desk-308/trial_T20190908_125200_737896
valid_unseen/look_at_obj_in_light-AlarmClock-None-DeskLamp-308/trial_T20190908_222917_366542
valid_unseen/pick_clean_then_place_in_recep-Bowl-None-Cabinet-10/trial_T20190909_061130_844814
valid_unseen/pick_heat_then_place_in_recep-Apple-None-Fridge-10/trial_T20190906_182259_116320
valid_unseen/pick_cool_then_place_in_recep-Bread-None-CounterTop-10/trial_T20190908_091747_866951
valid_unseen/pick_two_obj_and_place-CD-None-Safe-308/trial_T20190907_050942_897916
\end{Verbatim}

\subsection{\scienceworld}
We use the first variation of the test set for each of the 30 tasks.

\section{Annotated Logs: Criteria and Examples}\label{app:annotated}
\subsection{Labeling Criteria}
For spatial reasoning failures, we primarily label those actions that involve failing navigation through path finding,  backtracking, or iteratively searching spaces for necessary elements for completing the assigned task.
Deductive reasoning failures were primarily actions that involved interaction with game elements that cannot be reasonably linked back to the target objective or failures to solve in-game puzzles.
Inductive reasoning failures consisted mostly of actions that the agent had previously attempted in a near-identical state to negative feedback.
If there was any non-trivial change in game state compared to when the action was first attempted, we did not label the action as an inductive reasoning failure if the action did not fail due to a parser issue.
Grounded reasoning failures encompassed what is commonly referred to as hallucination: attempts to interact with game elements not mentioned in observations or feedback.

\subsection{Spatial Reasoning Failure Example}
\begin{Verbatim}[breaklines=true, breakanywhere=true]
Task Description:Your task is to melt lead. First, focus on the substance. Then, take actions that will cause it to change its state of matter...

```
This room is called the kitchen. In it, you see: 
	the agent
	a substance called air
	a chair. On the chair is: nothing.
	a counter. On the counter is: a bowl (containing a red apple, a banana, an orange, a potato), a drawer.
	a cupboard. The cupboard door is closed. 
	a freezer. The freezer door is closed. 
	a fridge. The fridge door is closed. 
	a glass jar (containing a substance called sodium chloride)
	a lighter
	a oven, which is turned off. The oven door is closed. 
	a painting
	a sink, which is turned off. In the sink is: nothing.
	a substance called soap
	a stopwatch, which is deactivated. 
	a stove, which is turned off. On the stove is: nothing.
	a table. On the table is: a glass cup (containing nothing).
	a thermometer, currently reading a temperature of 10 degrees celsius
You also see:
	A door to the bathroom (that is open)
	A door to the hallway (that is closed)
	A door to the outside (that is closed)
> 
Action: open oven

...

Action: go hallway

...

Action: go workshop

```
This room is called the workshop. In it, you see: 
	the agent
	a substance called air
	a table. On the table is: a battery, a green wire, a metal pot (containing a substance called lead), a orange light bulb, which is off, a orange wire, a red wire, a switch, which is off, a violet light bulb, which is off, a yellow light bulb, which is off.
	a ultra low temperature freezer. The ultra low temperature freezer door is closed. 
You also see:
	A door to the hallway (that is open)
> 
Action: pick up metal pot

You move the metal pot to the inventory.

Action: go kitchen

\end{Verbatim}

Above we provide an example of a spatial reasoning failure from Claude-3.5-Haiku in ScienceWorldMelt. The agent navigates to the workshop, through the hallway and thus should realize the path is kitchen to hallway to workshop. However, despite this, the agent attempts the action \textcolor{red}{go kitchen}.

\subsection{Deductive Reasoning Failure Example}
\begin{Verbatim}[breaklines=true, breakanywhere=true]
Task Description:Your task is to freeze lead. First, focus on the substance. Then, take actions that will cause it to change its state of matter.

...

Action: look around   

This room is called the kitchen. In it, you see: 
	the agent
	a substance called air
	a chair. On the chair is: nothing.
	a counter. On the counter is: a bowl (containing a red apple, a banana, an orange, a potato), a drawer.
	a cupboard. The cupboard door is closed. 
	a freezer. The freezer door is closed. 
	a fridge. The fridge door is closed. 
	a glass jar (containing a substance called sodium chloride)
	a lighter
	a oven, which is turned off. The oven door is closed. 
	a painting
	a sink, which is turned off. In the sink is: nothing.
	a substance called soap
	a stopwatch, which is deactivated. 
	a stove, which is turned off. On the stove is: nothing.
	a table. On the table is: a glass cup (containing nothing).
	a thermometer, currently reading a temperature of 10 degrees celsius
You also see:
	A door to the bathroom (that is open)
	A door to the hallway (that is closed)
	A door to the outside (that is closed)
    
Action: open freezer        

\end{Verbatim}

Above we provide an example of a deductive reasoning failure from Claude-3.7-Sonnet (non-thinking) in ScienceWorldFreeze. While looking for lead, the agent checks a kitchen freezer, an extremely unlikely place to find lead.

\subsection{Inductive Reasoning Failure Example}
\begin{Verbatim}[breaklines=true, breakanywhere=true]
...

If you pray enough, your prayers may be answered.

Action: pray

If you pray enough, your prayers may be answered.

Action: pray

If you pray enough, your prayers may be answered.

Action: pray

If you pray enough, your prayers may be answered.

Action: pray

...

\end{Verbatim}

Above we provide an example of an inductive reasoning failure from Claude-3.5-Haiku in \zork. While the game environment returns a signal that praying might result in some change, Claude-3.5-Haiku continues to do this for 36 steps, far past when most reasonable players would have moved on.

\subsection{Grounded Reasoning Failure Example}
\begin{Verbatim}[breaklines=true, breakanywhere=true]
This room is called the workshop. In it, you see: 
	the agent
	a substance called air
	a table. On the table is: a battery, a green wire, a orange light bulb, which is off, a orange wire, a red wire, a switch, which is off, a violet light bulb, which is off, a yellow light bulb, which is off.
	a ultra low temperature freezer. The ultra low temperature freezer door is closed. 
You also see:
	A door to the hallway (that is open)

...

Action: go to hallway

You move to the hallway.

...

Action: go to hallway     

You move to the hallway.

Action: go to kitchen       

You move to the kitchen.

Action: use lighter on lead 

The lighter heats up the lead a small amount.

Action: take red wire


\end{Verbatim}

Above we provide an example of a grounded reasoning failure from Claude-3.7-Sonnet (thinking) in ScienceWorldBoil. We see the agent attempt to interact with an item in its context despite having moved to another location where the item is no longer present.

\section{Data Contamination Evidence from Thinking Traces: Examples}\label{app:data_contamination}
\begin{Verbatim}[breaklines=true, breakanywhere=true]
The blue glow of my sword suggests there's a danger nearby, likely something from Zork lore (like a grue). Let me explore the available paths carefully. I have the lantern on, which should protect me from grues in dark areas. Let me try going north first.
\end{Verbatim}
An example of a thinking trace from the Claude-3.7-Sonnet thinking mode that leads us to strongly suspect data contamination. Most notably, the thinking trace references a \textit{grue}, a creature from \zork mythology that had not at all been previously mentioned in the game's history.
\section{Benchmark Run Command and Settings}\label{app:command}
The following Bash commands run one model configuration across the 5 seeds. All arguments not included are default values: action and chain-of-thought temperature 0, the game-specific default game seed, and admissible commands off. \texttt{\$LLM} is the model name.

Non-thinking LLMs use the zero-shot agent:

\begin{Verbatim}[breaklines=true, breakanywhere=true]
for i in {1..5}; do
  python benchmark.py --agent agents/llm.py zero-shot --envs jericho scienceworld textworld textworld_express alfworld --llm $LLM --context-limit 100 --nb-steps 100 --conversation --wandb --seed "20241106$i" &
  sleep 60
done
\end{Verbatim}

Thinking LLMs use the reasoning agent. Rows labeled (thinking) use reasoning effort \texttt{high}, and rows labeled (low), (medium) or (high) use that effort:

\begin{Verbatim}[breaklines=true, breakanywhere=true]
for i in {1..5}; do
  python benchmark.py --agent agents/reasoning.py reasoning --envs jericho scienceworld textworld textworld_express alfworld --llm $LLM --reasoning-effort high --context-limit 100 --nb-steps 100 --conversation --wandb --seed "20241106$i" &
  sleep 60
done
\end{Verbatim}

Rows labeled (1024) cap the chain of thought at 1024 tokens: they pass \verb|--cot-max-tokens 1024| in place of \verb|--reasoning-effort high|. Claude models released after claude-opus-4.6 also add \verb|--tool-actions| (Appendix \ref{app:tool_control}); as of \opusfivefive, the action tool is no longer needed.

\section{Predicting Score Increases Past 100 Steps: Additional Details}\label{app:100steps}
We evaluate ``deserves more than 100 steps'' using actual \jericho @100$\rightarrow$@300 gain as the target. Gains are computed on the ground-truth 54-game \jericho denominator with best-so-far normalized score; if a run terminates before a target budget, its terminal best-so-far score is carried forward. Predictors are computed from first-100-step behavior, primarily on non-\jericho trajectories, so the prediction does not depend on observing longer \jericho rollouts.

Across the seven-model extended-budget set (\texttt{claude-opus-4.6}, \texttt{claude-sonnet-4.6}, \texttt{gpt-5.1}, \texttt{gpt-5.4}, \texttt{claude-haiku-4.5}, \texttt{gpt-5.4-mini}, \texttt{gpt-5.4-nano}), the strongest predictors are trajectory-shape features rather than raw extension labels. \texttt{novelty\_decay} measures whether observations in steps 71--100 remain diverse relative to steps 1--30; high values indicate that the agent is still reaching distinct states rather than cycling locally, and this reaches Spearman $\rho = 0.929$ and Pearson $r = 0.965$. \texttt{plateau\_then\_recover\_rate} measures how often a run stalls for at least 15 steps after step 25 and then improves before step 100; it distinguishes temporary search bottlenecks from true exhaustion and reaches $\rho = 0.919$, $r = 0.801$. \texttt{best\_score\_recency} records how late the run first reaches its best score within the first 100 steps; late best scores indicate that progress has not saturated ($\rho = 0.786$, $r = 0.935$). Late score movement itself is captured by \texttt{gain\_76\_100}, the mean score gain in steps 76--100 ($\rho = 0.750$, $r = 0.888$), and \texttt{late\_gain\_consistency}, the fraction of runs with any gain in steps 76--100 ($\rho = 0.750$, $r = 0.845$).

\scienceworld provides a useful control: top models are already near ceiling by 100 steps, so post-100 score changes are small, while weaker models show essentially no consistent improvement. The pre-\scienceworld environments are similarly close to saturated under these agents, leaving \jericho as the framework where additional budget produces the clearest and most systematic gains.

The competency-based detector adds an interpretable diagnostic. Non-\jericho Rule Learning failure rate has $\rho = -0.821$ with later \jericho gain, while aggregate Situated Awareness failure rate is much weaker ($\rho = -0.393$). This suggests that extension benefit is limited less by occasional grounded-context mistakes than by repeated-action loops and failure to adapt to feedback. In practical terms, extra budget helps models that are still exploring productively; it does not help models whose first 100 steps already show high loopiness or disengagement.

\section{Impact of Privileged Information in RL}\label{app:RL}
We evaluate the impact of directly providing privileged and domain information to the model during RL fine-tuning by ablating over the inclusion of the admissible actions and the contents of \cmd{help} in the prompt across the synthetic environments \abbrtextworld, \abbrtwx, \abbralfworld, and \abbrscienceworld. 
Given the model finds little success in \abbrscienceworld, we exclude \jericho which is even more challenging.

\begin{figure*}[ht]
\centering
 \includegraphics[width=\textwidth]{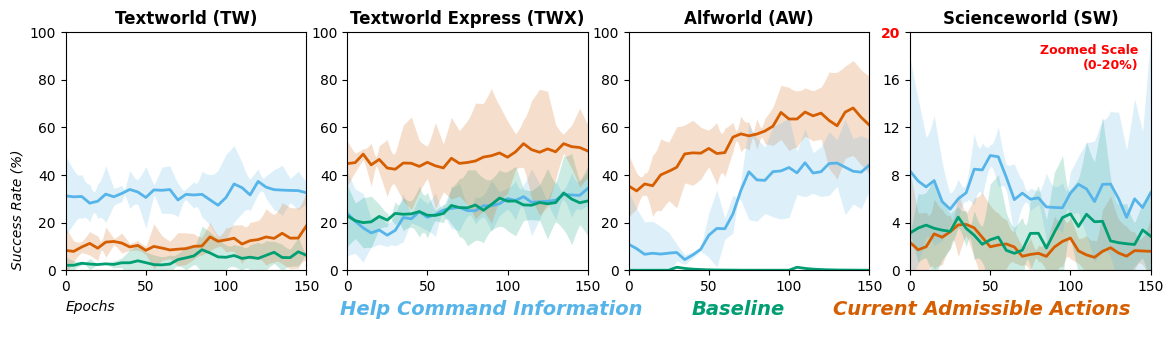}
\caption{Inclusion of \textcolor[HTML]{D55E00}{admissible actions} and \textcolor[HTML]{56B4E9}{the feedback from \cmd{help}} in an agent's prompt during RL fine-tuning can result in a significant increase in performance over an \textcolor[HTML]{009E73}{unscaffolded baseline}. \abbrtwx shows no change due to \cmd{help} providing only the task description.}
    \label{fig:tales_rl_think_graphs}
\end{figure*}

We use the verl-agent \citep{feng2025group} framework due to their pre-existing results on \abbralfworld. 
Integration of \ours is done standalone.
See Appendix~\ref{app:hyperparameters} for hyperparameters used.
Qwen3-8B~\citep{qwen3technicalreport} is used due to its strong performance as a thinking model. 
Figure~\ref{fig:tales_rl_think_graphs} shows results, averaged over three seeds.

\textbf{Simpler environments can be made significantly easier by the inclusion of privileged information.} 
The requirement to explore and iterate over what action commands are accepted by the game parser can be one of the most challenging aspects of an environment. 
The ability to do this iteration and keep track of the current state of the problem space tests a model's inductive, spatial, and grounded reasoning by forcing the choice of action to occur in the unbounded space of natural language.
Bounding the space from which the agent must select its action to a subset of precomputed commands can drastically reduce the difficulty of the environment, as we can see by the extremely large differential in performance for \abbrtwx and \abbralfworld when trained with and without admissible actions included in the prompt.

\textbf{Training LLM agents with privileged environment information should be avoided due to a failure to generalize.}
While capable of improving performance in simpler environments, leveraging privileged information as scaffolding can result in overfitting to the specified environment due to overreliance. 
Taking the admissible-actions baseline as an example, we see significant improvements in \abbrtwx and \abbralfworld.
However, we fail to see a similar improvement in \abbrtextworld with performance even dropping in \abbrscienceworld.
These environments significantly feature actions that can leave the agent entirely unable to complete the given task. 
Without context, these actions can appear innocuous when embedded in the admissible actions.
\abbrscienceworld in particular has sufficient complexity such that the admissible actions can quickly grow combinatorially large, saturating the context with useless information.
While not visible from a graph plotting success rate against epochs, computational cost on the environment can also become an issue\footnote{On an H100 cluster, the average wall-clock time for the training runs of \abbrscienceworld with admissible actions was roughly 54 hours, with the baseline and \cmd{help}-baseline needing only an average of 22 hours.}.

\textbf{Domain information can generalize as a strong deductive prior.}
With the exception of \abbrtwx, where \cmd{help} only repeats the task objective, we see improvements in all frameworks in the \cmd{help} baseline where the information included in the prompt is information the agent would be able to otherwise discover on its own.
While this still decreases the difficulty of the environment, this otherwise accessible domain information acts as more natural guidance compared to the privileged information from the admissible actions.

\textbf{LLM agents face difficulty improving performance when exposed to a large number of diverse environments simultaneously.} 
In the \cmd{help} and admissible actions baselines, we see an upward trend of success in \abbralfworld, where all tasks involve relatively uniform task specifications.
However, in \abbrtextworld, \abbrtwx, and \abbrscienceworld, we see little to no learning occurring due to the extremely noisy reward signal from a wide range of task variety and difficulty.

\section{Action-Tool Control}\label{app:tool_control}
Claude models released after claude-opus-4.6 are evaluated with an action tool; as of \opusfivefive, the action tool is no longer needed. One sentence is added to the system prompt (``Instead of typing the command directly, submit it by calling the \texttt{submit\_command} tool with the \texttt{command} argument.''), and the \texttt{command} argument is sent to the game as the action. To check that the tool does not confound comparisons with models evaluated without it, we re-ran claude-opus-4.6 (thinking) with the tool enabled and all other settings unchanged. Table \ref{tab:tool_control} compares the two configurations.

\begin{table}[h!]
    \footnotesize
    \centering
    \renewcommand{\arraystretch}{1.2}
    \resizebox{\textwidth}{!}{
    \begin{tabular}{l c c c c c c}
        \toprule
        \textbf{Action tool} & \textworld & \twx & \alfworld & \scienceworld & \jericho & \textbf{\ours Score} \\
        \midrule
        Off                               & 100.0 & 95.7 & 100.0 & 92.0 & 16.2 $\pm$ 1.2 & 60.4 $\pm$ 1.4 \\
        On (Table \ref{ref:score_table})  & 100.0 & 98.6 & 100.0 & 91.0 & 17.8 $\pm$ 1.8 & 61.2 $\pm$ 0.8 \\
        \bottomrule
    \end{tabular}
    }
    \caption{claude-opus-4.6 (thinking) with and without the action tool, 5 seeds each. $\pm$ is the standard deviation across seeds of the per-seed score.}
    \label{tab:tool_control}
\end{table}

\newpage
\section*{NeurIPS Paper Checklist}

\begin{enumerate}

\item {\bf Claims}
    \item[] Question: Do the main claims made in the abstract and introduction accurately reflect the paper's contributions and scope?
    \item[] Answer: \answerYes{} %
    \item[] Justification: Yes, we explicitly outline contributions in our introduction. Contribution 1: This paper, with code publicly available at \url{https://github.com/microsoft/tale-suite}. Contribution 2: Sections \ref{sec:tales} and \ref{ref:zero-shot}. Contribution 3: \url{https://huggingface.co/datasets/ccui46/tale_suite_trajectories}. Contribution 4: Section \ref{sec:error_analysis}.
    Contribution 5: Due to space limitations, we provide the top 10 overall-scoring models in Table \ref{ref:score_table} but include all scores on a per game and LLM basis in the appendix.

\item {\bf Limitations}
    \item[] Question: Does the paper discuss the limitations of the work performed by the authors?
    \item[] Answer: \answerYes{} %
    \item[] Justification: We include a distinct limitations section in our conclusion (Section \ref{sec:conclusion}).

\item {\bf Theory assumptions and proofs}
    \item[] Question: For each theoretical result, does the paper provide the full set of assumptions and a complete (and correct) proof?
    \item[] Answer: \answerNA{}
    \item[] Justification: The paper does not include theoretical results.

    \item {\bf Experimental result reproducibility}
    \item[] Question: Does the paper fully disclose all the information needed to reproduce the main experimental results of the paper to the extent that it affects the main claims and/or conclusions of the paper (regardless of whether the code and data are provided or not)?
    \item[] Answer: \answerYes{} %
    \item[] Justification: Aside from provided code, we also provide an outline of the settings used in our experiments in Section \ref{ref:zero-shot}, with the exact seeds given in Appendix \ref{app:command}. We further include all code used for analysis and release all trajectories generated by LLMs. An example template command used to run experiments across the seeds we used can be found in Appendix \ref{app:command}. However, we acknowledge that some API models may no longer be available.

\item {\bf Open access to data and code}
    \item[] Question: Does the paper provide open access to the data and code, with sufficient instructions to faithfully reproduce the main experimental results, as described in supplemental material?
    \item[] Answer: \answerYes{} %
    \item[] Justification: Our code used to run all experiments, as well as tables and figures in this paper, is publicly available at \url{https://github.com/microsoft/tale-suite}, and all trajectories at \url{https://huggingface.co/datasets/ccui46/tale_suite_trajectories}.

\item {\bf Experimental setting/details}
    \item[] Question: Does the paper specify all the training and test details (e.g., data splits, hyperparameters, how they were chosen, type of optimizer) necessary to understand the results?
    \item[] Answer: \answerYes{} %
    \item[] Justification: As our paper is a benchmark, our main results involve no training. We have baseline results that we detail how to replicate.
    We include additional RL experiments as an ablation in Appendix \ref{app:RL}. Code and hyperparameters used are based on the defaults used by the framework authors in their experiments.

\item {\bf Experiment statistical significance}
    \item[] Question: Does the paper report error bars suitably and correctly defined or other appropriate information about the statistical significance of the experiments?
    \item[] Answer: \answerYes{} %
    \item[] Justification: Main results are averaged over 5 seeds (claude-fable-5: one seed); the extended-budget runs in Section~\ref{sec:quant} use one seed and the RL experiments in Appendix~\ref{app:RL} three seeds. Standard deviations across seeds are reported in Appendix~\ref{app:all_stds} and Table~\ref{tab:tool_control}.

\item {\bf Experiments compute resources}
    \item[] Question: For each experiment, does the paper provide sufficient information on the computer resources (type of compute workers, memory, time of execution) needed to reproduce the experiments?
    \item[] Answer: \answerYes{} %
    \item[] Justification: We outline the hardware used, compute funding, run time and token use in Appendix \ref{app:resources}.
    Notably, we also provide results from a number of closed-source, deprecated API models. 
    As these are no longer available, experiments run cannot be replicated. However, we release transcripts from these runs.
    
\item {\bf Code of ethics}
    \item[] Question: Does the research conducted in the paper conform, in every respect, with the NeurIPS Code of Ethics \url{https://neurips.cc/public/EthicsGuidelines}?
    \item[] Answer: \answerYes{} %
    \item[] Justification: Our work adheres to the NeurIPS Code of Ethics. All human labor was done by researchers on the project and thus does not fall within the domain of human subjects research.

\item {\bf Broader impacts}
    \item[] Question: Does the paper discuss both potential positive societal impacts and negative societal impacts of the work performed?
    \item[] Answer:  \answerNA{} %
    \item[] Justification: We do not foresee any direct societal impacts of the work performed outside of second-hand outcomes that may come as a result of introducing additional methods of evaluating agent reasoning capabilities. 
    
\item {\bf Safeguards}
    \item[] Question: Does the paper describe safeguards that have been put in place for responsible release of data or models that have a high risk for misuse (e.g., pre-trained language models, image generators, or scraped datasets)?
    \item[] Answer: \answerNA{} %
    \item[] Justification: The paper poses no such risks. 

\item {\bf Licenses for existing assets}
    \item[] Question: Are the creators or original owners of assets (e.g., code, data, models), used in the paper, properly credited and are the license and terms of use explicitly mentioned and properly respected?
    \item[] Answer: \answerYes{} %
    \item[] Justification: To the best of our knowledge, the creators or original owners of assets used in the paper are properly credited, and license and terms of use properly respected. For games with commercial or restrictive licensing, we release only the agent actions, seeds, and scores, with regeneration instructions for anyone with access to the relevant story file.

\item {\bf New assets}
    \item[] Question: Are new assets introduced in the paper well documented and is the documentation provided alongside the assets?
    \item[] Answer: \answerYes{} %
    \item[] Justification: We provide code and documentation for the assets introduced by this work.

\item {\bf Crowdsourcing and research with human subjects}
    \item[] Question: For crowdsourcing experiments and research with human subjects, does the paper include the full text of instructions given to participants and screenshots, if applicable, as well as details about compensation (if any)? 
    \item[] Answer: \answerNA{} %
    \item[] Justification: The paper does not include crowdsourcing experiments or research with human subjects.

\item {\bf Institutional review board (IRB) approvals or equivalent for research with human subjects}
    \item[] Question: Does the paper describe potential risks incurred by study participants, whether such risks were disclosed to the subjects, and whether Institutional Review Board (IRB) approvals (or an equivalent approval/review based on the requirements of your country or institution) were obtained?
    \item[] Answer: \answerNA{} %
    \item[] Justification: The paper does not include human subjects.

\item {\bf Declaration of LLM usage}
    \item[] Question: Does the paper describe the usage of LLMs if it is an important, original, or non-standard component of the core methods in this research? Note that if the LLM is used only for writing, editing, or formatting purposes and does \emph{not} impact the core methodology, scientific rigor, or originality of the research, declaration is not required.
    \item[] Answer: \answerNA{} %
    \item[] Justification: Based on our understanding of this question, we answer NA. LLMs were not used to aid in development of research questions, research directions, or research methods. 

\end{enumerate}

\end{document}